\documentclass[letterpaper]{article} 
\usepackage{aaai25}  
\usepackage{times}  
\usepackage{helvet}  
\usepackage{courier}  
\usepackage[hyphens]{url}  
\usepackage{graphicx} 
\usepackage{natbib}  
\usepackage{caption} 
\usepackage{algorithm}
\usepackage{algorithmic}
\usepackage{amsmath}
\usepackage{amssymb}
\usepackage[dvipsnames]{xcolor}
\usepackage{pifont}
\newcommand{\cmark}{\ding{51}}
\newcommand{\xmark}{\ding{55}}
\usepackage{subcaption}
\usepackage{multirow}
\usepackage{newfloat}
\usepackage{listings}
\DeclareCaptionStyle{ruled}{labelfont=normalfont,labelsep=colon,strut=off} 
\floatstyle{ruled}
\newfloat{listing}{tb}{lst}{}
\floatname{listing}{Listing}
\newtheorem{theorem}{Theorem}
\newtheorem{remark}{Remark}
\newtheorem{assumption}{Assumption}

\newtheorem{lemma}{Lemma}
\newtheorem{definition}{Definition}
\newtheorem{proposition}{Proposition}

\DeclareMathOperator{\Er}{E}
\DeclareMathOperator{\N}{N}
\DeclareMathOperator{\diam}{diam}
\DeclareMathOperator{\CE}{CE}

\usepackage{xcolor}

\title{Open-Set Heterogeneous Domain Adaptation: Theoretical Analysis and Algorithm}
\author{
    Thai-Hoang Pham\textsuperscript{\rm 1,\rm 2},
    Yuanlong Wang\textsuperscript{\rm 1,\rm 2},
    Changchang Yin\textsuperscript{\rm 1,\rm 2},
    Xueru Zhang\textsuperscript{\rm 1},
    Ping Zhang\textsuperscript{\rm 1,\rm 2}
}
\affiliations{
    \textsuperscript{\rm 1}Department of Computer Science and Engineering, The Ohio State University, USA\\
    \textsuperscript{\rm 2}Department of Biomedical Informatics, The Ohio State University, USA\\
    \{pham.375,wang.16050,yin.731,zhang.12807,zhang.10631\}@osu.edu
}

\usepackage{bibentry}

\begin{document}

\maketitle

\begin{abstract}
Domain adaptation (DA) tackles the issue of distribution shift by learning a model from a source domain that generalizes
to a target domain. However, most existing DA methods are designed for scenarios where the source and target domain data lie within the same feature space, which limits their applicability in real-world situations. 
Recently, heterogeneous DA (HeDA) methods have been introduced to address the challenges posed by heterogeneous feature space between source and target domains. Despite their successes, current HeDA techniques fall short when there is a mismatch in both feature and label spaces. To address this, this paper explores a new DA scenario called open-set HeDA (OSHeDA). In OSHeDA, the model must not only handle heterogeneity in feature space but also identify samples belonging to novel classes. To tackle this challenge, we first develop a novel theoretical framework that constructs learning bounds for prediction error on target domain. Guided by this framework, we propose a new DA method called Representation Learning for OSHeDA (RL-OSHeDA). This method is designed to simultaneously transfer knowledge between heterogeneous data sources and identify novel classes. Experiments across text, image, and clinical data demonstrate the effectiveness of our algorithm. Model implementation is available at \url{https://github.com/pth1993/OSHeDA}.
\end{abstract}

\section{Introduction}

Machine learning (ML) techniques have achieved unprecedented success over the past decades in numerous areas~\cite{lecun2015deep}. However, ML systems are often built on the assumption that training and testing data are independent and identically distributed, which is commonly violated in real-world applications where the environment changes during model deployment. Existing works have shown that the performance of ML models often deteriorates due to distribution shifts between training and testing data~\cite{ben2010theory,quinonero2022dataset}. To learn a model robust under distribution shifts,
domain adaptation (DA)~\cite{ben2010theory,mansour2009domain} has been proposed to transfer knowledge from a source domain that possesses abundant labeled data to a different but relevant target domain. 

\begin{figure}[t]
    \centering
    \includegraphics[width=0.93\linewidth]{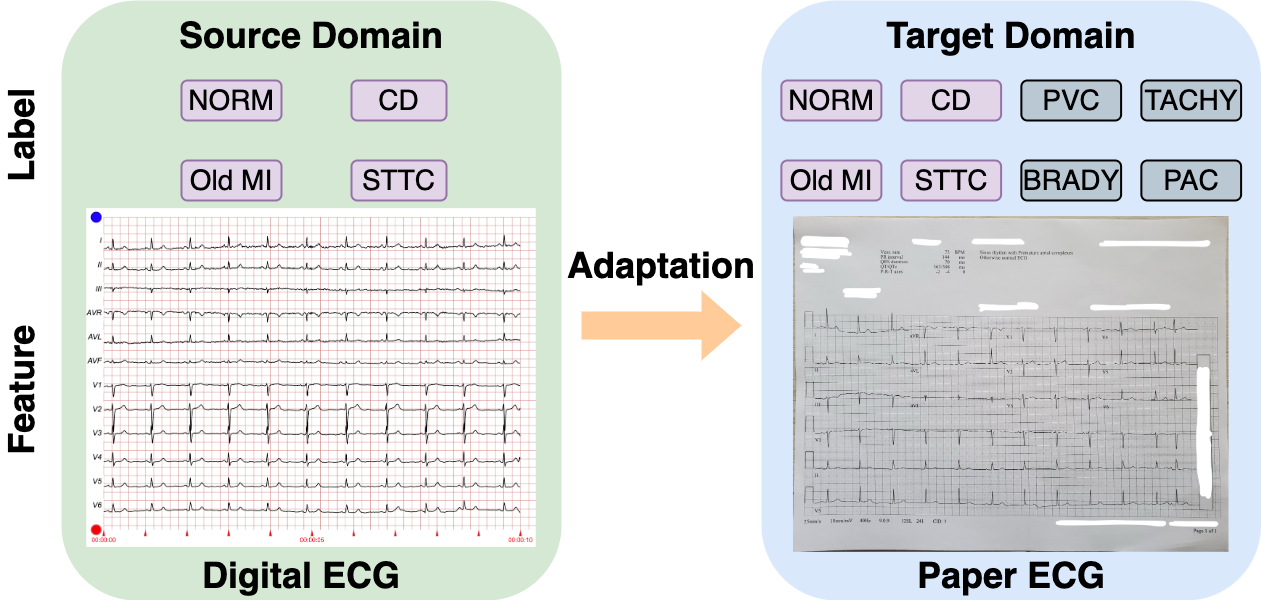}
    \caption{A motivating example about OSHeDA in the context of screening diseases using electrocardiogram (ECG) data. While digital ECGs comprise the majority of labeled data for training ML models for disease screening, physical or paper ECGs 
    remain prevalent worldwide. Thus, the transfer of knowledge from digital ECG datasets is essential to support the training of ML models that analyze paper ECGs. Moreover, ML systems must effectively manage rare abnormalities (indicated with gray boxes), which may not be available in training data, to prevent misdiagnosis.}
    \label{fig:1}
\end{figure}

Existing DA methods, however, typically assume a homogeneous scenario where the source and target domains have the same 
feature and label spaces. Consequently, they may fail when 
the source and target domain data lie in different spaces. 
For example, 
heterogeneous feature space is common in biomedical domains in which medical terms undergoes continuous evolution, leading to the retirement of outdated terms (e.g., ICD-9 coding system) and the introduction of novel ones (e.g., ICD-10 coding system)~\cite{grief2016simulation}. In such cases, acquiring training data that seamlessly aligns with target domain’s feature space can be impractical or excessively costly. Heterogeneous domain adaptation (HeDA) methods have emerged to handle the heterogeneity observed in distinct feature spaces, which often vary significantly between domains~\cite{li2020simultaneous,zhao2022semantic}. 

Despite the significant successes achieved by these HeDA methods, they face a major limitation: current HeDA techniques can only address heterogeneity in feature space and are inadequate when there is a mismatch in both feature and label spaces. This limitation restricts the practical application of HeDA methods in many real-world scenarios because neglecting label mismatch, such as new classes emerging in the target domain, can lead to negative transfer effects from the source to the target domains~\cite{liu2019separate}.

To overcome this limitation, this study explores a new DA scenario called open-set heterogeneous domain adaptation (OSHeDA). In OSHeDA, ML methods must not only manage heterogeneity in feature space between source and target domains but also identify samples belonging to novel classes in the target domain. 
Figure~\ref{fig:1} illustrates a real example from clinical applications for this novel learning scenario. In this instance, the adaptation process aims to transfer knowledge from digital electrocardiogram (ECG) to paper ECG formats (\textit{heterogeneous}). 
Moreover, ML models for ECG-based diagnosis must also detect rare abnormalities that were not included in the training data (\textit{open-set}). 

To address the challenge of feature and label mismatch in OSHeDA, we first develop a novel theoretical analysis that constructs learning bounds for the prediction error of ML models on the target domain. Guided by this theoretical analysis, we then design a novel representation learning method named \textbf{R}epresentation \textbf{L}earning for \textbf{O}pen-\textbf{S}et \textbf{He}terogeneous \textbf{D}omain \textbf{A}daptation (RL-OSHeDA). This method is proposed to transfer knowledge between heterogeneous data sources and identify novel class simultaneously. Unlike existing HeDA methods, RL-OSHeDA transfer knowledge from source to target domains by aligning representations between source and target domains for known classes while also enforcing the representations of novel class in target domains to move apart from the known classes of the source and  target domains. To effectively identify samples from novel class within unlabeled data, RL-OSHeDA optimizes a non-negative risk estimator for open-set and employs pseudo labeling to enrich the labeled data. 

In summary, the contributions of our work are as follows: 
\begin{itemize}
    \item We conduct a theoretical analysis to establish learning bounds in the OSHeDA scenario. This analysis emphasizes the importance of minimizing the distance between source and target domains for known classes, while maximizing the separation from unknown classes. Moreover, we investigate the impact of pseudo-label and the non-negative risk estimator for open-set in OSHeDA.
    \item Motivated by the theoretical results, we propose a novel algorithm (RL-OSHeDA) based on representation learning to transfer knowledge from source to target domains.
    \item We conduct experiments on real data from clinical, computer vision, and natural language processing domains to validate the effectiveness of our method for OSHeDA.
\end{itemize}

\section{Related Works}

In this section, we summarize existing research from related areas including heterogeneous domain adaptation, open-set domain adaptation, and open-set semi-supervised learning.

\noindent \textbf{Heterogeneous Domain Adaptation (HeDA).} HeDA aims to transfer knowledge across domains with distinct feature spaces and data distributions. Depending on whether unlabeled target data are used in the adaptation process, HeDA approaches are categorized into three types: supervised, semi-supervised, and unsupervised methods. Supervised HeDA methods utilize ample labeled data from both source and target domains for adaptation~\cite{hoffman2013efficient,li2013learning,hoffman2014asymmetric}. In contrast, semi-supervised HeDA methods require only a small number of labeled target domain data and utilize unlabeled instances from the target domain to facilitate transfer~\cite{yao2020discriminative,li2020simultaneous,fang2022semi,zhao2022semantic,yao2019heterogeneous}. Finally, unsupervised HeDA methods operate without any labeled target data, relying solely on unlabeled instances and labeled source data to align cross-domain feature representations~\cite{shen2018unsupervised,li2018heterogeneous,zou2018unsupervised}. However, successful transfer in unsupervised settings depends on specific assumptions about domain relationships~\cite{liu2020heterogeneous}. 

\noindent \textbf{Open-Set Domain Adaptation (OSDA).} OSDA represents a realistic and challenging scenario in DA where the target domain includes instances whose classes are not observed in the source domain, alongside a shift in feature distribution between the two domains. In contrast to the OSHeDA, OSDA assumes a homogeneous feature space between the source and target domains. Existing approaches for OSDA can be categorized into two main groups: adversarial learning and self-supervised learning. Adversarial learning methods employ adversarial networks to detect unknown samples and align the distributions of known samples between the source and target domains~\cite{saito2018open,luo2020progressive}. On the other hand, self-supervised learning methods utilize techniques like data augmentation to distinguish between known and unknown instances in the target domain~\cite{bucci2020effectiveness,li2021domain}.

\noindent \textbf{Open-Set Semi-supervised Learning (OS-SSL).} OS-SSL is a SSL scenario that addresses novel classes within unlabeled data during training. Unlike OSDA, OS-SSL requires only a small amount of labeled data. However, it assumes that both labeled and unlabeled data of known classes are drawn from the same distribution, and this setting does not account for novel classes during inference. Methods designed for OS-SSL can be broadly categorized into two types based on how they detect novel classes: criterion-based approaches and detector-based approaches. Criterion-based approaches use heuristic rules to identify novel classes~\cite{chen2020semi,huang2022they,du2023semi,he2022safe}. In contrast, detector-based approaches employ parameterized detectors to filter outliers~\cite{yu2020multi,huang2021trash,wang2023out,saito2021openmatch}. 
\section{Problem Formulation}
\textbf{Notations.} 
Let $\mathcal{X}^d$ and $\mathcal{Y}^d$ denote the feature and label spaces of a domain $d$ associated with a distribution $P_d(X_d,Y_d): \mathcal{X}^d\times \mathcal{Y}^d \to [0,1]$ and labeling function $h_{d}: \mathcal{X}^{d} \to \Delta\left(\mathcal{Y}^d\right)$ where $X_d$ and $Y_d$ are random variables that take values in $\mathcal{X}^d$ and $\mathcal{Y}^d$, and $\Delta\left(\mathcal{Y}^d\right)$ is a probability simplex over $\mathcal{Y}^d$. Consider a model $h: \mathcal{X}^d \to \Delta \left(\mathcal{Y}^{d}\right)$, then the \textit{expected error} of $h$ under domain $d$  for some loss function $L:  \Delta \left(\mathcal{Y}^{d}\right) \times \mathcal{Y}^d \rightarrow \mathbb{R}_+$ (e.g., 0-1, cross-entropy loss)  can be defined as
$\Er\left(P_d, h\right) = \mathbb{E}_{P_d}\left[L\left(h\left(X_d\right), Y_d\right)\right].$

\noindent \textbf{Open-Set Heterogeneous Domain Adaptation (OSHeDA) Setup.} 
In DA, we consider $d \in \{s, t\}$ where $s$ and $t$ denote the source and target domains, respectively. Different from conventional DA setup where feature and label spaces remain the same between source and target domains, in OSHeDA, we have $\mathcal{X}^s \neq \mathcal{X}^t$ (\textit{heterogeneous}) and $\mathcal{Y}^s \subset  \mathcal{Y}^t$ (\textit{open-set}). Because $\mathcal{Y}^s \subset  \mathcal{Y}^t$, we use $Y$ to denote the random variable of label in both source and target domains, and we have $P_s\left(Y \in \mathcal{Y}^t \setminus \mathcal{Y}^s\right) = 0$. Moreover, classes in the sets $\mathcal{Y}^t \setminus \mathcal{Y}^s $ are referred to as unknown in our setting. Given sets of samples $D_s = \{ x_i^s, y_i^s \}_{i=1}^{n_s} \overset{i.i.d}{\sim} P_s\left(X_s,Y\right)$ (\textit{source dataset}), $D_{t_l} = \{ x_i^t, y_i^t \}_{i=1}^{n_{t_l}} \overset{i.i.d}{\sim} P_t\left(X_t,Y | Y \in \mathcal{Y}^s\right)$ 
(\textit{labeled target dataset}), and $D_{t_u} = \{ x_i^t \}_{i=1}^{n_{t_u}} \overset{i.i.d}{\sim}  P_t\left(X_t,Y\right)$ (\textit{unlabeled target dataset}), where $n_s, n_{t_l}, n_{t_u}$ are size of datasets and $n_{t_l} \ll n_s, n_{t_u}$, the goal of OSHeDA is to learn a model $h: \mathcal{X}^t \to \Delta\left(\mathcal{Y}^t\right)$ from $D_s, D_{t_l}, D_{t_u}$ such that the expected error on the target domain $\Er\left(P_t, h\right)$ is small.

\noindent \textbf{Representation learning.} 
Representation learning is a common approach for transferring knowledge from a source to a target domain in DA~\cite{zhao2019learning,ganin2016domain,albuquerque2019generalizing,pham2023fairness}, and we will leverage this method in OSHeDA. 
Specifically, it maps the input spaces $\mathcal{X}^s$ and $\mathcal{X}^t$ of the source and target domains to a shared representation space $\mathcal{Z}$ using two representation mappings: $f_s: \mathcal{X}^s \to \mathcal{Z}$ and $f_t: \mathcal{X}^t \to \mathcal{Z}$. 
A shared classifier $h: \mathcal{Z} \to \Delta(\mathcal{Y}^t)$ can then be employed to make predictions from this representation space. Notably, $h$ can be utilized for both domains because $\mathcal{Y}^s \subset \mathcal{Y}^t$. 

\section{Theoretical Analysis}\label{sec:theory}

In our analysis, we consider Jensen–Shannon (JS) divergence ($\mathcal{D}_{JS}$) as the statistical distance between two domains. 
While different distances~\cite{ben2010theory} were used in domain adaptation literature, we adopt JS divergence because it is aligned with the training objective of adversarial learning~\cite{goodfellow2014generative}, a technique used in many representation learning-based domain adaptation works~\cite{zhao2019learning,ganin2016domain,pham2023fairness}. Next, we present the main theorems, with detailed proofs provided in Appendix~\ref{app:proof}. 

\subsection{Learning bounds for OSHeDA (infinite case)}\label{sec:infinite}

To simplify notations used in our following analysis, we denote $P_{t,k}(\cdot) = P_{t}(\cdot|Y \in \mathcal{Y}^s)$ and $P_{t,u}(\cdot) = P_{t}(\cdot|Y \notin \mathcal{Y}^s)$ as the distributions of target domain conditioned on known and unknown classes, respectively. We also introduce two distributions $P^u_s$ and $P^u_t$ induced from $P_s$ and $P_t$ by the two mappings $f^u_s$ and $f^u_t$ such that $f^u_s(X^s,Y) = (X^s, unk)$ and $f^u_t(X^t,Y) = (X^t, unk)$ where $unk$ denotes unknown class. 
In addition, we adopt an assumption commonly used in DA literature~\cite{nguyen2021kl,mansour2009domain,cortes2014domain} as follows.
\begin{assumption}[Bounded loss]\label{ass:1}Assume loss function $L$ defined on input space $\mathcal{X}$ and output space $\mathcal{Y}$ is upper bounded by a constant $C$, i.e., $\forall x \in \mathcal{X}, y \in \mathcal{Y}$, $h \in \mathcal{H}$, we have $L(h(x), y) \leq C$.
\end{assumption}
We note that this assumption is indeed reasonable rather than stringent. For example, while Assumption~\ref{ass:1} does not hold for the cross-entropy loss typically utilized in classification, we can adjust this loss to ensure that it satisfies Assumption~\ref{ass:1}~\cite{phamnon}. 
Based on this assumption, we then provide an upper bound for prediction error on the target domain in OSHeDA as follows.

\begin{theorem}\label{thm:1}
Given a loss function $L$ satisfying Assumption~\ref{ass:1}, then for any $h \in \mathcal{H}, f_s \in \mathcal{F}_s, f_t \in \mathcal{F}_t$, 
we have:
\begin{align*}
     \Er \left(P_t, h \circ f_t \right) &\leq \underbrace{\lambda \Er \left(P_s, h \circ f_s \right)}_{\textbf{source error}}  \nonumber \\  
     &+ \underbrace{\Er \left(P_t^u, h \circ f_t \right) - \lambda \Er \left(P_s^u, h \circ f_s \right)}_{\textbf{open-set difference}} \nonumber \\ 
     &+ \sqrt{2} \lambda C \left( \left(\mathcal{D}_{JS}\left( P_s(Z) \parallel P_{t,k}(Z) \right) \right)^{\frac{1}{2}}  \right. \nonumber \\
     &\underbrace{\left.+ \left(\mathcal{D}_{JS} \left( P_s(Z,Y) \parallel P_{t,k}(Z,Y) \right) \right)^{\frac{1}{2}} \right)}_{\textbf{domain distance}} 
\end{align*}
\end{theorem}
where $\lambda = P_t(Y \in \mathcal{Y}^s)$, $\mathcal{H}$, $\mathcal{F}_s$, $\mathcal{F}_t$ are hypothesis classes for $h, f_s, f_t$, and $P_s(Z)$ and $P_{t,k}(Z)$ are the distributions induced from $P_s(X_s)$ and $P_{t,k}(X_t)$ by $f_s$ and $f_t$, respectively. 

\begin{remark}\label{remark:1}
    The upper bound in Theorem~\ref{thm:1} shed a light on achieving good accuracy on target domain. Specifically, to minimize $\Er \left(P_t, h \circ f_t \right)$, the model need to optimize three terms: \textbf{(i)} the source error $\Er\left(P_s, h \circ f_s \right)$, \textbf{(ii)} the open-set difference $\Er \left(P_t^u, h \circ f_t \right) - \lambda \Er \left(P_s^u, h \circ f_s \right)$, and \textbf{(iii)} the distances of marginal and joint distributions between source domain and target domain conditioned on known labels $\mathcal{D}_{JS}\left( P_s(Z) \parallel P_{t,k}(Z) \right)$ and $\mathcal{D}_{JS}\left( P_s(Z,Y) \parallel P_{t,k}(Z,Y) \right)$. 
\end{remark}

We want to emphasize that minimizing the distance of the joint distribution between the source and target domains, $\mathcal{D}_{JS}\left( P_s(Z,Y) \parallel P_{t,k}(Z,Y) \right)$, requires knowledge of the label distribution in the target domain $P_{t,k}(Y)$. Therefore, access to labeled target data during training is essential to avoid negative transfer. 
Note that the concept of open-set difference is not exclusive to OSHeDA. This term also appears in existing works for OSDA \cite{fang2020open} and positive-unlabeled learning \cite{kiryo2017positive} which are special cases of our setting. 
Thus, this demonstrates the consistency between our work and the existing literature. Next, we present a lower bound for OSHeDA. 

\begin{proposition}\label{thm:3}
Given a loss function $L$ satisfying Assumption~\ref{ass:1}, then for any $h \in \mathcal{H}, f_s \in \mathcal{F}_s, f_t \in \mathcal{F}_t$, we have:
\begin{align*}
\Er \left(P_t, h \circ f_t \right) &\geq \lambda \Er \left(P_{t,k}, h \circ f_t \right) + (1 - \lambda) \Er \left(P_{s}^u, h \circ f_s \right) \nonumber \\
&- \sqrt{2}(1 - \lambda) C \left(\mathcal{D}_{JS} \left( P_s(Z) \parallel P_{t,u}(Z) \right) \right)^{\frac{1}{2}}
\end{align*}
\end{proposition}
where $P_{t,u}(Z)$ is distribution induced from $P_{t,u}(X_t)$ by $f_t$. 

\begin{remark}\label{remark:2}
    Theorem~\ref{thm:1} shows the necessity of reducing $\Er \left(P_{s}, h \circ f_s \right)$ to achieve high accuracy on target domain. However, it may unavoidably increase $ \Er \left(P_{s}^u, h \circ f_s \right)$. This observation, combined with Proposition~\ref{thm:3}, suggests that to avoid the large lower bound for the target error $\Er \left(P_t, h \circ f_t \right)$, we should increase the distance of the marginal distribution between the source domain and the unknown data in target domain, $\mathcal{D}_{JS} \left( P_s(Z) \parallel P_{t,u}(Z) \right)$. In other words, we should segregate the representations of known classes from those of unknown class.
\end{remark}

\subsection{Learning bound for OSHeDA (finite case)}
The learning bounds discussed in Section~\ref{sec:infinite} are only applicable for the setting when we have access to unlimited data from source and target domains. In such cases, minimizing JS divergence of data distribution between these domains is equivalent to achieving invariant representations through adversarial learning~\cite{goodfellow2014generative}. However, we only work with finite data in practice. Thus, we present the following result, which provides a guarantee for using adversarial learning to optimize JS divergence from finite data.


\begin{proposition}[Adapted from \citet{biau2020some}]\label{thm:5} 
The error in minimizing JS divergence of data distributions between source and target domains in representation space, using finite data, is up to $\mathcal{O} \left( \left(1 / \sqrt{n_s} + 1 / \sqrt{n_t} \right) \right)$.
\end{proposition}
where $n_s$ and $n_t$ are the size of source and target datasets.

\begin{remark}
    Proposition~\ref{thm:5} states that the performance of minimizing JS divergence from finite data is proportional to the dataset size. Note that in OSHeDA, we only have access to limited label data from target domain which then results in significant error in estimating JS divergence only from labeled source and target data. In essence, this result underscores the need for the development of an effective approach to utilize unlabeled target data for estimating the JS divergence, which involves techniques like pseudo-labeling.
\end{remark}

Therefore, we apply pseudo-labeling on unlabeled data to enrich labeled target data. Let $g$ be pseudo-label model and denote $\N(P_{t,k}, g) = \mathbb{E}\left[ \mathcal{D}_{JS} \left(  P_{t,k}(g(Z)) \parallel P_{t,k}(Y|Z) \right) \right]$ as the noise of $g$ with respect to the target domain conditioned on known labels. Then, the impact of pseudo-labeled data can be illustrated in a new bound for OSHeDA as follows.

\begin{theorem}\label{thm:6}
    Given a loss function $L$ satisfying Assumption~\ref{ass:1}, for any $0 < \delta < 1$, with probability at least $1 - \delta$, the following holds for all $h \in \mathcal{H}, f_s \in \mathcal{F}_s, f_t \in \mathcal{F}_t$:
\begin{align*}
     &\Er \left(P_t, h \circ f_t \right) \leq \lambda \widehat{\Er} \left(P_s, h \circ f_s \right) + \widehat{\Er} \left(P_t^u, h \circ f_t \right)  \nonumber \\ 
     &- \lambda \widehat{\Er} \left(P_s^u, h \circ f_s \right) + \sqrt{2} \lambda C \left( \left(\mathcal{D}_{JS}\left( P_s(Z) \parallel P_{t,k}(Z) \right) \right)^{\frac{1}{2}}  \right. \nonumber \\
     &\left.+ \left(\mathcal{D}_{JS} \left( P_s(Z,Y) \parallel P_{t,k}(Z,g(Z)) \right) \right)^{\frac{1}{2}} + \left(\N(P_{t,k}, g)\right)^{\frac{1}{2}}\right)  \nonumber \\
     & + \mathcal{O} \left( \lambda C  \sqrt{\frac{d_s \log{n_s} + d_s \log{|\mathcal{Y}^t| + \log{\frac{1}{\delta}}}}{n_s}}  \right. \nonumber \\
     &\left.+ C \sqrt{\frac{d_t \log{n_t} + d_t \log{|\mathcal{Y}^t| + \log{\frac{1}{\delta}}}}{n_t}} \right)
\end{align*}
where $\widehat{\Er}\left(P_s, h \circ f_s \right)$, $\widehat{\Er} \left(P_t^u, h \circ f_t \right)$, $\widehat{\Er} \left(P_s^u, h \circ f_s \right)$ are empirical errors calculated on samples from distributions $P_s$, $P_t^u$, $P_s^u$,  $n_t=n_{t_l}+n_{t_u}$, and $d_s$, $d_t$ are Natarajan dimension~\cite{natarajan1989learning} of hypothesis classes $\mathcal{H} \circ \mathcal{F}_s$, $\mathcal{H} \circ \mathcal{F}_t$.
\end{theorem}
Theorem~\ref{thm:6} shows that the error in the target domain depends on the quality of the pseudo-label model $g$, with higher-quality $g$ being more effective at reducing noise. Additionally, the bound emphasizes the importance of aligning the joint distributions between the source and target domains in OSHeDA. This makes OSHeDA more challenging compared to homogeneous DA (HoDA), where source and target data lie on the same space. In contrast, HoDA methods can attain good performance under certain conditions by solely aligning the marginal distributions of representations between source and target domains. We will illustrate this contrast through the bound for HoDA in Section~\ref{sec:hoda}.



\subsection{Learning bound for HoDA}\label{sec:hoda}

Before constructing the learning bound for HoDA, we introduce an assumption about the representation $Z$ as follows.

\begin{assumption}[Sufficient representation]\label{ass:2} Let $I_s(\cdot, \cdot)$ be the mutual information between two random variables in the source domain. We assume 
 $I_s(Z,Y) = I_s(X_s,Y)$. In particular, $I_s(Z,Y) = \mathcal{D}_{KL}\left( P_s(Z,Y) \parallel P_s(Z) \otimes P_s(Y) \right)$ and $I_s(X_s,Y) = \mathcal{D}_{KL}\left( P_s(X_s,Y) \parallel P_s(X_s) \otimes P_s(Y) \right)$ where $D_{KL}$ is KL divergence between two distributions.
\end{assumption}
Note that Assumption~\ref{ass:2} is reasonable because we have access to labeled data of source domain and the dimension of $\mathcal{Y}$ is often smaller than that of $\mathcal{Z}$. Based on this, we establish the learning bound in HoDA under the covariate shift below. 
\begin{proposition}\label{thm:2}
 Suppose Assumptions~\ref{ass:1} and~\ref{ass:2} hold and the distribution shift between source and target domains is covariate shift (i.e., $P_s(X) \neq P_t(X), P_s(Y|X) = P_t(Y|X)$), then for any $h \in \mathcal{H}$ and $f \in \mathcal{F}$, 
 we have:
\begin{align*}
\resizebox{0.48\textwidth}{!}{$
    \Er \left(P_t, h \circ f \right) \leq \Er \left(P_s, h \circ f \right) \nonumber + \sqrt{2} C \left( \mathcal{D}_{JS}\left(P_s(Z) \parallel P_t(Z) \right) \right)^{\frac{1}{2}}
$}
\end{align*}
\end{proposition}
In HoDA, due to the homogeneity of the input space, we can utilize a single representation mapping $f$ for both the source and target domains. Note that the bound in Proposition~\ref{thm:2} depends solely on the distance of the marginal distributions between the source and target domains, $\mathcal{D}_{JS}\left(P_s(Z) \parallel P_t(Z) \right)$, which can be effectively minimized even without access to labeled data in the target domain. Clearly, covariate shift assumption is only reasonable in HoDA, where the source and target data share the same feature and label spaces.
\section{Methodology}

\begin{figure*}[t]
    \centering
    \includegraphics[width=0.96\linewidth]{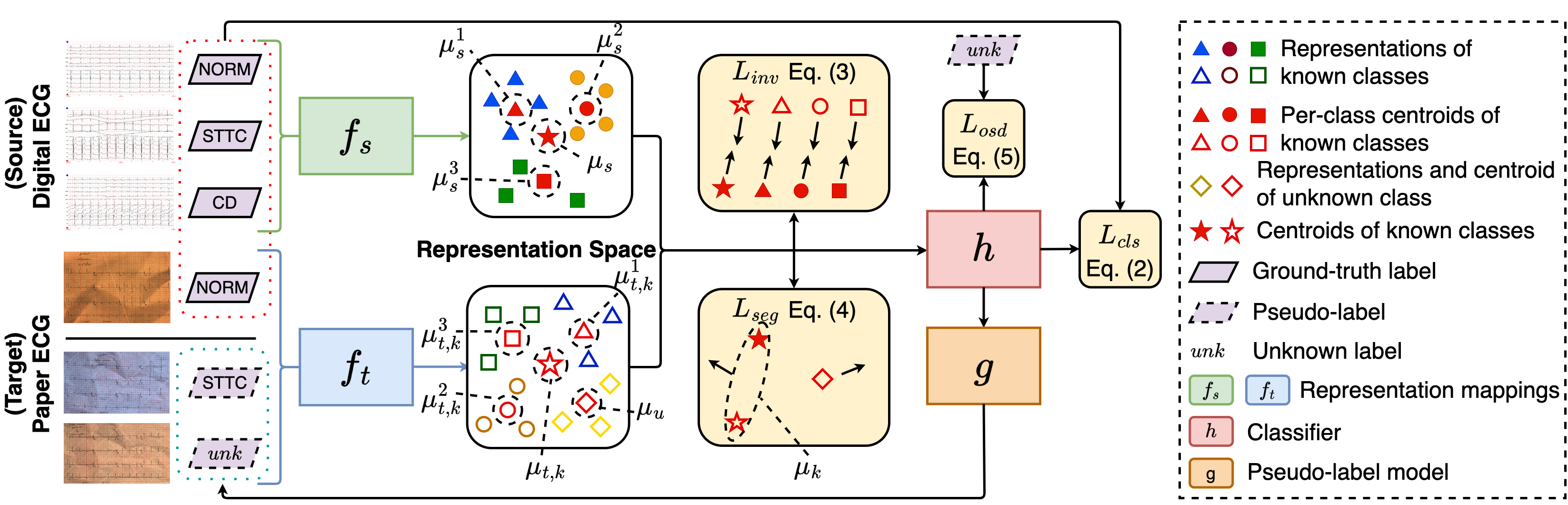}
    \caption{Overall architecture of RL-OSHeDA is illustrated with a motivating example from ECG-based diagnosis application. We leverage 2-stage learning process to update model parameters. In stage 1, model parameters are updated by optimizing $L_{cls}$. In stage 2, model parameters are updated by optimizing $L_{cls}$, $L_{inv}$, $L_{seg}$, and $L_{osd}$ with the help from pseudo-label model $g$.}
    \label{fig:2}
    \vspace{-0.3cm}
\end{figure*}

Motivated by theoretical results presented in Section~\ref{sec:theory}, we introduce RL-OSHeDA, a representation learning method specifically designed for OSHeDA. Our method aims to simultaneously optimize both the upper bound in Theorem~\ref{thm:6} and the lower bound in Proposition~\ref{thm:3}. RL-OSHeDA features two distinct representation mappings, $f_s$ and $f_t$, which map heterogeneous source and target feature spaces to a shared representation space, along with a classifier $h$ that makes predictions based on these representations. Figure~\ref{fig:2} presents the overall architecture of RL-OSHeDA, while pseudo code describing training process can be found in Appendix~\ref{app:training}. 

\subsection{Objective function}
To improve predictive performance in OSHeDA, our method targets the following: (i) minimizing prediction errors on both source and labeled target data, (ii) minimizing the distances of marginal and label-conditioned representation distributions for known classes between source and target data, (iii) maximizing the distances of marginal representation distributions between known and unknown classes, and (iv) minimizing the open-set difference. Specifically, RL-OSHeDA optimizes the following objective function:
\begin{align}
    L = L_{cls} + L_{inv} - L_{seg} + L_{osd}
\end{align}
where $L_{cls}$ is the classification error computed from source and labeled target datasets $D_s$ and $D_{t_l}$, defined as follows:
\begin{equation}
\resizebox{0.48\textwidth}{!}{$
 \displaystyle   L_{cls} =  \frac{\lambda}{n_s} \sum_{i=1}^{n_s} \CE\left( h \left( f_s \left( x_i^s \right) \right) , y_i^s \right) +  \frac{1}{n_{t_l}} \sum_{i=1}^{n_{t_l}} \CE\left( h \left( f_t \left( x_i^t \right) \right) , y_i^t \right)
$}
\end{equation}
\vspace{-0.7cm}

\noindent
Here $\CE$ is the cross-entropy loss. 

$L_{inv}$ denotes the distances of marginal and label-conditioned representation distributions for known classes between source and target datasets. Note that, we minimize the distance of the label-conditioned representation distribution $P(Z|Y)$, rather than the joint distribution $P(Z,Y)$, as noted by \citet{pham2023fairness}. As shown in Proposition~\ref{thm:5}, $L_{inv}$ can be defined based on JS divergence and minimized through adversarial learning. However, the number of discriminators required for this approach scales linearly with the number of classes, leading to instability in training when the dataset has a large number of classes. To address this issue, we implement $L_{inv}$ using maximum mean discrepancy (MMD) defined as follows:
\begin{align}
    L_{inv} = \left\| \mu_s - \mu_{t,k}  \right\|_2^2 + \sum_{m=1}^{\left|\mathcal{Y}^s\right|} \left\| \mu_s^m - \mu_{t,k}^m  \right\|_2^2
\end{align}
where $\mu_s$ (resp. $\mu_{t,k}$) is centroid of representations from source data (resp. target data belonging to known classes), and $\mu_s^m$ (resp. $\mu_{t,k}^m$) is centroid of representations from source data (resp. target data) belonging to known class $m$. Note that $\mu_{t,k}$ and $\mu_{t,k}^m$ are computed using both instances with ground-truth labels from labeled target data and those with high-quality pseudo-labels (see Section~\ref{sec:pl}) from unlabeled target data to provide a more accurate estimation. 

$L_{seg}$ is the distances between marginal representation distributions of known and unknown classes. Similarly, we implement $L_{seg}$ with MMD as follows:
\begin{align}
    L_{seg} = \left\| \mu_k - \mu_u  \right\|_2^2
\end{align}
where $\mu_k$ (resp. $\mu_u$) are centroids of representations from both source and target datasets belonging to ground-truth and pseudo known (resp. unknown) classes.

$L_{osd}$ represents the open-set difference, as detailed in Theorem~\ref{thm:6}. The optimal value for the open-set difference is $0$. However, due to the flexibility of deep neural networks, this term can become excessively negative during training and adversely affect model performance. To address this issue, we implement $L_{osd}$ as a non-negative risk estimator:
\begin{align}
\resizebox{0.48\textwidth}{!}{$\displaystyle
    L_{osd} = \max \left( 0, \frac{1}{n_t} \sum_{i=1}^{n_t} \CE\left( h \left( f_t \left( x_i^t \right) \right) , unk \right) - \frac{\lambda}{n_s} \sum_{i=1}^{n_s} \CE\left( h \left( f_s \left( x_i^s \right) \right) , unk \right)  \right)
$}
\end{align}
\vspace{-0.7cm}

\noindent
where $n_t = n_{t_l} + n_{t_u}$ is the size of the target dataset.

\begin{table*}[t]
\caption{Prediction performances ($HOS$, $OS^{\ast}$, $UNK$) of RL-OSHeDA and baselines for OSHeDA scenario on 7 datasets. We report average results over 10 random seeds for each dataset.}
\label{tab:1}
\centering
\resizebox{\textwidth}{!}{
\begin{tabular}{l|lll|lll|lll|lll}
\hline
     & \multicolumn{3}{c|}{CIFAR10 \& ILSVRC2012}                                      & \multicolumn{3}{c|}{ImageCLEF-DA}                                            & \multicolumn{3}{c|}{Multilingual Reuters Collection}                                                     & \multicolumn{3}{c}{NUSWIDE \& ImageNet}                                        \\\cline{2-13}
     & \multicolumn{1}{c}{$HOS$} & \multicolumn{1}{c}{$OS^{\ast}$} & \multicolumn{1}{c|}{$UNK$} & \multicolumn{1}{c}{$HOS$} & \multicolumn{1}{c}{$OS^{\ast}$} & \multicolumn{1}{c|}{$UNK$} & \multicolumn{1}{c}{$HOS$} & \multicolumn{1}{c}{$OS^{\ast}$} & \multicolumn{1}{c|}{$UNK$} & \multicolumn{1}{c}{$HOS$} & \multicolumn{1}{c}{$OS^{\ast}$} & \multicolumn{1}{c}{$UNK$} \\\hline
DS3L & 61.49$\pm$0.74                  & 59.04$\pm$1.00                   & 64.40$\pm$1.06                   & 58.62$\pm$2.04                   & 52.87$\pm$2.70                   & 66.74$\pm$2.77                   & 59.35$\pm$0.94                   & 52.92$\pm$1.27                   & 67.57$\pm$1.24                   & 67.61$\pm$1.65                   & 66.17$\pm$2.22                   & 69.20$\pm$2.30                   \\
KPG  & 57.27$\pm$0.50                    & 54.73$\pm$0.00                   & 60.30$\pm$1.11                    & 40.68$\pm$0.94                   & 34.60$\pm$0.00                   & 50.79$\pm$2.91                   & 11.27$\pm$0.07                   & 8.59$\pm$0.00                    & 17.04$\pm$0.96                   & 55.18$\pm$1.17                   & 52.60$\pm$0.00                   & 58.10$\pm$2.45                   \\
OPDA & 53.30$\pm$0.77                   & 48.22$\pm$1.00                   & 60.26$\pm$1.11                   & 53.17$\pm$1.83                   & 45.28$\pm$2.13                   & 65.76$\pm$2.76                   & 55.85$\pm$0.96                   & 48.47$\pm$1.23                   & 65.94$\pm$1.23                   & 71.06$\pm$1.44                   & 66.60$\pm$1.98                   & 76.38$\pm$2.09                   \\
PL        & 42.75$\pm$0.52                   & 37.12$\pm$0.49                   & 52.31$\pm$1.10                   & 39.20$\pm$1.62                   & 31.93$\pm$1.66                   & 54.34$\pm$2.91                   & 42.85$\pm$0.81                   & 34.56$\pm$1.00                   & 57.86$\pm$1.29                   & 42.43$\pm$0.26                   & 34.05$\pm$0.00                   & 61.15$\pm$2.10                   \\
SCT  & 59.61$\pm$0.75                   & 57.35$\pm$1.00                   & 62.33$\pm$1.08                   & 58.76$\pm$2.05                   & 53.09$\pm$2.71                   & 66.71$\pm$2.76                   & 61.17$\pm$0.94                   & \textbf{54.96$\pm$1.30}                   & 69.00$\pm$1.21                   & 70.42$\pm$1.49                   & 68.00$\pm$2.20                   & 73.10$\pm$1.99                   \\
SSAN & 60.38$\pm$0.73                   & 59.01$\pm$1.00                   & 62.01$\pm$1.08                   & 58.61$\pm$2.05                   & 53.18$\pm$2.74                   & 66.14$\pm$2.74                   & 58.25$\pm$0.93                   & 51.99$\pm$1.25                   & 66.26$\pm$1.24                   & 67.98$\pm$1.49                   & 66.25$\pm$2.04                   & 69.85$\pm$2.21                   \\
STN  & 61.59$\pm$0.72                   & 58.80$\pm$0.98                   & 64.87$\pm$1.05                   & 56.25$\pm$2.06                   & 49.80$\pm$2.69                   & 65.84$\pm$2.76                   & 59.21$\pm$0.96                   & 52.91$\pm$1.31                   & 67.24$\pm$1.23                   & 67.75$\pm$1.23                   & 64.80$\pm$1.42                   & 71.08$\pm$2.16                   \\
SL        & 60.74$\pm$0.74                   & 58.29$\pm$1.00                   & 63.67$\pm$1.08                   & 58.59$\pm$2.05                   & 52.84$\pm$2.70                   & 66.71$\pm$2.76                   & 58.53$\pm$0.96                   & 52.14$\pm$1.32                   & 66.74$\pm$1.21                   & 69.41$\pm$1.64                   & 66.63$\pm$2.26                   & 72.57$\pm$2.23                   \\
RL-OSHeDA & \textbf{72.33$\pm$0.70}                   & \textbf{67.88$\pm$0.98}                   & \textbf{77.81$\pm$0.97}                   & \textbf{63.98$\pm$2.04}                   & \textbf{56.63$\pm$2.72}                   & \textbf{74.80$\pm$2.51}                   & \textbf{65.39$\pm$0.91}                   & 54.47$\pm$1.21                   & \textbf{81.97$\pm$0.96}                   & \textbf{80.01$\pm$1.30}                   & \textbf{74.65$\pm$2.01}                   & \textbf{86.35$\pm$0.81}                   \\\hline
\end{tabular}}
\resizebox{\textwidth}{!}{
\begin{tabular}{l|lll|lll|lll|lll}
\hline
     & \multicolumn{3}{c|}{Office \& Caltech256}                                       & \multicolumn{3}{c|}{Wikipedia}                                               & \multicolumn{3}{c|}{PTB-XL}                                                  & \multicolumn{3}{c}{Average over datasets}                                                     \\\cline{2-13}
     & \multicolumn{1}{c}{$HOS$} & \multicolumn{1}{c}{$OS^{\ast}$} & \multicolumn{1}{c|}{$UNK$} & \multicolumn{1}{c}{$HOS$} & \multicolumn{1}{c}{$OS^{\ast}$} & \multicolumn{1}{c|}{$UNK$} & \multicolumn{1}{c}{$HOS$} & \multicolumn{1}{c}{$OS^{\ast}$} & \multicolumn{1}{c|}{$UNK$} & \multicolumn{1}{c}{$HOS$} & \multicolumn{1}{c}{$OS^{\ast}$} & \multicolumn{1}{c}{$UNK$} \\\hline
DS3L & 72.06$\pm$2.48                   & 67.41$\pm$3.68                   & 78.15$\pm$2.89                   & 56.00$\pm$2.01                   & 50.72$\pm$2.36                   & 66.24$\pm$2.80                   & 30.30$\pm$1.19		           & 34.95$\pm$0.58                   & 26.74$\pm$1.82                   & 57.92$\pm$1.58                   & 54.87$\pm$1.97                   & 62.72$\pm$2.12                   \\
KPG  & 34.46$\pm$0.99                   & 29.18$\pm$0.00                   & 45.34$\pm$3.58                   & 24.82$\pm$0.42                   & 16.40$\pm$0.00                   & 52.52$\pm$3.18                   & N/A                     & N/A                     &  N/A                    & 37.28$\pm$0.68                   & 32.68$\pm$0.00                   & 47.35$\pm$2.36                   \\
OPDA & 65.23$\pm$2.58                   & 57.70$\pm$3.43                   & 76.23$\pm$3.03                   & 52.66$\pm$1.92                   & 45.94$\pm$1.81                   & 65.42$\pm$3.12                   & 31.47$\pm$1.22                   & 36.35$\pm$0.63                   & 27.74$\pm$1.86                   & 54.67$\pm$1.53                   & 49.79$\pm$1.74                   & 62.53$\pm$2.17                   \\
PL        & 48.92$\pm$1.36                   & 40.38$\pm$1.13                   & 68.41$\pm$3.40                   & 41.87$\pm$1.65                   & 35.14$\pm$1.48                   & 58.40$\pm$3.10                   & 26.18$\pm$1.37                   & 36.43$\pm$0.55                   & 20.43$\pm$1.66                   & 40.60$\pm$1.08                   & 35.66$\pm$0.90                   & 53.27$\pm$2.22                   \\
SCT  & 75.72$\pm$2.14                   & 71.05$\pm$3.20                   & 81.79$\pm$2.54                   & 58.41$\pm$2.01                   & 52.86$\pm$2.39                   & 68.42$\pm$2.74                   & 26.23$\pm$1.65                   & \textbf{46.48$\pm$1.71}                   & 18.27$\pm$1.60                   & 59.89$\pm$1.58                   & 57.10$\pm$1.93                   & 64.49$\pm$1.99                   \\
SSAN & 72.95$\pm$2.36                   & 67.99$\pm$3.37                   & 79.67$\pm$2.88                   & 58.37$\pm$1.76                   & 52.76$\pm$1.95                   & 68.36$\pm$2.80                   & 25.16$\pm$1.47                   & 40.40$\pm$0.65                   & 18.27$\pm$1.54                   & 57.39$\pm$1.54                   & 55.94$\pm$1.86                   & 61.51$\pm$2.07                   \\
STN  & 72.26$\pm$2.28                   & 66.46$\pm$3.30                   & 79.84$\pm$2.75                   & 57.75$\pm$1.91                   & 51.40$\pm$2.13                   & 69.00$\pm$2.97                   & 27.08$\pm$0.96                   & 22.63$\pm$0.26                   & 33.72$\pm$1.65                   & 57.41$\pm$1.45                   & 52.40$\pm$1.73                   & 64.51$\pm$2.08                   \\
SL        & 72.14$\pm$2.54                   & 67.72$\pm$3.75                   & 77.89$\pm$2.96                   & 57.10$\pm$1.97                   & 51.60$\pm$2.19                   & 67.04$\pm$2.76                   & 25.74$\pm$1.55                   & 44.50$\pm$0.76                   & 18.11$\pm$1.52                   & 57.46$\pm$1.64                   & 56.24$\pm$2.00                   & 61.82$\pm$2.08                   \\
RL-OSHeDA & \textbf{78.18$\pm$2.05}                   & \textbf{73.04$\pm$2.91}                   & \textbf{85.25$\pm$2.50}                   & \textbf{63.10$\pm$1.89}                   & \textbf{57.26$\pm$2.45}                   & \textbf{73.04$\pm$2.37}                   & \textbf{47.48$\pm$1.25}                        & 44.30$\pm$1.39                        & \textbf{51.16$\pm$1.86}                        & \textbf{67.21$\pm$1.45}                   & \textbf{61.18$\pm$1.95}                   & \textbf{75.77$\pm$1.71}                   \\\hline      
\end{tabular}}
\vspace{-0.2cm}
\end{table*}

\subsection{Pseudo-labeling using 2-stage learning}\label{sec:pl}
The accuracy of $L_{inv}$ and $L_{seg}$ highly depends on the quality of pseudo-labels. Traditionally, the pseudo-label model $g$ is derived by modifying the classifier $h$ (e.g., using hard labels calculated from $h$'s outputs as pseudo-labels), which creates a coupling between $g$ and $h$. Specifically, $g$ is defined as $a \circ h$, where $a$ is an operator applied to the output of $h$ (e.g., $a := \arg\max$). When the distributions of the source and target domains are well-aligned, this coupling is harmless, as the optimal solution for $g$ also aligns with that for $h$. However, at the beginning of the training process, when the distributions of the source and target domains are not aligned, $g$ and $h$ have completely different objective functions, resulting in a trade-off between them. To address this issue, we propose a 2-stage learning approach as follows:
\begin{itemize}
    \item \textbf{Stage 1} ($epoch <T$): Update $f_s, f_t, h$ using $L_{cls}$.
    \item \textbf{Stage 2} ($epoch \geq T$): Update $f_s, f_t, h$ using $L$.
\end{itemize}
where $T$ is a threshold indicating when to switch from stage 1 to stage 2. In stage 1, optimizing $L_{cls}$ partially aligns the source and target domains, thereby reducing the trade-off between $g$ and $h$ during the optimization of $L$ in stage 2. Additionally, rather than simply using the hard labels with the largest logits from $h$ as the output of $g$, we propose generating pseudo-labels as follows:
\begin{itemize}
    \item First, select pseudo-labels as $g(x^t) = a'( h(f_t(x^t)))$ where $a'$ is $\arg\max$ operator applied to the logits of the known classes only.
    \item Then, select $1-\lambda$ fraction of instances with the smallest maximum logits and assign pseudo-labels $unk$ to them.
\end{itemize}
The motivation behind this design of the pseudo-label model $g$ is that, at the beginning of stage 2, there is no supervision signal for training the parameters of $h$ related to unknown class. Therefore, relying solely on logits to determine the unknown class is unreliable. Note that this strategy is only used to generate pseudo-labels during the training of stage 2. Once training is complete and $h$'s parameters for unknown class are well-trained by optimizing $L_{osd}$, we can simply use $\arg\max$ across all classes to generate predictions.

\section{Experiments}
Next, we empirically evaluate the performance of our methods across clinical, computer vision, and natural language processing applications. We focus on the OsHeDA scenario, characterized by heterogeneity in the feature space between the source and target domains, with the label space of the target domain encompassing both known and unknown classes. 

\subsection{Experimental setup}
\textbf{Datasets.} 
We conduct our experiments on 7 datasets including CIFAR10~\cite{krizhevsky2009learning} \& ILSVRC2012~\cite{russakovsky2015imagenet}; Wikipedia~\cite{rasiwasia2010new}; Multilingual Reuters Collection~\cite{amini2009learning}; NUSWIDE~\cite{chua2009nus} \& ImageNet~\cite{deng2009imagenet}; Office~\cite{saenko2010adapting} \& Caltech256~\cite{griffin2007caltech}; ImageCLEF-DA~\cite{griffin2007caltech}; PTB-XL~\cite{wagner2020ptb}. These datasets results in 56 DA tasks. Detailed descriptions and statistics of these datasets are provided in Appendix~\ref{app:dataset}. 

\noindent \textbf{Baselines.} 
We compare our method with several representative methods from \textbf{HeDA} (SSAN~\cite{li2020simultaneous}, STN~\cite{yao2019heterogeneous}, SCT~\cite{zhao2022semantic}, KPG~\cite{gu2022keypoint}), \textbf{OSDA} (OPDA~\cite{saito2018open}), and \textbf{OS-SSL} (DS3L~\cite{guo2020safe}) literature. For the HeDA methods, they are trained on both source and target data. In contrast, OSDA and OS-SSL methods are trained only on target data as they cannot handle heterogeneous feature spaces. During inference, HeDA and OS-SSL methods classify instances as $unk$ using the same method as our pseudo-label model $g$ (see Section~\ref{sec:pl}). Additionally, we explore supervised learning (SL) and pseudo-labeling (PL) methods trained on target data. Among all baselines, only KPG is designed to handle OSHeDA by combining Gromov-Wasserstein distance and partial optimal transport~\cite{xu2020joint}. Since $\lambda$ is a required input for most methods in our experiments, we utilize techniques from positive-unlabeled learning~\cite{zeiberg2020fast} to estimate $\lambda$. Detailed architectures of our model and the baselines are in Appendix~\ref{app:architect}. 

\noindent \textbf{Evaluation method.} 
We utilize $HOS$, the harmonic mean of $OS^{\ast}$ and $UNK$~\cite{bucci2020effectiveness}. $OS^{\ast}$ is the class-wise averaged accuracy of known classes, while $UNK$ measures the accuracy for the unknown class. $HOS$ is particularly suitable for OSHeDA because it emphasizes the ability to both correctly classify known classes and detect out-of-distribution instances simultaneously. In particular, this metric increases when the performance in both known and unknown classifications is high. 

\subsection{Experimental results}

\begin{figure}
    \centering
    \includegraphics[width=0.9\linewidth]{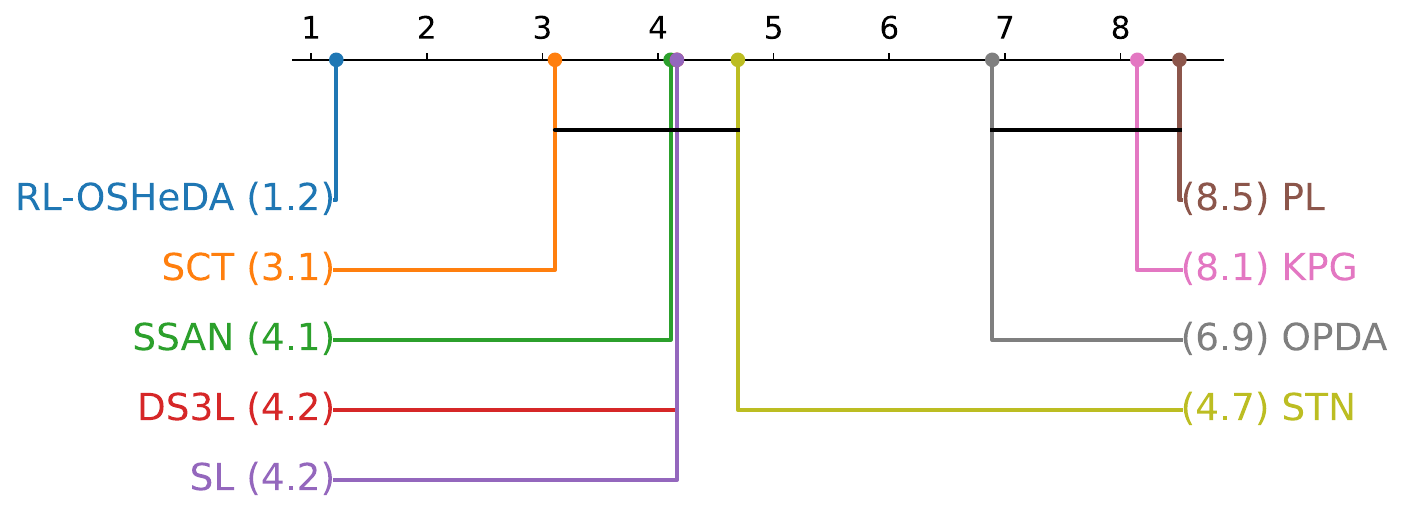}
    \caption{Critical Difference diagram for all methods calculated from 56 DA tasks. RL-OSHeDA is the highest ranked method on $HOS$ metric, and its performance is significantly better than baselines (as indicated by the lack of connections between RL-OSHeDA and baselines in the diagram).}
    \label{fig:3}
\end{figure}

\textbf{OSHeDA benchmark.} 
The prediction performance ($HOS$) of RL-OSHeDA and the baselines is summarized in Table~\ref{tab:1}. RL-OSHeDA consistently outperforms all baselines across all datasets, demonstrating its effectiveness in simultaneously addressing heterogeneity in the feature space and open-set in the label space during training. Among the baselines, KPG is specifically designed for OSHeDA by using optimal transport. Then, SVM trained on transported source and labeled target data is used to make prediction. However, this method underperforms in our evaluation due to its difficulty in correctly transporting from source to target data. Moreover, this method is not applicable for complex data structures, such as those found in PTB-XL dataset. Other baselines achieve better prediction performances, but their $HOS$ remains suboptimal due to their inability to handle novel classes or heterogeneous source data during training.



\begin{figure}[t]
    \centering
    \includegraphics[width=0.8\linewidth]{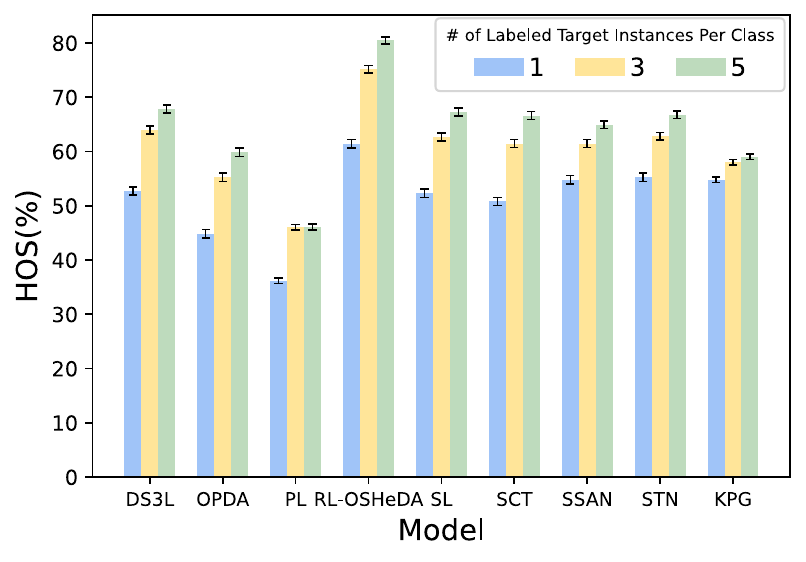}
    \vspace{-0.3cm}
    \caption{Performances w.r.t. different number of labeled target instances per class on CIFAR10 \& ILSVRC2012 dataset.} \vspace{-0.5cm}
    \label{fig:4}
\end{figure}

\begin{figure}[t]
    \centering
    \resizebox{\linewidth}{!}{
    \begin{subfigure}{.49\linewidth}
    \centering
    \includegraphics[width=\linewidth]{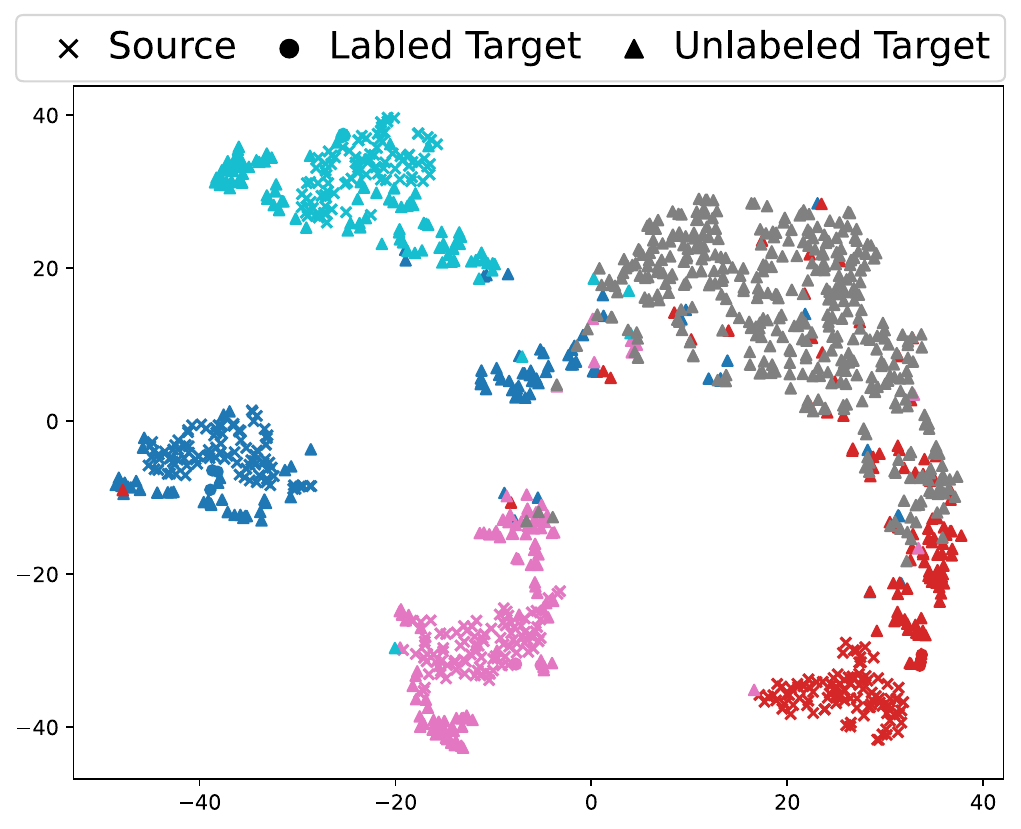}
    \caption{RL-OSHeDA}
    \end{subfigure}
    \begin{subfigure}{.49\linewidth}
    \centering
    \includegraphics[width=\linewidth]{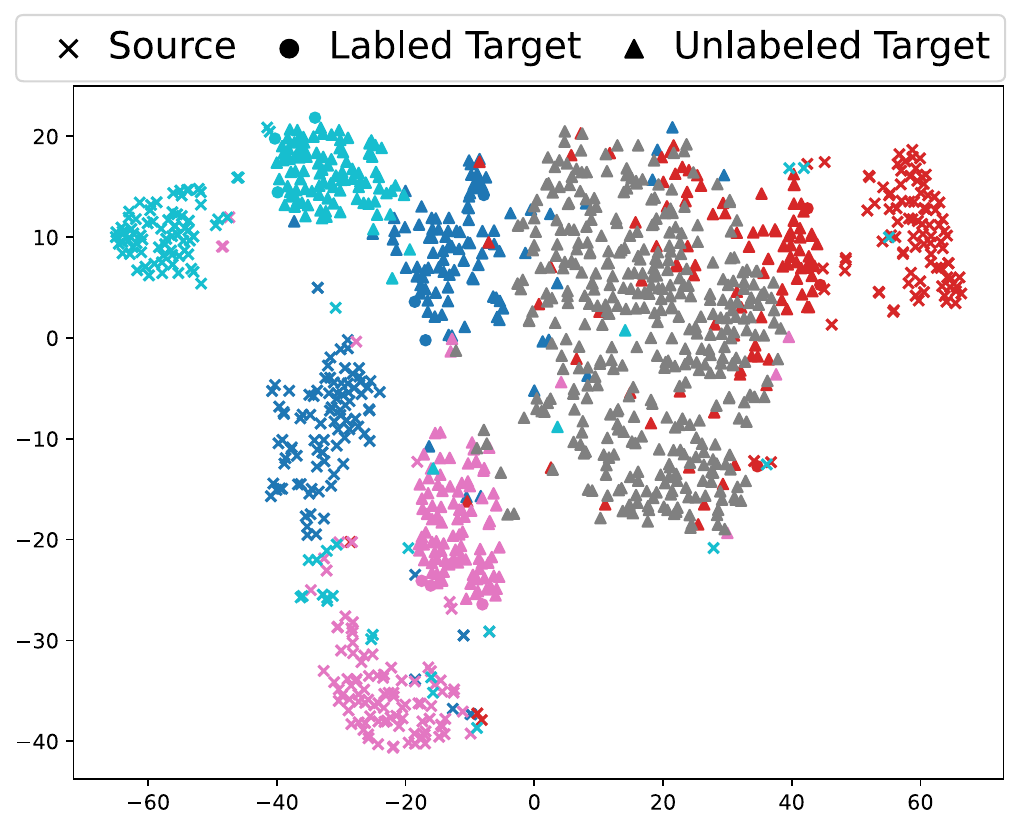}
    \caption{STN}
    \end{subfigure}
    }
    \vspace{-0.5cm}
    \caption{Visualization of representation spaces learned by RL-OSHeDA and STN for NUSWIDE \& ImageNet dataset. Different colors represent different classes, with the unknown class denoted in grey.}
    \label{fig:5}
\end{figure}

To further validate the superiority of RL-OSHeDA across all DA tasks, we conduct significance testing, including the Friedman test followed by the Nemenyi test~\cite{demvsar2006statistical}. The results (see Figure~\ref{fig:3}) show that our method significantly outperforms the baselines, with a P-value smaller than 0.05. Among all the baselines, SCT, SSAN, STN, SL, and DS3L exhibit better prediction performances than KPG, OPDA, and PL. Note that all methods, except OPDA and KPG, utilize our approach to detect the unknown class based on logits of known classes. This result suggests that while this approach can partially address the open-set issue, it cannot fully resolve it. For OPDA, although it is designed to handle open-set issue, its inability to leverage heterogeneous source data limits its performance to adapting with only a small labeled target dataset, resulting in suboptimal performance.

\noindent \textbf{Ablation study.} 
We conduct an ablation study to better understand the contribution of each component in the objective function of our method. As shown in Table~\ref{tab:2}, removing any component deteriorates model performance. This finding highlights the importance of achieving a good pseudo-label model using 2-stage learning approach as well as aligning the data distribution of known classes between source and target domains while simultaneously detecting and segregating unknown class from known ones for OSHeDA.

\begin{table}[t]
\caption{Ablation study for RL-OSHeDA on Multilingual Reuters Collection dataset. \textbf{Align} refers to 
using $L_{inv}$; \textbf{Segregate} refers to using $L_{seg}$; \textbf{OSD} refers to using $L_{osd}$; \textbf{2-stage} refers to using 2-stage learning approach.}
\label{tab:2}
\centering
\resizebox{\linewidth}{!}{
\begin{tabular}{cccc|lll}
\hline
Align & Segregate & OSD & 2-stage & \multicolumn{1}{c}{$HOS$} & \multicolumn{1}{c}{$OS^{\ast}$} & \multicolumn{1}{c}{$UNK$} \\\hline
\cmark    & \cmark      & \cmark              & \cmark  & \textbf{65.39}                   & \textbf{54.47}                   & \textbf{81.97}                   \\
\cmark    & \cmark      & \cmark              & \xmark  & 59.40                   & 49.47                   & 74.42                   \\
\cmark    & \xmark      & \cmark              & \cmark  & 61.92                   & 52.63                   & 75.37                   \\
\xmark    & \cmark      & \cmark              & \cmark  & 58.23                   & 51.13                   & 68.01                   \\
\cmark    & \cmark      & \xmark              & \cmark  & 59.96                   & 53.10                   & 68.97                   \\
\xmark    & \xmark      & \xmark              & \xmark  & 58.33                   & 51.86                   & 66.68    \\\hline              
\end{tabular}}
\vspace{-0.4cm}
\end{table}

\noindent \textbf{Impact of labeled target data.} 
We vary the number of instances per class in the labeled target data to investigate their impact on the DA process. Specifically, we conduct experiments on  CIFAR10 \& ILSVRC2012 dataset with 1, 3, and 5 instances per class in the labeled target data and visualize the result in Figure~\ref{fig:4}. Generally, we observe that increasing the number of labeled target instances facilitates better alignment and enhances the performance of all methods. This result demonstrates the importance of labeled target data for DA methods in OSHeDA.


\noindent \textbf{Visualization of representation space.} 
We perform a qualitative analysis to examine the learned representations of RL-OSHeDA and STN for NUSWIDE \& ImageNet dataset. Specifically, we use t-SNE~\cite{van2008visualizing} to project these representations into a 2-dimensional space. As shown in Figure~\ref{fig:5}, our method effectively aligns representations of the known classes between source and target domains while simultaneously segregating the representations of the unknown class (grey color). This results in improved $HOS$ scores compared to STN.
\section{Conclusion}
This paper studied a novel domain adaptation scenario called open-set heterogeneous domain adaptation (OSHeDA). We first conducted a theoretical analysis to establish learning bounds in OSHeDA. Based on these theorems, we proposed a 
representation learning method that aligns the data distribution of known classes between source and target domains while simultaneously detecting and segregating unknown class from known ones. The resulting models trained with the proposed method generalize well to target domains. Experiments on real datasets across diverse domains, including healthcare, natural language processing, and computer vision, demonstrate the effectiveness of our proposed method.

\section*{Acknowledgements}

\noindent This work was funded in part by the National Science Foundation under award number IIS-2145625 and by the National Institutes of Health under award number R01AI188576. We would like to express our sincere gratitude to Dr. Wei-Lun Chao from The Ohio State University for his valuable suggestions, which greatly improved this paper.

\bibliography{aaai25}

\begin{thebibliography}{64}
\providecommand{\natexlab}[1]{#1}

\bibitem[{Albuquerque et~al.(2019)Albuquerque, Monteiro, Darvishi, Falk, and Mitliagkas}]{albuquerque2019generalizing}
Albuquerque, I.; Monteiro, J.; Darvishi, M.; Falk, T.~H.; and Mitliagkas, I. 2019.
\newblock Generalizing to unseen domains via distribution matching.
\newblock \emph{arXiv preprint arXiv:1911.00804}.

\bibitem[{Amini, Usunier, and Goutte(2009)}]{amini2009learning}
Amini, M.~R.; Usunier, N.; and Goutte, C. 2009.
\newblock Learning from multiple partially observed views-an application to multilingual text categorization.
\newblock \emph{Advances in neural information processing systems}, 22.

\bibitem[{Ben-David et~al.(2010)Ben-David, Blitzer, Crammer, Kulesza, Pereira, and Vaughan}]{ben2010theory}
Ben-David, S.; Blitzer, J.; Crammer, K.; Kulesza, A.; Pereira, F.; and Vaughan, J.~W. 2010.
\newblock A theory of learning from different domains.
\newblock \emph{Machine learning}, 79: 151--175.

\bibitem[{Biau et~al.(2020)Biau, Cadre, Sangnier, and Tanielian}]{biau2020some}
Biau, G.; Cadre, B.; Sangnier, M.; and Tanielian, U. 2020.
\newblock {Some theoretical properties of GANS}.
\newblock \emph{The Annals of Statistics}, 48(3): 1539 -- 1566.

\bibitem[{Bucci, Loghmani, and Tommasi(2020)}]{bucci2020effectiveness}
Bucci, S.; Loghmani, M.~R.; and Tommasi, T. 2020.
\newblock On the effectiveness of image rotation for open set domain adaptation.
\newblock In \emph{European conference on computer vision}, 422--438. Springer.

\bibitem[{Chen et~al.(2020)Chen, Zhu, Li, and Gong}]{chen2020semi}
Chen, Y.; Zhu, X.; Li, W.; and Gong, S. 2020.
\newblock Semi-supervised learning under class distribution mismatch.
\newblock In \emph{Proceedings of the AAAI Conference on Artificial Intelligence}, volume~34, 3569--3576.

\bibitem[{Chua et~al.(2009)Chua, Tang, Hong, Li, Luo, and Zheng}]{chua2009nus}
Chua, T.-S.; Tang, J.; Hong, R.; Li, H.; Luo, Z.; and Zheng, Y. 2009.
\newblock Nus-wide: a real-world web image database from national university of singapore.
\newblock In \emph{Proceedings of the ACM international conference on image and video retrieval}, 1--9.

\bibitem[{Cortes and Mohri(2014)}]{cortes2014domain}
Cortes, C.; and Mohri, M. 2014.
\newblock Domain adaptation and sample bias correction theory and algorithm for regression.
\newblock \emph{Theoretical Computer Science}, 519: 103--126.

\bibitem[{Dem{\v{s}}ar(2006)}]{demvsar2006statistical}
Dem{\v{s}}ar, J. 2006.
\newblock Statistical comparisons of classifiers over multiple data sets.
\newblock \emph{The Journal of Machine learning research}, 7: 1--30.

\bibitem[{Deng et~al.(2009)Deng, Dong, Socher, Li, Li, and Fei-Fei}]{deng2009imagenet}
Deng, J.; Dong, W.; Socher, R.; Li, L.-J.; Li, K.; and Fei-Fei, L. 2009.
\newblock Imagenet: A large-scale hierarchical image database.
\newblock In \emph{2009 IEEE conference on computer vision and pattern recognition}, 248--255. Ieee.

\bibitem[{Donahue et~al.(2014)Donahue, Jia, Vinyals, Hoffman, Zhang, Tzeng, and Darrell}]{donahue2014decaf}
Donahue, J.; Jia, Y.; Vinyals, O.; Hoffman, J.; Zhang, N.; Tzeng, E.; and Darrell, T. 2014.
\newblock Decaf: A deep convolutional activation feature for generic visual recognition.
\newblock In \emph{International conference on machine learning}, 647--655. PMLR.

\bibitem[{Du et~al.(2023)Du, Zhao, Sheng, Li, and Chen}]{du2023semi}
Du, P.; Zhao, S.; Sheng, Z.; Li, C.; and Chen, H. 2023.
\newblock Semi-Supervised Learning via Weight-aware Distillation under Class Distribution Mismatch.
\newblock In \emph{Proceedings of the IEEE/CVF International Conference on Computer Vision}, 16410--16420.

\bibitem[{Fang et~al.(2020)Fang, Lu, Liu, Xuan, and Zhang}]{fang2020open}
Fang, Z.; Lu, J.; Liu, F.; Xuan, J.; and Zhang, G. 2020.
\newblock Open set domain adaptation: Theoretical bound and algorithm.
\newblock \emph{IEEE transactions on neural networks and learning systems}, 32(10): 4309--4322.

\bibitem[{Fang et~al.(2022)Fang, Lu, Liu, and Zhang}]{fang2022semi}
Fang, Z.; Lu, J.; Liu, F.; and Zhang, G. 2022.
\newblock Semi-supervised heterogeneous domain adaptation: Theory and algorithms.
\newblock \emph{IEEE Transactions on Pattern Analysis and Machine Intelligence}, 45(1): 1087--1105.

\bibitem[{Ganin et~al.(2016)Ganin, Ustinova, Ajakan, Germain, Larochelle, Laviolette, Marchand, and Lempitsky}]{ganin2016domain}
Ganin, Y.; Ustinova, E.; Ajakan, H.; Germain, P.; Larochelle, H.; Laviolette, F.; Marchand, M.; and Lempitsky, V. 2016.
\newblock Domain-adversarial training of neural networks.
\newblock \emph{The journal of machine learning research}, 17(1): 2096--2030.

\bibitem[{Goodfellow et~al.(2014)Goodfellow, Pouget-Abadie, Mirza, Xu, Warde-Farley, Ozair, Courville, and Bengio}]{goodfellow2014generative}
Goodfellow, I.; Pouget-Abadie, J.; Mirza, M.; Xu, B.; Warde-Farley, D.; Ozair, S.; Courville, A.; and Bengio, Y. 2014.
\newblock Generative adversarial nets.
\newblock \emph{Advances in neural information processing systems}, 27.

\bibitem[{Grief et~al.(2016)Grief, Patel, Kochendorfer, Green, Lussier, Li, Burton, and Boyd}]{grief2016simulation}
Grief, S.~N.; Patel, J.; Kochendorfer, K.~M.; Green, L.~A.; Lussier, Y.~A.; Li, J.; Burton, M.; and Boyd, A.~D. 2016.
\newblock Simulation of ICD-9 to ICD-10-CM transition for family medicine: simple or convoluted?
\newblock \emph{The Journal of the American Board of Family Medicine}, 29(1): 29--36.

\bibitem[{Griffin et~al.(2007)Griffin, Holub, Perona et~al.}]{griffin2007caltech}
Griffin, G.; Holub, A.; Perona, P.; et~al. 2007.
\newblock Caltech-256 object category dataset.
\newblock Technical report, Technical Report 7694, California Institute of Technology Pasadena.

\bibitem[{Gu et~al.(2022)Gu, Yang, Zeng, Sun, and Xu}]{gu2022keypoint}
Gu, X.; Yang, Y.; Zeng, W.; Sun, J.; and Xu, Z. 2022.
\newblock Keypoint-guided optimal transport with applications in heterogeneous domain adaptation.
\newblock \emph{Advances in Neural Information Processing Systems}, 35: 14972--14985.

\bibitem[{Guo et~al.(2020)Guo, Zhang, Jiang, Li, and Zhou}]{guo2020safe}
Guo, L.-Z.; Zhang, Z.-Y.; Jiang, Y.; Li, Y.-F.; and Zhou, Z.-H. 2020.
\newblock Safe deep semi-supervised learning for unseen-class unlabeled data.
\newblock In \emph{International conference on machine learning}, 3897--3906. PMLR.

\bibitem[{He et~al.(2022)He, Han, Lu, and Yin}]{he2022safe}
He, R.; Han, Z.; Lu, X.; and Yin, Y. 2022.
\newblock Safe-student for safe deep semi-supervised learning with unseen-class unlabeled data.
\newblock In \emph{Proceedings of the IEEE/CVF Conference on Computer Vision and Pattern Recognition}, 14585--14594.

\bibitem[{Hoffman et~al.(2013)Hoffman, Rodner, Donahue, Darrell, and Saenko}]{hoffman2013efficient}
Hoffman, J.; Rodner, E.; Donahue, J.; Darrell, T.; and Saenko, K. 2013.
\newblock Efficient learning of domain-invariant image representations.
\newblock \emph{arXiv preprint arXiv:1301.3224}.

\bibitem[{Hoffman et~al.(2014)Hoffman, Rodner, Donahue, Kulis, and Saenko}]{hoffman2014asymmetric}
Hoffman, J.; Rodner, E.; Donahue, J.; Kulis, B.; and Saenko, K. 2014.
\newblock Asymmetric and category invariant feature transformations for domain adaptation.
\newblock \emph{International journal of computer vision}, 109: 28--41.

\bibitem[{Huang et~al.(2021)Huang, Fang, Chen, Chai, Wei, Wei, Lin, and Li}]{huang2021trash}
Huang, J.; Fang, C.; Chen, W.; Chai, Z.; Wei, X.; Wei, P.; Lin, L.; and Li, G. 2021.
\newblock Trash to treasure: Harvesting ood data with cross-modal matching for open-set semi-supervised learning.
\newblock In \emph{Proceedings of the IEEE/CVF International Conference on Computer Vision}, 8310--8319.

\bibitem[{Huang, Yang, and Gong(2022)}]{huang2022they}
Huang, Z.; Yang, J.; and Gong, C. 2022.
\newblock They are not completely useless: Towards recycling transferable unlabeled data for class-mismatched semi-supervised learning.
\newblock \emph{IEEE Transactions on Multimedia}, 25: 1844--1857.

\bibitem[{Kiryo et~al.(2017)Kiryo, Niu, Du~Plessis, and Sugiyama}]{kiryo2017positive}
Kiryo, R.; Niu, G.; Du~Plessis, M.~C.; and Sugiyama, M. 2017.
\newblock Positive-unlabeled learning with non-negative risk estimator.
\newblock \emph{Advances in neural information processing systems}, 30.

\bibitem[{Kolesnikov et~al.(2019)Kolesnikov, Beyer, Zhai, Puigcerver, Yung, Gelly, and Houlsby}]{kolesnikov1912big}
Kolesnikov, A.; Beyer, L.; Zhai, X.; Puigcerver, J.; Yung, J.; Gelly, S.; and Houlsby, N. 2019.
\newblock Big transfer (BiT): General visual representation learning. arXiv 2020.
\newblock \emph{arXiv preprint arXiv:1912.11370}.

\bibitem[{Koltchinskii and Panchenko(2000)}]{koltchinskii2000rademacher}
Koltchinskii, V.; and Panchenko, D. 2000.
\newblock Rademacher processes and bounding the risk of function learning.
\newblock In \emph{High dimensional probability II}, 443--457. Springer.

\bibitem[{Krizhevsky(2009)}]{krizhevsky2009learning}
Krizhevsky, A. 2009.
\newblock Learning Multiple Layers of Features from Tiny Images.
\newblock \emph{Master's thesis, University of Tront}.

\bibitem[{LeCun, Bengio, and Hinton(2015)}]{lecun2015deep}
LeCun, Y.; Bengio, Y.; and Hinton, G. 2015.
\newblock Deep learning.
\newblock \emph{nature}, 521(7553): 436--444.

\bibitem[{Li et~al.(2021)Li, Kang, Zhu, Wei, and Yang}]{li2021domain}
Li, G.; Kang, G.; Zhu, Y.; Wei, Y.; and Yang, Y. 2021.
\newblock Domain consensus clustering for universal domain adaptation.
\newblock In \emph{Proceedings of the IEEE/CVF conference on computer vision and pattern recognition}, 9757--9766.

\bibitem[{Li et~al.(2018)Li, Lu, Huang, Zhu, and Shen}]{li2018heterogeneous}
Li, J.; Lu, K.; Huang, Z.; Zhu, L.; and Shen, H.~T. 2018.
\newblock Heterogeneous domain adaptation through progressive alignment.
\newblock \emph{IEEE transactions on neural networks and learning systems}, 30(5): 1381--1391.

\bibitem[{Li et~al.(2020)Li, Xie, Wu, Zhao, Liu, and Ding}]{li2020simultaneous}
Li, S.; Xie, B.; Wu, J.; Zhao, Y.; Liu, C.~H.; and Ding, Z. 2020.
\newblock Simultaneous semantic alignment network for heterogeneous domain adaptation.
\newblock In \emph{Proceedings of the 28th ACM international conference on multimedia}, 3866--3874.

\bibitem[{Li et~al.(2013)Li, Duan, Xu, and Tsang}]{li2013learning}
Li, W.; Duan, L.; Xu, D.; and Tsang, I.~W. 2013.
\newblock Learning with augmented features for supervised and semi-supervised heterogeneous domain adaptation.
\newblock \emph{IEEE Transactions on Pattern analysis and machine intelligence}, 36(6): 1134--1148.

\bibitem[{Liang(2016)}]{liang2016cs229t}
Liang, P. 2016.
\newblock CS229T/STAT231: Statistical learning theory (Winter 2016).

\bibitem[{Liu, Zhang, and Lu(2020)}]{liu2020heterogeneous}
Liu, F.; Zhang, G.; and Lu, J. 2020.
\newblock Heterogeneous domain adaptation: An unsupervised approach.
\newblock \emph{IEEE transactions on neural networks and learning systems}, 31(12): 5588--5602.

\bibitem[{Liu et~al.(2019)Liu, Cao, Long, Wang, and Yang}]{liu2019separate}
Liu, H.; Cao, Z.; Long, M.; Wang, J.; and Yang, Q. 2019.
\newblock Separate to adapt: Open set domain adaptation via progressive separation.
\newblock In \emph{Proceedings of the IEEE/CVF conference on computer vision and pattern recognition}, 2927--2936.

\bibitem[{Luo et~al.(2020)Luo, Wang, Huang, and Baktashmotlagh}]{luo2020progressive}
Luo, Y.; Wang, Z.; Huang, Z.; and Baktashmotlagh, M. 2020.
\newblock Progressive graph learning for open-set domain adaptation.
\newblock In \emph{International Conference on Machine Learning}, 6468--6478. PMLR.

\bibitem[{Mansour, Mohri, and Rostamizadeh(2009)}]{mansour2009domain}
Mansour, Y.; Mohri, M.; and Rostamizadeh, A. 2009.
\newblock Domain adaptation: Learning bounds and algorithms.
\newblock \emph{arXiv preprint arXiv:0902.3430}.

\bibitem[{Natarajan(1989)}]{natarajan1989learning}
Natarajan, B.~K. 1989.
\newblock On learning sets and functions.
\newblock \emph{Machine Learning}, 4: 67--97.

\bibitem[{Nguyen et~al.(2021)Nguyen, Tran, Gal, Torr, and Baydin}]{nguyen2021kl}
Nguyen, A.~T.; Tran, T.; Gal, Y.; Torr, P.; and Baydin, A.~G. 2021.
\newblock KL Guided Domain Adaptation.
\newblock In \emph{International Conference on Learning Representations}.

\bibitem[{Pham, Zhang, and Zhang(2023)}]{pham2023fairness}
Pham, T.-H.; Zhang, X.; and Zhang, P. 2023.
\newblock Fairness and Accuracy under Domain Generalization.
\newblock In \emph{The Eleventh International Conference on Learning Representations}.

\bibitem[{Pham, Zhang, and Zhang(2024)}]{phamnon}
Pham, T.-H.; Zhang, X.; and Zhang, P. 2024.
\newblock Non-stationary Domain Generalization: Theory and Algorithm.
\newblock In \emph{The 40th Conference on Uncertainty in Artificial Intelligence}.

\bibitem[{Qui{\~n}onero-Candela et~al.(2022)Qui{\~n}onero-Candela, Sugiyama, Schwaighofer, and Lawrence}]{quinonero2022dataset}
Qui{\~n}onero-Candela, J.; Sugiyama, M.; Schwaighofer, A.; and Lawrence, N.~D. 2022.
\newblock \emph{Dataset shift in machine learning}.
\newblock Mit Press.

\bibitem[{Rasiwasia et~al.(2010)Rasiwasia, Costa~Pereira, Coviello, Doyle, Lanckriet, Levy, and Vasconcelos}]{rasiwasia2010new}
Rasiwasia, N.; Costa~Pereira, J.; Coviello, E.; Doyle, G.; Lanckriet, G.~R.; Levy, R.; and Vasconcelos, N. 2010.
\newblock A new approach to cross-modal multimedia retrieval.
\newblock In \emph{Proceedings of the 18th ACM international conference on Multimedia}, 251--260.

\bibitem[{Russakovsky et~al.(2015)Russakovsky, Deng, Su, Krause, Satheesh, Ma, Huang, Karpathy, Khosla, Bernstein et~al.}]{russakovsky2015imagenet}
Russakovsky, O.; Deng, J.; Su, H.; Krause, J.; Satheesh, S.; Ma, S.; Huang, Z.; Karpathy, A.; Khosla, A.; Bernstein, M.; et~al. 2015.
\newblock Imagenet large scale visual recognition challenge.
\newblock \emph{International journal of computer vision}, 115: 211--252.

\bibitem[{Saenko et~al.(2010)Saenko, Kulis, Fritz, and Darrell}]{saenko2010adapting}
Saenko, K.; Kulis, B.; Fritz, M.; and Darrell, T. 2010.
\newblock Adapting visual category models to new domains.
\newblock In \emph{Computer Vision--ECCV 2010: 11th European Conference on Computer Vision, Heraklion, Crete, Greece, September 5-11, 2010, Proceedings, Part IV 11}, 213--226. Springer.

\bibitem[{Saito, Kim, and Saenko(2021)}]{saito2021openmatch}
Saito, K.; Kim, D.; and Saenko, K. 2021.
\newblock Openmatch: Open-set semi-supervised learning with open-set consistency regularization.
\newblock \emph{Advances in Neural Information Processing Systems}, 34: 25956--25967.

\bibitem[{Saito et~al.(2018)Saito, Yamamoto, Ushiku, and Harada}]{saito2018open}
Saito, K.; Yamamoto, S.; Ushiku, Y.; and Harada, T. 2018.
\newblock Open set domain adaptation by backpropagation.
\newblock In \emph{Proceedings of the European conference on computer vision (ECCV)}, 153--168.

\bibitem[{Shalev-Shwartz and Ben-David(2014)}]{shalev2014understanding}
Shalev-Shwartz, S.; and Ben-David, S. 2014.
\newblock \emph{Understanding machine learning: From theory to algorithms}.
\newblock Cambridge university press.

\bibitem[{Shen and Guo(2018)}]{shen2018unsupervised}
Shen, C.; and Guo, Y. 2018.
\newblock Unsupervised heterogeneous domain adaptation with sparse feature transformation.
\newblock In \emph{Asian conference on machine learning}, 375--390. PMLR.

\bibitem[{Simonyan and Zisserman(2014)}]{simonyan2014very}
Simonyan, K.; and Zisserman, A. 2014.
\newblock Very deep convolutional networks for large-scale image recognition.
\newblock \emph{arXiv preprint arXiv:1409.1556}.

\bibitem[{Van~der Maaten and Hinton(2008)}]{van2008visualizing}
Van~der Maaten, L.; and Hinton, G. 2008.
\newblock Visualizing data using t-SNE.
\newblock \emph{Journal of machine learning research}, 9(11).

\bibitem[{Wagner et~al.(2020)Wagner, Strodthoff, Bousseljot, Kreiseler, Lunze, Samek, and Schaeffter}]{wagner2020ptb}
Wagner, P.; Strodthoff, N.; Bousseljot, R.-D.; Kreiseler, D.; Lunze, F.~I.; Samek, W.; and Schaeffter, T. 2020.
\newblock PTB-XL, a large publicly available electrocardiography dataset.
\newblock \emph{Scientific data}, 7(1): 1--15.

\bibitem[{Wang et~al.(2023)Wang, Qiao, Liu, Song, Zheng, and Chen}]{wang2023out}
Wang, Y.; Qiao, P.; Liu, C.; Song, G.; Zheng, X.; and Chen, J. 2023.
\newblock Out-of-distributed semantic pruning for robust semi-supervised learning.
\newblock In \emph{Proceedings of the IEEE/CVF Conference on Computer Vision and Pattern Recognition}, 23849--23858.

\bibitem[{Xu et~al.(2020)Xu, Liu, Zhang, Cai, Wang, Liang, Ying, and Yin}]{xu2020joint}
Xu, R.; Liu, P.; Zhang, Y.; Cai, F.; Wang, J.; Liang, S.; Ying, H.; and Yin, J. 2020.
\newblock Joint Partial Optimal Transport for Open Set Domain Adaptation.
\newblock In \emph{IJCAI}, 2540--2546.

\bibitem[{Yao et~al.(2019)Yao, Zhang, Li, and Ye}]{yao2019heterogeneous}
Yao, Y.; Zhang, Y.; Li, X.; and Ye, Y. 2019.
\newblock Heterogeneous domain adaptation via soft transfer network.
\newblock In \emph{Proceedings of the 27th ACM MM}, 1578--1586.

\bibitem[{Yao et~al.(2020)Yao, Zhang, Li, and Ye}]{yao2020discriminative}
Yao, Y.; Zhang, Y.; Li, X.; and Ye, Y. 2020.
\newblock Discriminative distribution alignment: A unified framework for heterogeneous domain adaptation.
\newblock \emph{Pattern Recognition}, 101: 107165.

\bibitem[{Yu et~al.(2020)Yu, Ikami, Irie, and Aizawa}]{yu2020multi}
Yu, Q.; Ikami, D.; Irie, G.; and Aizawa, K. 2020.
\newblock Multi-task curriculum framework for open-set semi-supervised learning.
\newblock In \emph{Computer Vision--ECCV 2020: 16th European Conference, Glasgow, UK, August 23--28, 2020, Proceedings, Part XII 16}, 438--454. Springer.

\bibitem[{Zaheer et~al.(2020)Zaheer, Guruganesh, Dubey, Ainslie, Alberti, Ontanon, Pham, Ravula, Wang, Yang et~al.}]{zaheer2020big}
Zaheer, M.; Guruganesh, G.; Dubey, K.~A.; Ainslie, J.; Alberti, C.; Ontanon, S.; Pham, P.; Ravula, A.; Wang, Q.; Yang, L.; et~al. 2020.
\newblock Big bird: Transformers for longer sequences.
\newblock \emph{Advances in neural information processing systems}, 33: 17283--17297.

\bibitem[{Zeiberg, Jain, and Radivojac(2020)}]{zeiberg2020fast}
Zeiberg, D.; Jain, S.; and Radivojac, P. 2020.
\newblock Fast nonparametric estimation of class proportions in the positive-unlabeled classification setting.
\newblock In \emph{Proceedings of the AAAI Conference on Artificial Intelligence}, volume~34, 6729--6736.

\bibitem[{Zhao et~al.(2019)Zhao, Des~Combes, Zhang, and Gordon}]{zhao2019learning}
Zhao, H.; Des~Combes, R.~T.; Zhang, K.; and Gordon, G. 2019.
\newblock On learning invariant representations for domain adaptation.
\newblock In \emph{International conference on machine learning}, 7523--7532. PMLR.

\bibitem[{Zhao et~al.(2022)Zhao, Li, Zhang, Liu, Cao, Wang, and Tian}]{zhao2022semantic}
Zhao, Y.; Li, S.; Zhang, R.; Liu, C.~H.; Cao, W.; Wang, X.; and Tian, S. 2022.
\newblock Semantic correlation transfer for heterogeneous domain adaptation.
\newblock \emph{IEEE Transactions on Neural Networks and Learning Systems}.

\bibitem[{Zou et~al.(2018)Zou, Yu, Kumar, and Wang}]{zou2018unsupervised}
Zou, Y.; Yu, Z.; Kumar, B.; and Wang, J. 2018.
\newblock Unsupervised domain adaptation for semantic segmentation via class-balanced self-training.
\newblock In \emph{Proceedings of the European conference on computer vision (ECCV)}, 289--305.

\end{thebibliography}

\onecolumn

\appendix


\section{Proofs}\label{app:proof}

\subsection{Additional Lemmas}

\begin{lemma}\label{lemma:s1}
Given two domains $s$ and $t$ associated with two distributions $P_{s}(X,Y)$ and $P_{t}(X,Y)$, respectively, then for any classifier $h: \mathcal{X} \rightarrow \Delta\left(\mathcal{Y}\right)$, the expected  error of $h$ in domain $t$ can be upper bounded:
\begin{align*}
\Er \left( P_t, h \right) \leq  \Er \left( P_s, h \right) + \sqrt{2}C \times  \mathcal{D}_{JS}\left(P_t(X,Y) \parallel P_s(X,Y) \right)^{1/2}
\end{align*}
where $\mathcal{D}_{JS}\left( \cdot \parallel \cdot \right)$ is JS-divergence between two distributions.
\end{lemma}

\paragraph{Proof of Lemma~\ref{lemma:s1}}
Let $\mathcal{D}_{KL} \left( \cdot \parallel \cdot \right)$ be KL-divergence, $p_s$, $p_t$ are probability density functions associated with $P_s$, $P_t$, and $U = (X,Y)$ and $L(U) = L\left(h(X), Y\right)$. We first prove $\int_{\mathcal{E}} \left | p_{t}(u) - p_{s}(u)  \right | du = \frac{1}{2} \int \left | p_{t}(u) - p_{s}(u)  \right | du$ where $\mathcal{E}$ is the event that $p_{t}(u) \geq p_{s}(u)$ ($\ast$) as follows:

\begin{align*}
    \int_{\mathcal{E}} \left | p_{t}(u) - p_{s}(u)  \right | du &= \int_{\mathcal{E}} \left ( p_{t}(u) - p_{s}(u)  \right ) du \\
    &= \int_{\mathcal{E} \cup \overline{\mathcal{E}}} \left ( p_{t}(u) - p_{s}(u)  \right ) du - \int_{ \overline{\mathcal{E}}} \left ( p_{t}(u) - p_{s}(u)  \right ) du \\
    &\overset{(1)}{=}  \int_{ \overline{\mathcal{E}}} \left ( p_{s}(u) - p_{t}(u)  \right ) du \\
    &=  \int_{ \overline{\mathcal{E}}} \left | p_{t}(u) - p_{s}(u)  \right | du \\
    &= \frac{1}{2} \int \left | p_{t}(u) - p_{s}(u)  \right | du
\end{align*}

where $\overline{\mathcal{E}}$ is the complement of $\mathcal{E}$. We have $\overset{(1)}{=}$ because $\int_{\mathcal{E} \cup \overline{\mathcal{E}}} \left ( p_{t}(u) - p_{s}(u)  \right ) du = \int_{\mathcal{U}} \left ( p_{t}(u) - p_{s}(u)  \right ) du = 0$. Then, we have:

\begin{align}
\Er\left(P_t,h\right) &= \mathbb{E}_{P_t}\left[ L(U) \right] \nonumber \\
&= \int_{\mathcal{U}} L(u) p_{t}(u) du \nonumber \\
&= \int_{\mathcal{U}} L(u) p_{s}(u) du + \int_{\mathcal{U}} L(u) \left( p_{t}(u) - p_{s}(u) \right) du \nonumber \\
&= \mathbb{E}_{P_s}\left[ L(U) \right] + \int_{\mathcal{U}} L(u) \left( p_{t}(u) - p_{s}(u) \right) du \nonumber \\
&= \Er\left(P_s,h \right) + \int_{\mathcal{E}} L(u) \left( p_{t}(u) - p_{s}(u) \right) du + \int_{\overline{\mathcal{E}}} L(u) \left( p_{t}(u) - p_{s}(u) \right) du \nonumber \\
&\overset{(2)}{\leq} \Er\left(P_s,h \right) + \int_{\mathcal{E}} L(u) \left( p_{t}(u) - p_{s}(u) \right) du \nonumber \\
&\overset{(3)}{\leq} \Er\left(P_s,h \right) + C \int_{\mathcal{E}} \left( p_{t}(u) - p_{s}(u) \right) du \nonumber \\
&= \Er\left(P_s,h \right) + C \int_{\mathcal{E}} \left| p_{t}(u) - p_{s}(u) \right| du \nonumber \\
&\overset{(4)}{=} \Er\left(P_s,h \right) + \frac{C}{2} \int \left| p_{t}(u) - p_{s}(u) \right| du \nonumber \\
&\overset{(5)}{\leq} \Er\left(P_s,h \right) + \frac{C}{2} \sqrt{2  \min \left(\mathcal{D}_{KL}\left(P_{s}(U) \parallel P_{t}(U) \right) , \mathcal{D}_{KL}\left(P_{t}(U) \parallel P_{s}(U) \right) \right)} \nonumber \\
&\leq \Er\left(P_s,h \right) + \frac{C}{\sqrt{2}} \sqrt{ \mathcal{D}_{KL}\left(P_{t}(U) \parallel P_{s}(U) \right) } \label{eq:s0}
\end{align}
We have $\overset{(2)}{\leq}$ because $\int_{\overline{\mathcal{E}}} L(u) \left( p_{t}(u) - p_{s}(u) \right) du \leq 0$; $\overset{(3)}{\leq}$ because $L(u)$ is non-negative function and is bounded by $C$; $\overset{(4)}{=}$ by using ($\ast$); $\overset{(5)}{\leq}$ by using Pinsker’s inequality between total variation norm and KL-divergence. 

Let $P_{s,t}(U) = \frac{1}{2} \left ( P_{t}(U) + P_{s}(U) \right )$. Apply Eq.(\ref{eq:s0}) for two distributions $P_t$ and $P_{s,t}$, we have:
\begin{align}\label{eq:s1}
    \Er\left(P_t,h\right) \leq \Er\left(P_{s,t},h\right) + \frac{C}{\sqrt{2}}\sqrt{\mathcal{D}_{KL}\left( P_{t}(U) \parallel P_{s,t}(U) \right)}
\end{align}
Apply Eq.(\ref{eq:s0}) again for two distributions $P_{s,t}$ and $P_s$, we have:
\begin{align}\label{eq:s2}
    \Er\left(P_{s,t},h\right) \leq \Er\left(P_s,h \right) + \frac{C}{\sqrt{2}}\sqrt{\mathcal{D}_{KL}\left( P_{s}(U) \parallel P_{s,t}(U) \right)}
\end{align}
Adding Eq. (\ref{eq:s1}) to Eq. (\ref{eq:s2}) and subtracting $\Er\left(P_{s,t},h\right)$, we have:
\begin{align*}
    \Er\left(P_t,h\right) &\leq \Er\left(P_s,h \right) + \frac{C}{\sqrt{2}} \left ( \sqrt{\mathcal{D}_{KL}\left( P_{t}(U) \parallel P_{s,t}(U) \right)} + \sqrt{\mathcal{D}_{KL}\left( P_{s}(U) \parallel P_{s,t}(U) \right)} \right ) \\
    &\overset{(6)}{\leq} \Er\left(P_s,h \right) + \frac{C}{\sqrt{2}}  \sqrt{2 \left(\mathcal{D}_{KL}\left( P_{t}(U) \parallel P_{s,t}(U) \right) + \mathcal{D}_{KL}\left( P_{s}(U) \parallel P_{s,t}(U) \right) \right)} \\
    &= \Er\left(P_s,h \right) + \frac{C}{\sqrt{2}}  \sqrt{4 \mathcal{D}_{JS}\left( P_{s}(U) \parallel P_{t}(U) \right) } \\
    &= \Er\left(P_s,h \right) + \sqrt{2}C \sqrt{\mathcal{D}_{JS}\left( P_{s}(U) \parallel P_{t}(U) \right) } \\
\end{align*}
We have $\overset{(6)}{\leq}$ by using Cauchy–Schwarz inequality.

\begin{lemma} \label{lemma:s2}
Given two domains $s$ and $t$ associated with two distributions $P_{s}(X,Y)$ and $P_{t}(X,Y)$, respectively, then JS-divergence $\mathcal{D}_{JS}\left(P_s(X,Y) \parallel P_t(X,Y) \right)$ can be decomposed as follows:
\begin{align*}
    \mathcal{D}_{JS}\left(P_s(X,Y) \parallel P_t(X,Y) \right) &\leq \mathcal{D}_{JS}\left(P_s(Y) \parallel P_t(Y) \right) + \mathbb{E}_{P_s}\left[\mathcal{D}_{JS}\left(P_s(X|Y) \parallel P_t(X|Y) \right) \right] \\
    &+ \mathbb{E}_{P_t}\left[\mathcal{D}_{JS}\left(P_s(X|Y) \parallel P_t(X|Y) \right) \right]
\end{align*}
\end{lemma}

\paragraph{Proof of Lemma~\ref{lemma:s2}}

First, we show the decomposition formulation for KL-divergence as follow.

\begin{align}
    &\mathcal{D}_{KL}\left(P_s(X,Y) \parallel P_t(X,Y) \right) \nonumber \\
    &= \mathbb{E}_{P_s}\left[\log p_s(X,Y) - \log p_t(X,Y) \right] \nonumber \\
    &= \mathbb{E}_{P_s}\left [\log{p_s(Y)} + \log{p_s(X|Y)} \right] - \mathbb{E}_{p_s}\left[ \log{p_t(Y)} + \log{p_t(X|Y)} \right] \nonumber \\
    &= \mathbb{E}_{P_s}\left[\log{p_s(Y)} - \log{p_t(Y)}\right] + \mathbb{E}_{P_s}\left[\log{p_s(X|Y)} - \log{p_t(X|Y)}\right] \nonumber \\
    &= \mathbb{E}_{P_s}\left[\log{p_s(Y)} - \log{p_t(Y)}\right] + \mathbb{E}_{y \sim P_s(Y)}\left[ \mathbb{E}_{x \sim P_s(X|Y)}\left[\log{p_s(X|Y)} - \log{p_t(X|Y)} \right] \right] \nonumber \\
    &= \mathcal{D}_{KL}\left(P_s(Y) \parallel P_t(Y) \right) + \mathbb{E}_{P_s}\left[\mathcal{D}_{KL}\left(P_s(X|Y) \parallel P_t(X|Y) \right) \right] \label{eq:s3}
\end{align}

For JS-divergence, we have:
    
\begin{align*}
    &\mathcal{D}_{JS}\left(P_s(X,Y) \parallel P_t(X,Y) \right) \\
    &= \frac{1}{2} \left(\mathcal{D}_{KL}\left(P_s(X,Y) \parallel P_{s,t}(X,Y) \right) \right) + \frac{1}{2} \left( \mathcal{D}_{KL}\left(P_t(X,Y) \parallel P_{s,t}(X,Y) \right) \right) \\
    &\overset{(1)}{=} \frac{1}{2} \left( \mathcal{D}_{KL}\left(P_s(Y) \parallel P_{s,t}(Y) \right) \right) + \frac{1}{2} \left( \mathbb{E}_{P_s}\left[\mathcal{D}_{KL}\left(P_s(X|Y) \parallel P_{s,t}(X|Y) \right) \right] \right) \\
    &+ \frac{1}{2} \left(\mathcal{D}_{KL}\left(P_t(Y) \parallel P_{s,t}(Y) \right) \right) + \frac{1}{2} \left( \mathbb{E}_{P_t}\left[\mathcal{D}_{KL}\left(P_t(X|Y) \parallel P_{s,t}(X|Y) \right) \right] \right) \\
    &= \mathcal{D}_{JS}\left(P_s(Y) \parallel P_t(Y) \right) + \frac{1}{2} \left(   \mathbb{E}_{P_s}\left[\mathcal{D}_{KL}\left(P_s(X|Y) \parallel P_{s,t}(X|Y) \right) \right] \right) + \frac{1}{2} \left(   \mathbb{E}_{P_t}\left[\mathcal{D}_{KL}\left(P_t(X|Y) \parallel P_{s,t}(X|Y) \right) \right] \right) \\
    &\leq \mathcal{D}_{JS}\left(P_s(Y) \parallel P_t(Y) \right) + \frac{1}{2} \left(   \mathbb{E}_{P_s}\left[\mathcal{D}_{KL}\left(P_s(X|Y) \parallel P_{s,t}(X|Y) \right) \right] \right) + \frac{1}{2} \left(   \mathbb{E}_{P_s}\left[\mathcal{D}_{KL}\left(P_t(X|Y) \parallel P_{s,t}(X|Y) \right) \right] \right) \\ 
    &+ \frac{1}{2} \left(   \mathbb{E}_{P_t}\left[\mathcal{D}_{KL}\left(P_t(X|Y) \parallel P_{s,t}(X|Y) \right) \right] \right) + \frac{1}{2} \left(   \mathbb{E}_{P_t}\left[\mathcal{D}_{KL}\left(P_s(X|Y) \parallel P_{s,t}(X|Y) \right) \right] \right) \\
    &= \mathcal{D}_{JS}\left(P_s(Y) \parallel P_t(Y) \right) + \mathbb{E}_{P_s}\left[\mathcal{D}_{JS}\left(P_s(X|Y) \parallel P_t(X|Y) \right) \right] + \mathbb{E}_{P_t}\left[\mathcal{D}_{JS}\left(P_s(X|Y) \parallel P_t(X|Y) \right) \right]
\end{align*}

We have $\overset{(1)}{=}$ by applying Eq. (\ref{eq:s3}) for $\mathcal{D}_{KL}\left(P_s(X,Y) \parallel P_{s,t}(X,Y) \right)$ and $ \mathcal{D}_{KL}\left(P_t(X,Y) \parallel P_{s,t}(X,Y) \right)$.

\begin{lemma} \label{lemma:s3}
Given two domains $s$ and $t$ associated with two distributions $P_{s}(X,Y)$ and $P_{t}(X,Y)$, respectively, let $f: \mathcal{X} \rightarrow \mathcal{Z}$ be the representation mapping from input space $\mathcal{X}$ to representation space $\mathcal{Z}$. If the shift between domains $s$ and $t$ is covariate shift (i.e., $P_s(Y|X) = P_t(Y|X)$), and Assumption~\ref{ass:2} holds for the representation $Z$, then the shift between these two domains in representation space is also covariate shift (i.e., $P_s(Y|Z) = P_t(Y|Z)$).
\end{lemma}

\paragraph{Proof of Lemma~\ref{lemma:s3}}

We have:

\begin{align}
    \log p_s(y|x) &= \log \left( \int p_s(y,z|x) dz \right) \nonumber \\
    &= \log \left( \int p_s(y|z) p(z|x) dz \right) \nonumber \\
    &= \log \left( \mathbb{E}_{P(Z|x)} \left[ p_s(y|Z) \right] \right) \nonumber \\
    &\overset{(1)}{\geq} \mathbb{E}_{P(Z|x)} \left[ \log p_s(y|Z) \right] \label{eq:s4}
\end{align}

We have $\overset{(1)}{\geq}$ by using Jensen's inequality. Taking expectation w.r.t. $P_t(X,Y)$ over both sides, we have:

\begin{align}
    &\mathbb{E}_{P_t(X,Y)} \left[ \log p_s(Y|X) - \mathbb{E}_{P(Z|X)} \left[ \log p_s(Y|Z) \right] \right]  \nonumber \\
    &= \int \int \left( \log p_s(y|x) - \mathbb{E}_{P(Z|x)} \left[ \log p_s(Y|Z) \right] \right) p_t(x,y) dx dy \nonumber \\
    &= \int \int \left( \log p_s(y|x) - \mathbb{E}_{P(Z|x)} \left[ \log p_s(Y|Z) \right] \right) p_s(x,y) \frac{p_t(x,y)}{p_s(x,y)} dx dy \nonumber \\
    &=\mathbb{E}_{P_s(X,Y)} \left[ \left( \log p_s(Y|X) - \mathbb{E}_{P(Z|X)} \left[ \log p_s(Y|Z) \right] \right) \frac{p_t(X,Y)}{p_s(X,Y)} \right]  \nonumber \\
    &\overset{(1)}{\leq} \left( \max_{x,y}\frac{p_t(x,y)}{p_s(x,y)} \right) \mathbb{E}_{P_s(X,Y)} \left[  \log p_s(Y|X) - \mathbb{E}_{P(Z|X)} \left[ \log p_s(Y|Z) \right] \right]  \nonumber \\
    &= \left( \max_{x,y}\frac{p_t(x,y)}{p_s(x,y)} \right) \left( \mathbb{E}_{P_s(X,Y)} \left[ \log p_s(Y|X) \right] - \mathbb{E}_{P_s(Z,Y)} \left[   \log p_s(Y|Z) \right] \right)  \nonumber \\
    &= \left( \max_{x,y}\frac{p_t(x,y)}{p_s(x,y)} \right) \left( H_s(Y,X) - H_s(Y,Z) \right) \nonumber \\
    &= \left( \max_{x,y}\frac{p_t(x,y)}{p_s(x,y)} \right) \left( \left( H_s(Y) - H_s(Y,Z) \right) - \left( H_s(Y) - H_s(Y,X) \right) \right) \nonumber \\
    &= \left( \max_{x,y}\frac{p_t(x,y)}{p_s(x,y)} \right) \left( I_s(Y,Z) - I_s(Y,X) \right) \nonumber \\
    &\overset{(2)}{=} 0 \label{eq:s5}
\end{align}

We have $\overset{(1)}{\leq}$ because $\log p_s(y|x) - \mathbb{E}_{P(Z|x)} \left[ \log p_s(y|z) \right] \geq 0$ according to Eq. (\ref{eq:s4}); $\overset{(2)}{=}$ because $I_s(Y,Z) = I_s(Y,X)$ according to Assumption~\ref{ass:2}. Based on Eq. (\ref{eq:s5}), we have:

\begin{align}
    \mathbb{E}_{P_t(X,Y)} \left[ \log p_s(Y|X) \right]  &=\mathbb{E}_{P_t(X,Y)} \left[ \mathbb{E}_{P(Z|X)} \left[ \log p_s(Y|Z) \right] \right] \nonumber \\
    &=\mathbb{E}_{P_t(Y,Z)} \left[ \log p_s(Y|Z) \right] \label{eq:s6}
\end{align}

We also have:

\begin{align}
    \mathbb{E}_{P_t(Y,Z)} \left[ \log P_t(Y|Z) \right] &= - H_t(Y|Z) \nonumber \\
    &= I_t(Y,Z) - H_t(Y) \nonumber \\
    &\overset{(1)}{\leq} I_t(Y,X) - H_t(Y) \nonumber \\
    &= - H_t(Y|X) \nonumber \\
    &= \mathbb{E}_{P_t(X,Y)} \left[ \log P_t(Y|X) \right] \label{eq:s7}
\end{align}

We have $\overset{(1)}{\leq}$ by using data processing inequality. Finally, we have:

\begin{align}
    & \;\;\;\;\; \mathbb{E}_{P_t(Z)} \left[ \mathcal{D}_{KL} \left( P_t(Y|Z) \parallel P_s(Y|Z) \right) \right] \nonumber \\
    &\overset{(1)}{=} \mathbb{E}_{P_t(Z)} \left[ \mathcal{D}_{KL} \left( P_t(Y|Z) \parallel P_s(Y|Z) \right) \right] - \mathbb{E}_{P_t(X)} \left[ \mathcal{D}_{KL} \left( P_t(Y|X) \parallel P_s(Y|X) \right) \right] \nonumber \\
    &= \mathbb{E}_{P_t(Y,Z)} \left[ \log p_t(Y|Z) - \log p_s(Y|Z) \right] - \mathbb{E}_{P_t(X,Y)} \left[ \log p_t(Y|X) - \log p_s(Y|X) \right] \nonumber \\
    &= \left( \mathbb{E}_{P_t(Y,Z)} \left[ \log p_t(Y|Z) \right] - \mathbb{E}_{P_t(X,Y)} \left[ \log p_t(Y|X) \right] \right) + \left( \mathbb{E}_{P_t(X,Y)} \left[ \log p_s(Y|X) \right] - \mathbb{E}_{P_t(Y,Z)} \left[ \log p_s(Y|Z) \right] \right) \nonumber \\
    &\overset{(2)}{=} 0 \label{eq:s8}
\end{align}

We have $\overset{(1)}{=}$ because the shift between two domains w.r.t. input space $\mathcal{X}$ is covariate shift; $\overset{(2)}{=}$ by using Eq. (\ref{eq:s6}) and Eq. (\ref{eq:s7}) and the fact that KL-divergence is non-negative. Note that Eq. (\ref{eq:s8}) implies that the shift between these two domains w.r.t. representation space $\mathcal{Z}$ is also covariate shift (i.e., $P_s(Y|Z) = P_t(Y|Z)$).

\begin{lemma}\label{lemma:s4}
Given domain $d$ associated with a distribution $P_d(X,Y)$, then for any $\delta > 0$, with probability at least $1 - \delta$ over sample $S$ of size $n$ drawn i.i.d from domain $d$,  for all $h \in \mathcal{H}: \mathcal{X} \rightarrow \Delta\left(\mathcal{Y}\right)$, the expected error of $h$ in domain $d$ can be upper bounded:
\begin{align*}
    \Er \left( P_d, h \right) \leq  \widehat{\Er} \left( P_d, h \right) + 2\mathcal{R}_S\left( \mathcal{L} \circ \mathcal{H} \right) + 3C\sqrt{\frac{\log(2 / \delta)}{2n}}
\end{align*}
\end{lemma}
where $\mathcal{L} \circ \mathcal{H} = \left\{ (x,y) \rightarrow L\left(h\left(x\right),y\right): h \in \mathcal{H} \right\}$ and $\mathcal{R}_S\left( \mathcal{L} \circ \mathcal{H} \right)$ is an empirical Rademacher complexity of the function class $\mathcal{L} \circ \mathcal{H}$ computed from the sample $S$.

\paragraph{Proof of Lemma~\ref{lemma:s4}}
We start from the Rademacher bound~\cite{koltchinskii2000rademacher} which is stated as follows.
\paragraph{Rademacher Bounds.}
  Let $\mathcal{F}$ be a family of functions mapping from $Z$ to $[0,1]$. Then, for any $0 < \delta < 1$, with probability at least $1 - \delta$ over sample $S = \{z_1,\cdots,z_n\}$, the following holds for all $f \in \mathcal{F}$:
\begin{align*}
    \mathbb{E}\left[ f^{Z} \right] \leq \frac{1}{n} \sum_{i=1}^{n} f(z_i) + 2 \mathcal{R}_S(\mathcal{F}) + 3\sqrt{\frac{\log (2 / \delta)}{2n}}
\end{align*}
where $\mathcal{R}_S \left(\mathcal{F}\right)$ is an empirical Rademacher complexity of function class $\mathcal{F}$ computed from the sample $S$.

We then apply this result to our setting with $Z = (X,Y)$, the loss function $L$ bounded by $C$, and the function class $\mathcal{L} \circ \mathcal{H} = \left\{ (x,y) \rightarrow L\left(h\left(x\right),y\right): h \in \mathcal{H} \right\}$. In particular, we scale the loss function $L$ to $[0,1]$ by dividing by C and denote the new class of scaled loss functions as $\mathcal{L} \circ \mathcal{H} / C$. Then, for any $\delta > 0$, with probability at least $1 - \delta$, we have:
\begin{align}
   \frac{\Er \left( P_d, h \right)}{C} &\leq \frac{\widehat{\Er} \left( P_d, h \right)}{C} + 2 \mathcal{R}_S\left(\mathcal{L} \circ \mathcal{H} / C\right) + 3\sqrt{\frac{\log (2 / \delta)}{2n}} \nonumber \\
   &\overset{(1)}{=} \frac{\widehat{\Er} \left( P_d, h \right)}{C} + \frac{2}{C}\mathcal{R}_S\left(\mathcal{L} \circ \mathcal{H}\right) + 3\sqrt{\frac{\log (2 / \delta)}{2n}} \label{eq:s9}
\end{align}
We have $\overset{(1)}{=}$ by using the property of empirical Redamacher complexity that $\mathcal{R}_S(\alpha\mathcal{F}) = \alpha \mathcal{R}_S(\mathcal{F})$. We derive Lemma~\ref{lemma:s4} by multiplying Eq. (\ref{eq:s9}) by C.

\begin{lemma}\label{lemma:s5}
Given a loss function $L$ satisfied Assumption~\ref{ass:1} and a sample $S = \left\{ (x_1, y_1), \cdots, (x_n, y_n) \right\}$ of size $n$, then an empirical Rademacher complexity $\mathcal{R}_S\left( \mathcal{L} \circ \mathcal{H} \right)$ computed from the sample $S$ is upper bounded as follows. 
\begin{align*}
    \mathcal{R}_S\left(\mathcal{L} \circ \mathcal{H}\right) \leq C \sqrt{\frac{2d \log n + 4d \log  \left|\mathcal{Y} \right|  }{n}}
\end{align*}
\end{lemma}
where $d$ is Natarajan dimension of hypothesis class $\mathcal{H}$.

\paragraph{Proof of Lemma~\ref{lemma:s5}}

We have:
\begin{align*}
    \mathcal{R}_S\left(\mathcal{L} \circ \mathcal{H}\right) &\overset{(1)}{\leq} C \sqrt{\frac{2 \log \left| \mathcal{L} \circ \mathcal{H} \right|}{n}} \\
    &\overset{(2)}{\leq} C \sqrt{\frac{2d \log n + 4d \log  \left|\mathcal{Y} \right|  }{n}}
\end{align*}

We have $\overset{(1)}{\leq}$ by applying Massart's finite lemma (Lemma 5 in~\citet{liang2016cs229t}) for function class $\mathcal{L} \circ \mathcal{H}$ and note that $\sup_{h \in \mathcal{H}} \frac{1}{n} \sum_{i=1}^n L(h(x_i), y_i)^2 \leq C^2$ because of Assumption~\ref{ass:2}; $\overset{(2)}{\leq}$ by using the fact that $\left| \mathcal{L} \circ \mathcal{H}\right| \leq \left| \mathcal{H}\right|$ and then applying Natarajan lemma (Lemma 29.4 in~\citet{shalev2014understanding}) for hypothesis class $\mathcal{H}$ with Natarajan dimension $d$.


\subsection{Proof of main theorems}

\paragraph{Proof of Theorem~\ref{thm:1}}

We have:

\begin{align}
    \Er\left( P_t, h \circ f_t \right) &= \mathbb{E}_{P_t(Y,Z)} \left[ L(h(Z), Y) \right] \nonumber \\
    &= \int \int L(h(z), y) p_{t}(y,z) dydz \nonumber \\
    &= \int \int L(h(z), y) \left( p_{t}(y,z|y \in \mathcal{Y}^s) p_{t}(y \in \mathcal{Y}^s) + p_{t}(y,z|y \notin \mathcal{Y}^s) p_{t}(y \notin \mathcal{Y}^s) \right) dydz \nonumber \\
    &= \lambda \mathbb{E}_{P_{t,k}(Y,Z)} \left[ L(h(Z), Y) \right] + (1 - \lambda) \mathbb{E}_{P_{t,u}(Y,Z)} \left[ L(h(Z), Y) \right] \nonumber \\
    &= \lambda \Er\left( P_{t,k}, h \circ f_t \right) + (1 - \lambda) \Er\left( P_{t,u}, h \circ f_t \right) \label{eq:s11}
\end{align}

Applying Lemma~\ref{lemma:s1} for the two distributions $P_{t,k}(Z,Y)$ and $P_s(Z,Y)$, we have:

\begin{align}
    \Er\left( P_{t,k}, h \circ f_t \right) \leq \Er\left( P_s, h \circ f_s \right) + \sqrt{2}C \left( \mathcal{D}_{JS} \left( P_s(Y,Z), P_{t,k}(Y,Z) \right) \right)^{1/2} \label{eq:s12}
\end{align}

Let $P_{t,k}^u$ is the distribution induced from $P_{t,k}$ by the mapping $f_t^u$. Then, we have:

\begin{align}
    \Er\left( P_t^u, h \circ f_t \right) &= \mathbb{E}_{P_t^u(Y,Z)} \left[ L(h(Z), Y) \right] \nonumber \\
    &= \int \int L(h(z), y) p^u_{t}(y,z) dydz \nonumber \\
    &\overset{(1)}{=} \int \int L(h(z), y) \left( p_{t,k}^u(y,z) p_{t}(y \in \mathcal{Y}^s) + p_{t,u}(y,z) p_{t}(y \notin \mathcal{Y}^s) \right) dydz \nonumber \\
    &= \lambda \mathbb{E}_{P_{t,k}^u(Y,Z)} \left[ L(h(Z), Y) \right] + (1 - \lambda) \mathbb{E}_{P_{t,u}(Y,Z)} \left[ L(h(Z), Y) \right] \nonumber \\
    &= \lambda \Er\left( P_{t,k}^u, h \circ f_t \right) + (1 - \lambda) \Er\left( P_{t,u}, h \circ f_t \right) \nonumber \\
    &\overset{(2)}{\geq} \lambda \left( \Er\left( P_{s}^u, h \circ f_s \right) - \sqrt{2}C \left( \mathcal{D}_{JS} \left( P_{t,k}^u (Y,Z) \parallel P_{s}^u (Y,Z) \right) \right)^{1/2} \right)  + (1 - \lambda) \Er\left( P_{t,u}, h \circ f_t \right) \nonumber \\ 
    &\overset{(3)}{\geq} \lambda \left( \Er\left( P_{s}^u, h \circ f_s \right) - \sqrt{2}C \left(\mathcal{D}_{JS} \left( P_{t,k}^u (Z) \parallel P_{s}^u (Z) \right)  + \mathbb{E}_{P^u_{t,k}(Z)} \left[ P^u_{t,k}(Y|Z) \parallel P_{s}^u (Y|Z) \right] \right. \right. \nonumber \\
    &+ \left. \left. \mathbb{E}_{P^u_{s}(Z)} \left[ P^u_{t,k}(Y|Z) \parallel P_{s}^u (Y|Z) \right] \right)^{1/2} \right)  + (1 - \lambda) \Er\left( P_{t,u}, h \circ f_t \right) \nonumber \\
    &\overset{(4)}{=} \lambda \Er\left( P_{s}^u, h \circ f_s \right) - \sqrt{2}\lambda C \left(\mathcal{D}_{JS} \left( P_{t,k}^u (Z) \parallel P_{s}^u (Z) \right) \right)^{1/2}  + (1 - \lambda) \Er\left( P_{t,u}, h \circ f_t \right)
    \label{eq:s13}
\end{align}

We have $\overset{(1)}{=}$ by using the fact that $P_t(Y,Z) = \lambda P_{t,k}(Y,Z) + (1- \lambda) P_{t,u}(Y,Z)$ and $f_t^u$ is the mapping such that $f_t^u\left(X^t, Y \right) = (X^t, unk)$; $\overset{(2)}{\geq}$ by using Lemma~\ref{lemma:s1}; $\overset{(3)}{\geq}$ by using Lemma~\ref{lemma:s2}; $\overset{(4)}{=}$ because $P^u_{t,k}(Y|Z) \parallel P_{s}^u (Y|Z)$ (support of $Y = \left\{ unk \right\}$). Finally, by combining Eq. (\ref{eq:s11}), Eq. (\ref{eq:s12}), and Eq. (\ref{eq:s13}), we have:

\begin{align*}
     \Er \left(P_t, h \circ f_t \right) &\leq \underbrace{\lambda \Er \left(P_s, h \circ f_s \right)}_{\textbf{source error}}  + \underbrace{\Er \left(P_t^u, h \circ f_t \right) - \lambda \Er \left(P_s^u, h \circ f_s \right)}_{\textbf{open-set difference}} \nonumber \\ 
     &+ \underbrace{\sqrt{2} \lambda C \left( \left(\mathcal{D}_{JS}\left( P_s(Z) \parallel P_{t,k}(Z) \right) \right)^{\frac{1}{2}}  
     + \left(\mathcal{D}_{JS} \left( P_s(Z,Y) \parallel P_{t,k}(Z,Y) \right) \right)^{\frac{1}{2}} \right)}_{\textbf{domain distance}} 
\end{align*}

\paragraph{Proof of Proposition~\ref{thm:3}}

We have:

\begin{align}
    \Er\left( P_t, h \circ f_t \right) &= \mathbb{E}_{P_t(Y,Z)} \left[ L(h(Z), Y) \right] \nonumber \\
    &= \int \int L(h(z), y) p_{t}(y,z) dydz \nonumber \\
    &= \int \int L(h(z), y) \left( p_{t}(y,z|y \in \mathcal{Y}^s) p_{t}(y \in \mathcal{Y}^s) + p_{t}(y,z|y \notin \mathcal{Y}^s) p_{t}(y \notin \mathcal{Y}^s) \right) dydz \nonumber \\
    &= \lambda \mathbb{E}_{P_{t,k}(Y,Z)} \left[ L(h(Z), Y) \right] + (1 - \lambda) \mathbb{E}_{P_{t,u}(Y,Z)} \left[ L(h(Z), Y) \right] \nonumber \\
    &= \lambda \Er\left( P_{t,k}, h \circ f_t \right) + (1 - \lambda) \Er\left( P_{t,u}, h \circ f_t \right) \label{eq:s14}
\end{align}

Next, applying Lemma~\ref{lemma:s1} for the two distributions $P_{t,u}(Y,Z)$ and $P_s^u(Y,Z)$, we have:

\begin{align}
    \Er \left( P_s^u, h \circ f_s \right) &\leq \Er \left( P_{t,u}, h \circ f_t \right) + \sqrt{2}C \left(\mathcal{D}_{JS} \left( P_s^u (Y,Z) \parallel P_{t,u} (Y,Z) \right) \right)^{1/2} \nonumber \\
    &\overset{(1)}{\leq} \Er \left( P_{t,u}, h \circ f_t \right) + \sqrt{2}C \left( \mathcal{D}_{JS} \left( P_s^u (Z) \parallel P_{t,u} (Z) \right) \right. \nonumber \\
    &+ \left. \mathbb{E}_{P_s^u (Z)} \left[ \mathcal{D}_{JS} \left( P_s^u (Y|Z) \parallel P_{t,u} (Y|Z) \right) \right] + \mathbb{E}_{P_{t,u} (Z)} \left[ \mathcal{D}_{JS} \left( P_s^u (Y|Z) \parallel P_{t,u} (Y|Z) \right) \right] \right)^{1/2} \nonumber \\
    &\overset{(2)}{=} \Er \left( P_{t,u}, h \circ f_t \right) + \sqrt{2}C \left( \mathcal{D}_{JS} \left( P_s (Z) \parallel P_{t,u} (Z) \right) \right)^{1/2} \label{eq:s15}
\end{align}

We have $\overset{(1)}{\leq}$ by applying Lemma~\ref{lemma:s2}; $\overset{(2)}{=}$ because $P_s^u (Y|Z) = P_{t,u} (Y|Z)$ (support of $Y = \{unk\}$) and $P^u_s (Z) = P_s (Z)$. Combining Eq. (\ref{eq:s14}) and Eq. (\ref{eq:s15}), we have:

\begin{align*}
    \Er\left( P_t, h \circ f_t \right) \geq \lambda \Er\left( P_{t,u}, h \circ f_t \right) + (1 - \lambda) \Er\left( P_s^u, h \circ f_s \right) - \sqrt{2}(1 - \lambda)C \left( \mathcal{D}_{JS} \left( P_s (Z) \parallel P_{t,u} (Z) \right) \right)^{1/2}
\end{align*}

\paragraph{Proof of Proposition~\ref{thm:5}}

Before giving the proof, we first introduce following definition about adversarial learning for invariant representation.

\begin{definition}\label{def:1} Given dataset $D_s=\{x^s_i\}_{i=1}^{n_s}$ and $D_t=\{x^t_i\}_{i=1}^{n_t}$ associated with the distributions $P_s(X_s)$ and $P_t(X_t)$, respectively, the goal of adversarial learning approach for invariant representation is to achieve $\widehat{L}_{adv} = \inf_{\alpha,\beta} \sup_{\gamma} \left( \frac{1}{n_s} \sum_{i=1}^{n_s} \log \left ( D_{\gamma}(F_{\alpha}(x^s_i)) \right ) \right.$ $\left.+ \frac{1}{n_t} \sum_{i=1}^{n_t} \log \left ( 1 - D_{\gamma}(F_{\beta}(x^t_i)) \right ) \right) $ where $F_\alpha$, $F_\beta$ are the mappings from the feature spaces $\mathcal{X}^s, \mathcal{X}^t$ to the representation space $\mathcal{Z}$ parameterized by $\alpha \in \mathcal{A}$ and $\beta \in \mathcal{B}$, and $D_{\gamma}$ are the discriminator parameterized by $\gamma \in \Gamma$.
\end{definition}

Then Proposition~\ref{thm:5} are formally state as follows.

\paragraph{Proposition 2 (Formal).} Let $\alpha^{\ast}, \beta^{\ast}, \gamma^{\ast}$be the parameters learned by optimizing $L_{adv}$ and $\widehat{\alpha}, \widehat{\beta}, \widehat{\gamma}$ be the parameters learned by optimizing $\widehat{L}_{adv}$. We have:
\begin{align*}
    &\mathbb{E} \left[\mathcal{D}_{JS}\left(P_{\widehat{\alpha}}(Z) \parallel P_{\widehat{\beta}}(Z) \right) \right] \leq \mathcal{D}_{JS}\left(P_{\alpha^{\ast}}(Z) \parallel P_{\beta^{\ast}}(Z) \right) \nonumber \\ 
    &+ \mathcal{O} \left( \left(\frac{1}{\sqrt{n_s}} + \frac{1}{\sqrt{n_t}} \right) \times C(\mathcal{A}, \mathcal{B}, \Gamma) \right)
\end{align*}
where $C(\mathcal{A}, \mathcal{B}, \Gamma)$ is a constant specified by the parameter spaces $\mathcal{A}, \mathcal{B}, \Gamma$, and $L_{adv} = \inf_{\alpha,\beta} \sup_{\gamma} \int_{\mathcal{X}^s} \log \left ( D_{\gamma}(F_{\alpha}(x^s)) \right ) p_s(x^s) dx^s + \int_{\mathcal{X}^t} \log \left ( 1 - D_{\gamma}(F_{\beta}(x^t)) \right ) p_t(x^t) dx^t$ is the objective function of adversarial learning for invariant representation with infinite data.

The proof for Proposition~\ref{thm:5} is based on the proof provided for GAN model by~\citet{biau2020some}. Let $L(\alpha, \beta, \gamma) = \int_{\mathcal{Z}} \left(\log \left ( D_{\gamma}(z) \right ) p_{\alpha}(z) + \log \left ( 1 - D_{\gamma}(z) \right ) p_{\beta}(z)\right) dz $, we have:
\begin{align*}
    2 \mathcal{D}_{JS} \left( P_{\widehat{\alpha}}(Z) \parallel P_{\widehat{\beta}}(Z) \right) &= L(\widehat{\alpha}, \widehat{\beta}, \widehat{\gamma}) + \log(4) \\
    &\leq \sup_{\gamma} L(\widehat{\alpha}, \widehat{\beta}, \gamma) + \log(4) \\
    &\leq \sup_{\gamma} \left(  \widehat{L}(\widehat{\alpha}, \widehat{\beta}, \gamma) + \left| \widehat{L}(\widehat{\alpha}, \widehat{\beta}, \gamma) - L(\widehat{\alpha}, \widehat{\beta}, \gamma) \right| \right) + \log(4) \\
    &\leq \sup_{\gamma}   \widehat{L}(\widehat{\alpha}, \widehat{\beta}, \gamma) + \sup_{\gamma}  \left| \widehat{L}(\widehat{\alpha}, \widehat{\beta}, \gamma) - L(\widehat{\alpha}, \widehat{\beta}, \gamma) \right|  + \log(4) \\
    &\leq \inf_{\alpha,\beta}\sup_{\gamma}   \widehat{L}(\alpha, \beta, \gamma) + \sup_{\alpha,\beta,\gamma}  \left| \widehat{L}(\alpha, \beta, \gamma) - L(\alpha, \beta, \gamma) \right|  + \log(4) \\
    &\leq \inf_{\alpha,\beta}\sup_{\gamma}   L(\alpha, \beta, \gamma) + \left| \inf_{\alpha,\beta}\sup_{\gamma}   \widehat{L}(\alpha, \beta, \gamma) - \inf_{\alpha,\beta}\sup_{\gamma}   L(\alpha, \beta, \gamma) \right| \\
    &+ \sup_{\alpha,\beta,\gamma}  \left| \widehat{L}(\alpha, \beta, \gamma) - L(\alpha, \beta, \gamma) \right|  + \log(4) \\
    &\overset{(1)}{\leq} \inf_{\alpha,\beta}\sup_{\gamma}   L(\alpha, \beta, \gamma) + \sup_{\alpha,\beta}\left| \sup_{\gamma}   \widehat{L}(\alpha, \beta, \gamma) - \sup_{\gamma}   L(\alpha, \beta, \gamma) \right| \\
    &+ \sup_{\alpha,\beta,\gamma}  \left| \widehat{L}(\alpha, \beta, \gamma) - L(\alpha, \beta, \gamma) \right|  + \log(4) \\
    &\overset{(2)}{\leq} \inf_{\alpha,\beta}\sup_{\gamma}   L(\alpha, \beta, \gamma) + 2 \sup_{\alpha,\beta,\gamma}  \left| \widehat{L}(\alpha, \beta, \gamma) - L(\alpha, \beta, \gamma) \right|  + \log(4) \\
    &= 2 \mathcal{D}_{JS}\left(P_{\alpha^{\ast}}(Z) \parallel P_{\beta^{\ast}}(Z) \right) + 2 \sup_{\alpha,\beta,\gamma}  \left| \widehat{L}(\alpha, \beta, \gamma) - L(\alpha, \beta, \gamma) \right| 
\end{align*}
We have $\overset{(1)}{\leq}$ by using inequality $|\inf A - \inf B| \leq \sup |A-B|$, $\overset{(2)}{\leq}$ by using inequality $|\sup A - \sup B| \leq \sup |A-B|$.
Take the expectation and rearrange the both sides, we have:
\begin{align*}
    &\mathbb{E}\left[\mathcal{D}_{JS} \left( P_{\widehat{\alpha}}(Z) \parallel P_{\widehat{\beta}}(Z) \right) \right] - \mathcal{D}_{JS}\left(P_{\alpha^{\ast}}(Z) \parallel P_{\beta^{\ast}}(Z) \right) \\
    &\leq \mathbb{E}\left[\sup_{\alpha,\beta,\gamma}  \left| \widehat{L}(\alpha, \beta, \gamma) - L(\alpha, \beta, \gamma) \right| \right] \\
    &= \mathbb{E}\left[\sup_{\alpha,\beta,\gamma}  \left| \frac{1}{n_s} \sum_{i=1}^{n_s} \log \left ( D_{\gamma}((z^s_i)) \right ) + \frac{1}{n_t} \sum_{i=1}^{n_t} \log \left ( 1 - D_{\gamma}(z^t_i) \right ) \right. \right. \\
    &\left. \left. - \int_{\mathcal{Z}} \left(\log \left ( D_{\gamma}(z) \right ) p_{\alpha}(z) + \log \left ( 1 - D_{\gamma}(z) \right ) p_{\beta}(z)\right) dz \right| \right] \\
    &\leq \mathbb{E}\left[\sup_{\alpha,\beta,\gamma}  \left| \underbrace{\frac{1}{n_s} \sum_{i=1}^{n_s} \log \left ( D_{\gamma}((z^s_i)) \right ) - \int_{\mathcal{Z}} \left(\log \left ( D_{\gamma}(z) \right ) p_{\alpha}(z) \right) dz}_{A_s(\alpha, \beta, \gamma)} \right| \right] \\
    &+ \mathbb{E}\left[\sup_{\alpha,\beta,\gamma}  \left| \underbrace{\frac{1}{n_t} \sum_{i=1}^{n_t} \log \left ( 1 - D_{\gamma}(z^t_i) \right ) - \int_{\mathcal{Z}} \left(\log \left ( 1 - D_{\gamma}(z) \right ) p_{\beta}(z) \right) dz}_{A_t(\alpha, \beta, \gamma)} \right| \right] \\
\end{align*}
Note that $\left(A_s\left(\alpha, \beta, \gamma \right)\right)_{\alpha \in \mathcal{A}, \beta \in \mathcal{B}, \gamma \in \Gamma}$ and $\left(A_t\left(\alpha, \beta, \gamma \right)\right)_{\alpha \in \mathcal{A}, \beta \in \mathcal{B}, \gamma \in \Gamma}$ are the subgaussian processes in the metric spaces $\left(\mathcal{A} \times \mathcal{B} \times \Gamma, C_1 \left \| \cdot \right \| / \sqrt{n_s}\right)$ and $\left(\mathcal{A} \times \mathcal{B} \times \Gamma, C_1 \left \| \cdot \right \| / \sqrt{n_t}\right)$ where $C_1$  is a constant and $\left \| \cdot \right \|$ is the Euclidean norm on $\mathcal{A} \times \mathcal{B} \times \Gamma$. Then using Dudley's entropy integral, we have:
\begin{align*}
    &\mathbb{E}\left[\mathcal{D}_{JS} \left( P_{\widehat{\alpha}}(Z) \parallel P_{\widehat{\beta}}(Z) \right) \right] - \mathcal{D}_{JS}\left(P_{\alpha^{\ast}}(Z) \parallel P_{\beta^{\ast}}(Z) \right) \\
    &\leq  \mathbb{E}\left[\sup_{\alpha,\beta,\gamma} A_s\left(\alpha, \beta, \gamma \right) \left| \right| \right] + \mathbb{E}\left[\sup_{\alpha,\beta,\gamma} A_t\left(\alpha, \beta, \gamma \right) \left| \right| \right] \\
    &\leq 12 \int_{0}^{\infty} \left(\sqrt{\log N(\mathcal{A} \times \mathcal{B} \times \Gamma,  C \left \| \cdot \right \| / \sqrt{n_s}, \epsilon)} + \sqrt{\log N(\mathcal{A} \times \mathcal{B} \times \Gamma,  C \left \| \cdot \right \| / \sqrt{n_t}, \epsilon)} \right) d\epsilon \\
    &= 12C_1 \left(\frac{1}{\sqrt{n_s}} + \frac{1}{\sqrt{n_t}}\right) \int_{0}^{\infty}\sqrt{\log N(\mathcal{A} \times \mathcal{B} \times \Gamma, \left \| \cdot \right \|, \epsilon)} d\epsilon \\
    &\overset{(3)}{=} 12C_1 \left(\frac{1}{\sqrt{n_s}} + \frac{1}{\sqrt{n_t}}\right) \int_{0}^{\diam(\mathcal{A} \times \mathcal{B} \times \Gamma)}\sqrt{\log N(\mathcal{A} \times \mathcal{B} \times \Gamma, \left \| \cdot \right \|, \epsilon)} d\epsilon \\
    &\overset{(4)}{\leq} 12C_1 \left(\frac{1}{\sqrt{n_s}} + \frac{1}{\sqrt{n_t}}\right) \int_{0}^{\diam(\mathcal{A} \times \mathcal{B} \times \Gamma)}\sqrt{\log \left( \left( \frac{2C_2 \sqrt{\dim(\mathcal{A} \times \mathcal{B} \times \Gamma)}}{\epsilon} \right) ^{\dim(\mathcal{A} \times \mathcal{B} \times \Gamma)} \right)} d\epsilon \\
    &= \mathcal{O}\left( \left(\ \frac{1}{\sqrt{n_s}} + \frac{1}{\sqrt{n_t}} \right) \times C(\mathcal{A}, \mathcal{B}, \Gamma)\right) \\
\end{align*}
where $\diam(\cdot)$ and $\dim(\cdot)$ are the diameter and the dimension of the metric space, and $C(\mathcal{A}, \mathcal{B}, \Gamma)$ is the function of $\diam(\mathcal{A} \times \mathcal{B} \times \Gamma)$ and $\dim(\mathcal{A} \times \mathcal{B} \times \Gamma)$. We have $\overset{(3)}{=}$ because $N(\mathcal{A} \times \mathcal{B} \times \Gamma, \left \| \cdot \right \|, \epsilon) = 1$ for $\epsilon > \diam(\mathcal{A} \times \mathcal{B} \times \Gamma)$, $\overset{(4)}{\leq}$ by using inequality $N(\mathcal{T}, \| \cdot \|, \epsilon) \leq \left(\frac{2C_2\sqrt{d}}{\epsilon}\right)^{d}$ where $\mathcal{T}$ lied in Euclidean space $\mathbb{R}^d$ is the set of vectors whose length is at most $C_2$.

\paragraph{Proof of Theorem~\ref{thm:6}}
We have:
\begin{align}
    \mathcal{D}_{JS} \left( P_s(Z,Y) \parallel P_{t,k}(Z,Y) \right) &\overset{(1)}{\leq} \mathcal{D}_{JS} \left( P_s(Z,Y) \parallel P_{t,k}(Z, g(Z)) \right) + \mathcal{D}_{JS} \left( P_{t,k}(Z, g(Z)) \parallel P_{t,k}(Z, Y) \right) \nonumber \\
    &\overset{(2)}{\leq} \mathcal{D}_{JS} \left( P_s(Z,Y) \parallel P_{t,k}(Z, g(Z)) \right) + \mathcal{D}_{JS} \left( P_{t,k}(Z) \parallel P_{t,k}(Z) \right) \nonumber \\
    &+ \mathbb{E}_{P_{t,k}(Z)} \left[ \mathcal{D}_{JS} \left( P_{t,k}(g(Z)|Z) \parallel P_{t,k}(Y|Z) \right) \right] \nonumber \\
    &= \mathcal{D}_{JS} \left( P_s(Z,Y) \parallel P_{t,k}(Z, g(Z)) \right) + \N\left(P_{t,k},g\right) \label{eq:s10}
\end{align}

We have $\overset{(1)}{\leq}$ by using triangle inequality for JS-divergence; $\overset{(2)}{\leq}$ by applying Lemma~\ref{lemma:s2}. According to Theorem~\ref{thm:1}, we have:

\begin{align}
     \Er \left(P_t, h \circ f_t \right) &\leq \lambda \Er \left(P_s, h \circ f_s \right)  + \Er \left(P_t^u, h \circ f_t \right) - \lambda \Er \left(P_s^u, h \circ f_s \right) \nonumber \\ 
     &+ \sqrt{2} \lambda C \left( \left(\mathcal{D}_{JS}\left( P_s(Z) \parallel P_{t,k}(Z) \right) \right)^{\frac{1}{2}}  
     + \left(\mathcal{D}_{JS} \left( P_s(Z,Y) \parallel P_{t,k}(Z,Y) \right) \right)^{\frac{1}{2}} \right) \nonumber \\ 
     &\overset{(1)}{\leq} \lambda \Er \left(P_s, h \circ f_s \right)  + \Er \left(P_t^u, h \circ f_t \right) - \lambda \Er \left(P_s^u, h \circ f_s \right) \nonumber \\
     &+ \sqrt{2} \lambda C \left( \left(\mathcal{D}_{JS}\left( P_s(Z) \parallel P_{t,k}(Z) \right) \right)^{\frac{1}{2}}  
     + \left(\mathcal{D}_{JS} \left( P_s(Z,Y) \parallel P_{t,k}(Z,g(Z)) \right) \right)^{\frac{1}{2}} + \left( \N\left(P_{t,k},g\right) \right)^{\frac{1}{2}} \right) \label{eq:s18}
\end{align}

We have $\overset{(1)}{\leq}$ by applying Eq. (\ref{eq:s10}) and using inequality $\sqrt{a+b} \leq \sqrt{a} + \sqrt{b}$. Finally, by applying Lemma~\ref{lemma:s4} for $\Er \left(P_s, h \circ f_s \right)$, $\Er \left(P_t^u, h \circ f_t \right)$, and $\Er \left(P_s^u, h \circ f_s \right)$ in Eq. (\ref{eq:s18}), then with probability at least $1 - \delta$ over the choice of source and target datasets $D_s$ and $D_t$, we have:

\begin{align}
     \Er \left(P_t, h \circ f_t \right) &\leq \lambda \widehat{\Er} \left(P_s, h \circ f_s \right)  + \widehat{\Er} \left(P_t^u, h \circ f_t \right) - \lambda \widehat{\Er} \left(P_s^u, h \circ f_s \right) \nonumber \\ 
     &+ \sqrt{2} \lambda C \left( \left(\mathcal{D}_{JS}\left( P_s(Z) \parallel P_{t,k}(Z) \right) \right)^{\frac{1}{2}}  
     + \left(\mathcal{D}_{JS} \left( P_s(Z,Y) \parallel P_{t,k}(Z,g(Z)) \right) \right)^{\frac{1}{2}} + \left( \N\left(P_{t,k},g\right) \right)^{\frac{1}{2}} \right) \nonumber \\
     &+ 2\lambda\mathcal{R}_{D_s} \left( \mathcal{L} \circ \mathcal{H} \circ \mathcal{F}_s \right) + 2\mathcal{R}_{D_t^u} \left( \mathcal{L} \circ \mathcal{H} \circ \mathcal{F}_t \right) + 2\lambda\mathcal{R}_{D_s^u} \left( \mathcal{L} \circ \mathcal{H} \circ \mathcal{F}_s \right) + \mathcal{O}\left( C \sqrt{\log (1 / \delta)} \left( \frac{\lambda}{\sqrt{n_s}} + \frac{1}{\sqrt{n_t}} \right) \right) \nonumber \\
     &\overset{(1)}{\leq} \lambda \widehat{\Er} \left(P_s, h \circ f_s \right) + \widehat{\Er} \left(P_t^u, h \circ f_t \right)  - \lambda \widehat{\Er} \left(P_s^u, h \circ f_s \right) \nonumber \\
     &+ \sqrt{2} \lambda C \left( \left(\mathcal{D}_{JS}\left( P_s(Z) \parallel P_{t,k}(Z) \right) \right)^{\frac{1}{2}} + \left(\mathcal{D}_{JS} \left( P_s(Z,Y) \parallel P_{t,k}(Z,g(Z)) \right) \right)^{\frac{1}{2}} + \left(\N(P_{t,k}, g)\right)^{\frac{1}{2}}\right)  \nonumber \\
     & + \mathcal{O} \left( \lambda C  \sqrt{\frac{d_s \log{n_s} + d_s \log{|\mathcal{Y}^t| + \log{\frac{1}{\delta}}}}{n_s}} + C \sqrt{\frac{d_t \log{n_t} + d_t \log{|\mathcal{Y}^t| + \log{\frac{1}{\delta}}}}{n_t}} \right)
\end{align}
where $D_s^u$ and $D_t^u$ are datasets induced from $D_s$ and $D_t$ using mappings $f_s^u$ and $f_t^u$, respectively. We have $\overset{(1)}{\leq}$ by applying Lemma~\ref{lemma:s5} for $\mathcal{R}_{D_s} \left( \mathcal{L} \circ \mathcal{H} \circ \mathcal{F}_s \right)$, $\mathcal{R}_{D_t^u} \left( \mathcal{L} \circ \mathcal{H} \circ \mathcal{F}_t \right)$, and $\mathcal{R}_{D_s^u} \left( \mathcal{L} \circ \mathcal{H} \circ \mathcal{F}_s \right)$.

\paragraph{Proof of Proposition~\ref{thm:2}}

We have:

\begin{align*}
    \Er \left( P_t, h \circ f \right) &\overset{(1)}{\leq} \Er \left( P_s, h \circ f \right) + \sqrt{2}C \left( \mathcal{D}_{JS} \left( P_s(Y,Z) \parallel P_t(Y,Z) \right) \right)^{1/2} \\
    &\overset{(2)}{\leq} \Er \left( P_s, h \circ f \right) + \sqrt{2}C \left( \mathcal{D}_{JS} \left( P_s(Z) \parallel P_t(Z) \right) + 2 \mathbb{E}_{P_{s,t}}\left[\mathcal{D}_{JS}\left(P_s(Y|Z) \parallel P_t(Y|Z) \right) \right] \right)^{1/2} \\
    &\overset{(3)}{=} \Er \left( P_s, h \circ f \right) + \sqrt{2}C \left( \mathcal{D}_{JS} \left( P_s(Z) \parallel P_t(Z) \right) + 2 \mathbb{E}_{P_{s,t}}\left[\mathcal{D}_{JS}\left(P_s(Y|Z) \parallel P_s(Y|Z) \right) \right] \right)^{1/2} \\
    &= \Er \left( P_s, h \circ f \right) + \sqrt{2}C \left( \mathcal{D}_{JS} \left( P_s(Z) \parallel P_t(Z) \right) \right)^{1/2}
\end{align*}

We have $\overset{(1)}{\leq}$ by applying Lemma~\ref{lemma:s1} for two distributions $P_s(Z,Y)$ and $P_t(Z,Y)$, and note that $\Er \left( P_d, h \circ f \right) = \mathbb{E}_{P_d(X,Y)}\left[ L(h(f(X)),Y) \right] = \mathbb{E}_{P_d(Z,Y)}\left[ L(h(Z),Y) \right] = \Er \left( P_d, h \right)$ for $d \in \left\{s,t\right\}$; $\overset{(2)}{\leq}$ by applying Lemma~\ref{lemma:s2} for $\mathcal{D}_{JS} \left( P_s(Y,Z) \parallel P_t(Y,Z) \right)$; $\overset{(3)}{=}$ by applying Lemma~\ref{lemma:s3}.
\section{Model Details}\label{app:model}
In this section, we provide a comprehensive overview of the architectures utilized in our experiment, along with the pseudocode for training our proposed method.

\subsection{Backbone Architecture Details}\label{app:architect}
To ensure fair comparisons between methods, we use the same backbone architectures across all approaches. Most methods involve representation mapping and classifier networks, except for SSAN and SCT, which include an additional discriminator network, and KPG, which employs optimal transport to transform data within the input space. For representation mapping, we implement multi-layer feed-forward neural networks on the CIFAR10 \& ILSVRC2012, Wikipedia, Multilingual Reuters Collection, NUSWIDE \& ImageNet, Office \& Caltech256, and ImageCLEF-DA datasets, while using ResNet/WideResNet to encode paper and digital ECG data in the PTB-XL dataset. A linear classifier is applied across all datasets. The specifics of these networks are detailed in Tables~\ref{tab:s1}-\ref{tab:s2} below.

\begin{table}[H]
    \centering
    \caption{Details of backbone networks used in CIFAR10 \& ILSVRC2012, Wikipedia, Multilingual Reuters Collection, NUSWIDE \& ImageNet, Office \& Caltech256, and ImageCLEF-DA datasets. \textbf{d\_source}, \textbf{d\_target}, and \textbf{n\_output} represent the dimensions of the source feature space, target feature space, and output space, respectively. For RL-OSHeDA and OPDA, \textbf{n\_output} is set to $\left | \mathcal{Y}^t \right|$ while for other methods, \textbf{n\_output} is set to $\left | \mathcal{Y}^s \right|$.}
    \label{tab:s1}
    \begin{tabular}{ll}
    \hline
Networks                                      & Layers                                                       \\\hline
\multirow{5}{*}{Representation Mapping $f_s$} & Linear(input dim=\textbf{d\_source}, output dim=(\textbf{d\_source} + 256)/2) \\
                                              & LeakyReLU(negative\_slope=0.2)                               \\
                                              & Linear(input dim=(\textbf{d\_source} + 256)/2, output dim=256)       \\
                                              & LeakyReLU(negative\_slope=0.2)                               \\
                                              & Normalize(p=2)                                               \\\hline
\multirow{5}{*}{Representation Mapping $f_t$} & Linear(input dim=\textbf{d\_target}, output dim=(\textbf{d\_target} + 256)/2) \\
                                              & LeakyReLU(negative\_slope=0.2)                               \\
                                              & Linear(input dim=(\textbf{d\_target} + 256)/2, output dim=256)       \\
                                              & LeakyReLU(negative\_slope=0.2)                               \\
                                              & Normalize(p=2)                                               \\\hline
\multirow{2}{*}{Classifier $h$}               & Linear(input dim=256, output dim=\textbf{n\_output})                  \\
                                              & LeakyReLU(negative\_slope=0.2)                              \\\hline
\end{tabular}
\end{table}

\begin{table}[H]
    \centering
    \caption{Details of backbone networks used in PTB-XL dataset. The representation mapping networks are constructed from multiple ResNetBlock1d (for digital ECG) and ResNetBlock2d (for paper ECG) sub-modules. \textbf{n\_output} represent the dimensions of output space. For RL-OSHeDA and OPDA, \textbf{n\_output} is set to $\left | \mathcal{Y}^t \right|$ while for other methods, \textbf{n\_output} is set to $\left | \mathcal{Y}^s \right|$.}
    \label{tab:my_label}
    \begin{tabular}{ll}
    \hline
Networks                           & Layers                              \\\hline
\multirow{6}{*}{ResNetBlock1d(i\_ch, o\_ch)}   & Conv1d(input channel=\textbf{i\_ch}, output channel=\textbf{o\_ch}, kernel=3, stride=\textbf{o\_ch} / \textbf{i\_ch}, padding=1) \\
                                               & BatchNorm1d                                                                                  \\
                                               & ReLU                                                                                         \\
                                               & Conv1d(input channel=\textbf{o\_ch}, output channel=\textbf{o\_ch}, kernel=3, padding=1)                       \\
                                               & BatchNorm1d                                                                                  \\
                                               & ReLU                                                                                         \\\hline
\multirow{6}{*}{ResNetBlock2d(i\_ch, o\_ch)}   & Conv2d(input channel=\textbf{i\_ch}, output channel=\textbf{o\_ch}, kernel=3, stride=\textbf{o\_ch} / \textbf{i\_ch}, padding=1) \\
                                               & BatchNorm2d                                                                                  \\
                                               & ReLU                                                                                         \\
                                               & Conv2d(input channel=\textbf{o\_ch}, output channel=\textbf{o\_ch}, kernel=3, padding=1)                       \\
                                               & BatchNorm2d                                                                                  \\
                                               & ReLU                                                                                         \\\hline
\multirow{16}{*}{ResNet1d}                     & Conv1d(input channel=1, output channel=64, kernel=7, stride=2, padding=3)                    \\
                                               & BatchNorm1d                                                                                  \\
                                               & ReLU                                                                                         \\
                                               & MaxPool1d                                                                                    \\
                                               & ResNetBlock1d(i\_ch=64, o\_ch=64) $\times$ 2                                                            \\
                                               & ResNetBlock1d(i\_ch=64, o\_ch=128)                                                           \\
                                               & ResNetBlock1d(i\_ch=128, o\_ch=128)                                                          \\
                                               & ResNetBlock1d(i\_ch=128, o\_ch=256)                                                          \\
                                               & ResNetBlock1d(i\_ch=256, o\_ch=256)                                                          \\
                                               & ResNetBlock1d(i\_ch=256, o\_ch=512)                                                          \\
                                               & ResNetBlock1d(i\_ch=512, o\_ch=512)                                                          \\
                                               & BatchNorm1d                                                                                  \\
                                               & ReLU                                                                                         \\
                                               & AdaptiveAvgPool1d(output size=7)                                                             \\
                                               & Linear(input dim=3584, output dim=256)                                                       \\\hline
\multirow{2}{*}{Representation Mapping $f_s$}  & Concatenation of 12 ResNet1d modules                                                            \\
                                               & Linear(input dim=256 $\times$ 12, output dim=256)                                             \\\hline
\multirow{14}{*}{Representation Mapping $f_t$} & Conv2d(input channel=3, output channel=64, kernel=7, stride=2, padding=3)                    \\
                                               & BatchNorm2d                                                                                  \\
                                               & ReLU                                                                                         \\
                                               & MaxPool2d                                                                                    \\
                                               & ResNetBlock2d(i\_ch=64, o\_ch=64) $\times$ 2                                                            \\
                                               & ResNetBlock2d(i\_ch=64, o\_ch=128)                                                           \\
                                               & ResNetBlock2d(i\_ch=128, o\_ch=128)                                                          \\
                                               & ResNetBlock2d(i\_ch=128, o\_ch=256)                                                          \\
                                               & ResNetBlock2d(i\_ch=256, o\_ch=256)                                                          \\
                                               & ResNetBlock2d(i\_ch=256, o\_ch=512)                                                          \\
                                               & ResNetBlock2d(i\_ch=512, o\_ch=512)                                                          \\
                                               & AdaptiveAvgPool2d(output size=1)                                                             \\
                                               & Linear(input dim=512, output dim=256)                                                        \\\hline
\multirow{2}{*}{Classifier $h$}                & Linear(input dim=256, output dim=\textbf{n\_output})                                                  \\
                                               & LeakyReLU(negative\_slope=0.2)                                       \\\hline
\end{tabular}
\end{table}

\begin{table}[H]
    \centering
    \caption{Details of discriminator network used in SSAN and SCT methods.}
    \label{tab:my_label}
    \begin{tabular}{ll}
    \hline
Networks                           & Layers                              \\\hline
\multirow{2}{*}{Discriminator $D$} & Linear(input dim=256, output dim=1) \\
                                   & Sigmoid                              \\\hline
\end{tabular}
\end{table}

\begin{table}[H]
    \centering
    \caption{Details of the WideResNet backbone. This network is employed in DS3L for the PTB-XL dataset, in place of the standard ResNet backbone.}
    \label{tab:s2}
    \begin{tabular}{ll}
    \hline
Networks                           & Layers                              \\\hline
\multirow{6}{*}{ResNetUnit(i\_ch, o\_ch)}     & BatchNorm2d                                                                                  \\
                                              & LeakyReLU(negative\_slope=0.1)                                                               \\
                                              & Conv2d(input channel=\textbf{i\_ch}, output channel=\textbf{o\_ch}, kernel=3, stride=\textbf{o\_ch} / \textbf{i\_ch}, padding=1) \\
                                              & BatchNorm2d                                                                                  \\
                                              & LeakyReLU(negative\_slope=0.1)                                                               \\
                                              & Conv2d(input channel=\textbf{o\_ch}, output channel=\textbf{o\_ch}, kernel=3, stride=1, padding=1)             \\\hline
\multirow{2}{*}{ResNetBlock(i\_ch, o\_ch)}    & ResNetUnit(i\_ch=\textbf{i\_ch}, o\_ch=\textbf{o\_ch})                                                         \\
                                              & ResNetUnit(i\_ch=\textbf{o\_ch}, o\_ch=\textbf{o\_ch}) $\times$ 3                                              \\\hline
\multirow{8}{*}{Representation Mapping $f_t$} & Conv2d(input channel=3, output channel=16, kernel=3, padding=1)                              \\
                                              & ResNetBlock(i\_ch=16, o\_ch=32)                                                              \\
                                              & ResNetBlock(i\_ch=32, o\_ch=64)                                                              \\
                                              & ResNetBlock(i\_ch=64, o\_ch=128)                                                             \\
                                              & BatchNorm2d                                                                                  \\
                                              & LeakyReLU(negative\_slope=0.1)                                                               \\
                                              & AveragePool2d(output size=1)                                                                        \\
                                              & Linear(input dim=128, output dim=256)                                                        \\\hline
\end{tabular}                              
\end{table}

\subsection{Training Details}\label{app:training}
We utilize a two-stage learning approach for our proposed method, RL-OSHeDA. Specifically, in stage 1, we update the model parameters by optimizing $L_{cls}$ as defined in Eq. (2). In stage 2, we update the model parameters by optimizing $L$ as specified in Eq. (1), with the assistance of pseudo-labels. The details of this training process are outlined in Algorithm~\ref{alg:train} below.

\begin{algorithm}
\caption{Two-stage learning process for RL-OSHeDA}\label{alg:train}
\begin{algorithmic}[1]
\STATE \textbf{Inputs:} Source datasets $D_s = \left\{ x_i^s, y_i^s \right\}_{i=1}^{n_s}$, labeled target datasets $D_{t_l} = \left\{ x_i^t, y_i^t \right\}_{i=1}^{n_{t_l}}$, unlabeled target datasets $D_{t_u} = \left\{ x_i^t \right\}_{i=1}^{n_{t_u}}$, known class prior $\lambda$, threshold $T$, number of epochs $T_{max}$
\STATE \textbf{Outputs:} Trained network parameters $\theta_{f_s}$, $\theta_{f_t}$, $\theta_{h}$
\STATE Initialize network parameters $\theta_{f_s}$, $\theta_{f_t}$, $\theta_{h}$
\FOR{$epoch = 1$ \textbf{to} $T_{max}$}
    \STATE Sample source minibatch $B_s$
    \STATE Sample labeled target minibatch $B_{t_l}$
    \STATE Sample unlabeled target minibatch $B_{t_u}$
    \IF{$epoch < T$}
        \STATE{/* Stage 1 */}
        \STATE Calculate $L_{cls}$ in Eq. (2) using $B_s$, $B_{t_l}$, $B_{t_u}, \lambda$
        \STATE Update parameters: \\
        $\theta_{f_s} = \theta_{f_s} - \gamma \nabla_{\theta_{f_s}} L_{cls}$ \\
        $\theta_{f_t} = \theta_{f_t} - \gamma \nabla_{\theta_{f_t}} L_{cls}$ \\
        $\theta_{h} = \theta_{h} - \gamma \nabla_{\theta_{h}} L_{cls}$
    \ELSE
        \STATE{/* Stage 2 */}
        \STATE Generate pseudo-labels for $B_{t_u}$
        \STATE Calculate $L_{cls}$ in Eq. (2) using $B_s$, $B_{t_l}$, $B_{t_u}, \lambda$
        \STATE Calculate $L_{inv}$ in Eq. (3) using $B_s$, $B_{t_l}$, $B_{t_u}$, pseudo-labels
        \STATE Calculate $L_{seg}$ in Eq. (4) using $B_s$, $B_{t_l}$, $B_{t_u}$, pseudo-labels
        \STATE Calculate $L_{osd}$ in Eq. (5) using $B_s$, $B_{t_l}$, $B_{t_u}, \lambda$
        \STATE Update parameters: \\
        $\theta_{f_s} = \theta_{f_s} - \gamma \nabla_{\theta_{f_s}} \left( L_{cls} + L_{inv} - L_{seg} + L_{osd} \right)$ \\
        $\theta_{f_t} = \theta_{f_t} - \gamma \nabla_{\theta_{f_t}} \left( L_{cls} + L_{inv} - L_{seg} + L_{osd} \right)$ \\
        $\theta_{h} = \theta_{h} - \gamma \nabla_{\theta_{h}} \left( L_{cls} + L_{inv} - L_{seg} + L_{osd} \right)$
    \ENDIF
\ENDFOR
\RETURN Trained parameters $\theta_{f_s}$, $\theta_{f_t}$, $\theta_{h}$
\end{algorithmic}
\end{algorithm}

\section{Experimental Details}\label{app:exp}
\subsection{Datasets}\label{app:dataset}
We conduct experiments using seven datasets covering the clinical, computer vision, and natural language processing domains. Detailed descriptions of these datasets are provided below, with corresponding data statistics presented in Tables~\ref{tab:s3}-\ref{tab:s4}.

\paragraph{CIFAR-10~\cite{krizhevsky2009learning} \& ILSVRC2012~\cite{russakovsky2015imagenet}.} These two datasets are used for the image-to-image adaptation task. Big Transfer-M with ResNet-50 and ResNet-101~\cite{kolesnikov1912big} are utilized to extract features from the images. In the target domain (ILSVRC2012), 4 out of 8 shared classes are designated as unknown classes. For the source (CIFAR-10) and unlabeled target data, 50 instances are randomly selected for each class, and we randomly choose 1, 3, or 5 instances per class as labeled target data. This process results in 6 DA tasks.

\paragraph{Wikipedia~\cite{rasiwasia2010new}.} This dataset, consisting of text-image pairs, is used for image-to-text and text-to-image adaptation tasks. Big Transfer-M with ResNet-101 and Big Bird~\cite{zaheer2020big} are used to extract features for image and text, respectively. 5 out of 10 classes are designated as unknown classes. All data in the source domain are used as labeled source data. For the target domain, we randomly select 5 instances per class as labeled target data and randomly select 50 instances per class from the remaining data as unlabeled target data. This process results in 2 DA tasks.

\paragraph{Multilingual Reuters Collection~\cite{amini2009learning}.} This dataset, consisting of articles in 5 languages, is used for text-to-text adaptation tasks. Bag-of-Words with TF-IDF, followed by Principal Component Analysis, is used to generate features for each article. English, French, Italian, and German are used as the source domains, while Spanish is used as the target domain. 3 out of 6 classes are designated as unknown classes. For the source and unlabeled target datasets, 100 and 500 instances are randomly selected for each class, respectively, and we randomly choose 20 instances per class as labeled target data. This process results in 4 DA tasks.

\paragraph{NUS-WIDE~\cite{chua2009nus} and ImageNet~\cite{deng2009imagenet}.} These datasets are used for the text-to-image adaptation task. We utilize the tag information from NUS-WIDE as the source domain (text) and the image data from ImageNet as the target domain (image). 4 out of 8 shared classes between the two datasets are designated as unknown classes. Features for NUS-WIDE tags are extracted using a pre-trained 5-layer neural network, while DeCAF6~\cite{donahue2014decaf} features are used for images in ImageNet. For NUS-WIDE, 100 instances per class are selected, whereas for ImageNet, 3 instances per class are sampled as labeled target data, with all remaining images used as unlabeled target data.

\paragraph{Office~\cite{saenko2010adapting} and Caltech-256~\cite{griffin2007caltech}.} These datasets, which include 4 domains Amazon, Webcam, DSLR from Office, and Caltech from Caltech-256, are used for the image-to-image adaptation task. Amazon, Webcam, and Caltech are used as source domains while Amazon, Webcam, DSLR, and Caltech are used as target domains. SURF~\cite{saenko2010adapting} and DeCAF6 are utilized as 2 different feature sets for images in these datasets. 5 out of 10 classes are designated as unknown classes, and 3 instances per class are sampled as labeled target data. This process results in 18 DA tasks.

\paragraph{ImageCLEF-DA~\cite{griffin2007caltech}.} This data, which include 4 domains Caltech, ImageNet, Bing, and PascalVOC, are used for the image-to-image adaptation task. ResNet50 and VGG-19~\cite{simonyan2014very} are utilized as 2 different feature sets for images in this dataset. 6 out of 12 classes are designated as unknown classes, and 3 instances per class are sampled as labeled target data. This process results in 24 DA tasks.

\paragraph{PTB-XL~\cite{wagner2020ptb}.} This dataset is used for the digital-to-paper electrocardiogram (ECG) adaptation task. Frequent classes including NORM, Old MI, STTC, and CD are designated as known classes, while the remaining classes are considered unknown. For each known class, we sample 2,000 instances per class to construct source (1,000 instances) and target (1,000 instances) datasets. All instances from the unknown class are used for the target dataset. This process results in 1 DA task.

\begin{table}[H]
    \centering
    \caption{Data statistics for each domain adaptation task in CIFAR10 \& ILSVRC2012 dataset.}
    \label{tab:s3}
    \resizebox{\textwidth}{!}{
    \begin{tabular}{l|ccc|ccc|ccc}
    \hline
    \multicolumn{10}{c}{CIFAR10 \& ILSVRC2012}\\
    \hline
                                                          & \multicolumn{3}{c|}{Source Dataset}            & \multicolumn{3}{c|}{Labeled Target Dataset}    & \multicolumn{3}{c}{Unlabeled Target Dataset}  \\\cline{2-10}
                                                          & \# of classes & \# of instances & feature dim & \# of classes & \# of instances & feature dim & \# of classes & \# of instances & feature dim \\\hline
ImageNet (ResNet-101) $\Rightarrow$ CIFAR (ResNet-50) (1) & 4             & 2000            & 2048        & 4             & 4               & 2048        & 5             & 4000            & 2048        \\
ImageNet (ResNet-101) $\Rightarrow$ CIFAR (ResNet-50) (3) & 4             & 2000            & 2048        & 4             & 12              & 2048        & 5             & 4000            & 2048        \\
ImageNet (ResNet-101) $\Rightarrow$ CIFAR (ResNet-50) (5) & 4             & 2000            & 2048        & 4             & 20              & 2048        & 5             & 4000            & 2048        \\
ImageNet (ResNet-50) $\Rightarrow$ CIFAR (ResNet-101) (1) & 4             & 2000            & 2048        & 4             & 4               & 2048        & 5             & 4000            & 2048        \\
ImageNet (ResNet-50) $\Rightarrow$ CIFAR (ResNet-101) (3) & 4             & 2000            & 2048        & 4             & 12              & 2048        & 5             & 4000            & 2048        \\
ImageNet (ResNet-50) $\Rightarrow$ CIFAR (ResNet-101) (5) & 4             & 2000            & 2048        & 4             & 20              & 2048        & 5             & 4000            & 2048                \\\hline             
\end{tabular}
}
\end{table}

\begin{table}[H]
    \centering
    \caption{Data statistics for each domain adaptation task in Wikipedia dataset.}
    \label{tab:my_label}
    \resizebox{\textwidth}{!}{
    \begin{tabular}{l|ccc|ccc|ccc}
    \hline
    \multicolumn{10}{c}{Wikipedia}\\
    \hline
                                                          & \multicolumn{3}{c|}{Source Dataset}            & \multicolumn{3}{c|}{Labeled Target Dataset}    & \multicolumn{3}{c}{Unlabeled Target Dataset}  \\\cline{2-10}
                                                          & \# of classes & \# of instances & feature dim & \# of classes & \# of instances & feature dim & \# of classes & \# of instances & feature dim \\\hline
Image $\Rightarrow$ Text & 5             & 1472            & 768         & 5             & 25              & 2048        & 6              & 500             & 2048        \\
Text $\Rightarrow$ Image & 5             & 1472            & 2048        & 5             & 25              & 768         & 6              & 500             & 768         \\\hline      
\end{tabular}
}
\end{table}

\begin{table}[H]
    \centering
    \caption{Data statistics for each domain adaptation task in Multilingual Reuters Collection dataset.}
    \label{tab:my_label}
    \resizebox{\textwidth}{!}{
    \begin{tabular}{l|ccc|ccc|ccc}
    \hline
    \multicolumn{10}{c}{Multilingual Reuters Collection}\\
    \hline
                                                          & \multicolumn{3}{c|}{Source Dataset}            & \multicolumn{3}{c|}{Labeled Target Dataset}    & \multicolumn{3}{c}{Unlabeled Target Dataset}  \\\cline{2-10}
                                                          & \# of classes & \# of instances & feature dim & \# of classes & \# of instances & feature dim & \# of classes & \# of instances & feature dim \\\hline
English $\Rightarrow$ Spanish & 3             & 300             & 1131        & 3             & 60              & 807         & 4              & 2910            & 807         \\
French $\Rightarrow$ Spanish  & 3             & 300             & 1230        & 3             & 60              & 807         & 4              & 2910            & 807         \\
German $\Rightarrow$ Spanish  & 3             & 300             & 1417        & 3             & 60              & 807         & 4              & 2910            & 807         \\
Italian $\Rightarrow$ Spanish & 3             & 300             & 1041        & 3             & 60              & 807         & 4              & 2910            & 807         \\\hline 
\end{tabular}
}
\end{table}

\begin{table}[H]
    \centering
    \caption{Data statistics for each domain adaptation task in NUSWIDE \& ImageNet dataset.}
    \label{tab:my_label}
    \resizebox{\textwidth}{!}{
    \begin{tabular}{l|ccc|ccc|ccc}
    \hline
    \multicolumn{10}{c}{NUSWIDE \& ImageNet}\\
    \hline
                                                          & \multicolumn{3}{c|}{Source Dataset}            & \multicolumn{3}{c|}{Labeled Target Dataset}    & \multicolumn{3}{c}{Unlabeled Target Dataset}  \\\cline{2-10}
                                                          & \# of classes & \# of instances & feature dim & \# of classes & \# of instances & feature dim & \# of classes & \# of instances & feature dim \\\hline
Text $\Rightarrow$ Image & 4             & 400             & 64          & 4             & 12              & 4096        & 5              & 800             & 4096            \\\hline  
\end{tabular}
}
\end{table}

\begin{table}[H]
    \centering
    \caption{Data statistics for each domain adaptation task in Office \& Caltech256 dataset.}
    \label{tab:my_label}
    \resizebox{\textwidth}{!}{
    \begin{tabular}{l|ccc|ccc|ccc}
    \hline
    \multicolumn{10}{c}{Office \& Caltech256}\\
    \hline
                                                          & \multicolumn{3}{c|}{Source Dataset}            & \multicolumn{3}{c|}{Labeled Target Dataset}    & \multicolumn{3}{c}{Unlabeled Target Dataset}  \\\cline{2-10}
                                                          & \# of classes & \# of instances & feature dim & \# of classes & \# of instances & feature dim & \# of classes & \# of instances & feature dim \\\hline
Amazon (DeCAF6) $\Rightarrow$ Caltech (SURF) & 5             & 100             & 4096        & 5             & 15              & 800         & 6             & 1093            & 800         \\
Amazon (DeCAF6) $\Rightarrow$ DSLR (SURF)    & 5             & 100             & 4096        & 5             & 15              & 800         & 6             & 127             & 800         \\
Amazon (DeCAF6) $\Rightarrow$ Webcam (SURF)  & 5             & 100             & 4096        & 5             & 15              & 800         & 6             & 265             & 800         \\
Amazon (SURF) $\Rightarrow$ Caltech (DeCAF6) & 5             & 100             & 800         & 5             & 15              & 4096        & 6             & 1093            & 4096        \\
Amazon (SURF) $\Rightarrow$ DSLR (DeCAF6)    & 5             & 100             & 800         & 5             & 15              & 4096        & 6             & 127             & 4096        \\
Amazon (SURF) $\Rightarrow$ Webcam (DeCAF6)  & 5             & 100             & 800         & 5             & 15              & 4096        & 6             & 265             & 4096        \\
Caltech (DeCAF6) $\Rightarrow$ Amazon (SURF) & 5             & 100             & 4096        & 5             & 15              & 800         & 6             & 928             & 800         \\
Caltech (DeCAF6) $\Rightarrow$ DSLR (SURF)   & 5             & 100             & 4096        & 5             & 15              & 800         & 6             & 127             & 800         \\
Caltech (DeCAF6) $\Rightarrow$ Webcam (SURF) & 5             & 100             & 4096        & 5             & 15              & 800         & 6             & 265             & 800         \\
Caltech (SURF) $\Rightarrow$ Amazon (DeCAF6) & 5             & 100             & 800         & 5             & 15              & 4096        & 6             & 928             & 4096        \\
Caltech (SURF) $\Rightarrow$ DSLR (DeCAF6)   & 5             & 100             & 800         & 5             & 15              & 4096        & 6             & 127             & 4096        \\
Caltech (SURF) $\Rightarrow$ Webcam (DeCAF6) & 5             & 100             & 800         & 5             & 15              & 4096        & 6             & 265             & 4096        \\
Webcam (DeCAF6) $\Rightarrow$ Amazon (SURF)  & 5             & 100             & 4096        & 5             & 15              & 800         & 6             & 928             & 800         \\
Webcam (DeCAF6) $\Rightarrow$ Caltech (SURF) & 5             & 100             & 4096        & 5             & 15              & 800         & 6             & 1093            & 800         \\
Webcam (DeCAF6) $\Rightarrow$ DSLR (SURF)    & 5             & 100             & 4096        & 5             & 15              & 800         & 6             & 127             & 800         \\
Webcam (SURF) $\Rightarrow$ Amazon (DeCAF6)  & 5             & 100             & 800         & 5             & 15              & 4096        & 6             & 928             & 4096        \\
Webcam (SURF) $\Rightarrow$ Caltech (DeCAF6) & 5             & 100             & 800         & 5             & 15              & 4096        & 6             & 1093            & 4096        \\
Webcam (SURF) $\Rightarrow$ DSLR (DeCAF6)    & 5             & 100             & 800         & 5             & 15              & 4096        & 6             & 127             & 4096        \\\hline             
\end{tabular}
}
\end{table}

\begin{table}[H]
    \centering
    \caption{Data statistics for each domain adaptation task in ImageCLEF-DA dataset.}
    \label{tab:my_label}
    \resizebox{\textwidth}{!}{
    \begin{tabular}{l|ccc|ccc|ccc}
    \hline
    \multicolumn{10}{c}{ImageCLEF-DA}\\
    \hline
                                                          & \multicolumn{3}{c|}{Source Dataset}            & \multicolumn{3}{c|}{Labeled Target Dataset}    & \multicolumn{3}{c}{Unlabeled Target Dataset}  \\\cline{2-10}
                                                          & \# of classes & \# of instances & feature dim & \# of classes & \# of instances & feature dim & \# of classes & \# of instances & feature dim \\\hline
Bing (Reset-50) $\Rightarrow$ Caltech (VGG-19)       & 6             & 120             & 2048        & 6             & 18              & 4096        & 7             & 564             & 4096        \\
Bing (Reset-50) $\Rightarrow$ ImageNet (VGG-19)      & 6             & 120             & 2048        & 6             & 18              & 4096        & 7             & 564             & 4096        \\
Bing (Reset-50) $\Rightarrow$ PascalVOC (VGG-19)     & 6             & 120             & 2048        & 6             & 18              & 4096        & 7             & 564             & 4096        \\
Bing (VGG-19) $\Rightarrow$ Caltech (Reset-50)       & 6             & 120             & 4096        & 6             & 18              & 2048        & 7             & 564             & 2048        \\
Bing (VGG-19) $\Rightarrow$ ImageNet (Reset-50)      & 6             & 120             & 4096        & 6             & 18              & 2048        & 7             & 564             & 2048        \\
Bing (VGG-19) $\Rightarrow$ PascalVOC (Reset-50)     & 6             & 120             & 4096        & 6             & 18              & 2048        & 7             & 564             & 2048        \\
Caltech (Reset-50) $\Rightarrow$ Bing (VGG-19)       & 6             & 120             & 2048        & 6             & 18              & 4096        & 7             & 564             & 4096        \\
Caltech (Reset-50) $\Rightarrow$ ImageNet (VGG-19)   & 6             & 120             & 2048        & 6             & 18              & 4096        & 7             & 564             & 4096        \\
Caltech (Reset-50) $\Rightarrow$ PascalVOC (VGG-19)  & 6             & 120             & 2048        & 6             & 18              & 4096        & 7             & 564             & 4096        \\
Caltech (VGG-19) $\Rightarrow$ Bing (Reset-50)       & 6             & 120             & 4096        & 6             & 18              & 2048        & 7             & 564             & 2048        \\
Caltech (VGG-19) $\Rightarrow$ ImageNet (Reset-50)   & 6             & 120             & 4096        & 6             & 18              & 2048        & 7             & 564             & 2048        \\
Caltech (VGG-19) $\Rightarrow$ PascalVOC (Reset-50)  & 6             & 120             & 4096        & 6             & 18              & 2048        & 7             & 564             & 2048        \\
ImageNet (Reset-50) $\Rightarrow$ Bing (VGG-19)      & 6             & 120             & 2048        & 6             & 18              & 4096        & 7             & 564             & 4096        \\
ImageNet (Reset-50) $\Rightarrow$ Caltech (VGG-19)   & 6             & 120             & 2048        & 6             & 18              & 4096        & 7             & 564             & 4096        \\
ImageNet (Reset-50) $\Rightarrow$ PascalVOC (VGG-19) & 6             & 120             & 2048        & 6             & 18              & 4096        & 7             & 564             & 4096        \\
ImageNet (VGG-19) $\Rightarrow$ Bing (Reset-50)      & 6             & 120             & 4096        & 6             & 18              & 2048        & 7             & 564             & 2048        \\
ImageNet (VGG-19) $\Rightarrow$ Caltech (Reset-50)   & 6             & 120             & 4096        & 6             & 18              & 2048        & 7             & 564             & 2048        \\
ImageNet (VGG-19) $\Rightarrow$ PascalVOC (Reset-50) & 6             & 120             & 4096        & 6             & 18              & 2048        & 7             & 564             & 2048        \\
PascalVOC (Reset-50) $\Rightarrow$ Bing (VGG-19)     & 6             & 120             & 2048        & 6             & 18              & 4096        & 7             & 564             & 4096        \\
PascalVOC (Reset-50) $\Rightarrow$ Caltech (VGG-19)  & 6             & 120             & 2048        & 6             & 18              & 4096        & 7             & 564             & 4096        \\
PascalVOC (Reset-50) $\Rightarrow$ ImageNet (VGG-19) & 6             & 120             & 2048        & 6             & 18              & 4096        & 7             & 564             & 4096        \\
PascalVOC (VGG-19) $\Rightarrow$ Bing (Reset-50)     & 6             & 120             & 4096        & 6             & 18              & 2048        & 7             & 564             & 2048        \\
PascalVOC (VGG-19) $\Rightarrow$ Caltech (Reset-50)  & 6             & 120             & 4096        & 6             & 18              & 2048        & 7             & 564             & 2048        \\
PascalVOC (VGG-19) $\Rightarrow$ ImageNet (Reset-50) & 6             & 120             & 4096        & 6             & 18              & 2048        & 7             & 564             & 2048        \\\hline             
\end{tabular}
}
\end{table}

\begin{table}[H]
    \centering
    \caption{Data statistics for each domain adaptation task in PTB-XL dataset.}
    \label{tab:s4}
    \resizebox{\textwidth}{!}{
    \begin{tabular}{l|ccc|ccc|ccc}
    \hline
    \multicolumn{10}{c}{PTB-XL}\\
    \hline
                                                          & \multicolumn{3}{c|}{Source Dataset}            & \multicolumn{3}{c|}{Labeled Target Dataset}    & \multicolumn{3}{c}{Unlabeled Target Dataset}  \\\cline{2-10}
                                                          & \# of classes & \# of instances & feature dim & \# of classes & \# of instances & feature dim & \# of classes & \# of instances & feature dim \\\hline
Digital ECG $\Rightarrow$ Paper ECG & 4             & 4000            & 12$\times$5000 & 4             & 80              & 3$\times$224$\times$224 & 5             & 4602            & 3$\times$224$\times$224        \\\hline             
\end{tabular}
}
\end{table}

\subsection{Baselines}\label{app:baseline}
We experiment with diverse baselines from heterogeneous domain adaptation, open-set domain adaptation, open-set semi-supervised learning, supervised learning, and semi-supervised learning. The details of these baselines are as follows.

\paragraph{Heterogeneous Domain Adaptation.}
Heterogeneous domain adaptation methods are trained on both source and target data. During inference, these methods classify instances as unknown using the same method as our pseudo-label model $g$ (see Section~\ref{sec:pl}).

\begin{itemize}
    \item SSAN~\cite{li2020simultaneous}: This method maps heterogeneous source features into a shared representation space and then makes predictions from this space. To adapt from the source to the target domains, it aligns the marginal and label-conditional representation distributions between the two domains using Maximum Mean Discrepancy (MMD). Additionally, it employs pseudo-labels generated from geometric similarity and hard predictions made by the classifier.
    \item STN~\cite{yao2019heterogeneous}: This method is similar to SSAN, but it calculates the MMD distance using soft labels (i.e., softmax probabilities) rather than hard labels.
    \item SCT~\cite{zhao2022semantic}: This method is similar to SSAN but differs in that it aligns marginal and label-conditional representation distributions between source and target domains using cosine similarity. Additionally, it generates pseudo-labels based on geometric distances from the source data.
    \item KPG~\cite{gu2022keypoint}: This method utilizes partial optimal transport and the Gromov-Wasserstein distance to map features from the source domain to the target domain. An SVM, trained on the transported source data and labeled target data, is then used for making predictions.
\end{itemize}

\paragraph{Open-Set Domain Adaptation.}

\begin{itemize}
    \item OPDA~\cite{saito2018open}: This method is trained exclusively on target data. It employs adversarial training to train a classifier that distinguishes between labeled and unlabeled target samples, while a generator is trained to push the unlabeled samples away from the decision boundary. This setup provides the generator with two options: aligning unlabeled samples with labeled ones or classifying them as unknown. Consequently, this approach enables the extraction of features that effectively differentiate between known and unknown target samples. Unlike other baselines, OPDA can directly classify instances as unknown during inference based on the classifier's output.
\end{itemize}

\paragraph{Open-Set Semi-Supervised Learning.}

\begin{itemize}
    \item DS3L~\cite{guo2020safe}: This method is trained exclusively on target data and selectively uses unlabeled data while monitoring its impact to mitigate performance risks. Specifically, DS3L weakens the influence of unlabeled data with unknown classes to enhance distribution matching and maintain strong generalization. Simultaneously, it reinforces the use of labeled data to prevent performance degradation. These considerations are integrated into a unified bi-level optimization framework. During inference, DS3L classifies instances as unknown using the same method as our pseudo-label model $g$.
\end{itemize}

\paragraph{Supervised Learning.}

\begin{itemize}
    \item Supervised Learning (SL): We train the model directly on the labeled target dataset by minimizing the classification loss. During inference, SL classifies instances as unknown using the same method as our pseudo-label model $g$.
\end{itemize}

\paragraph{Semi-Supervised Learning.}

\begin{itemize}
    \item Pseudo Labeling (PL): Similar to supervised learning, but we use the model to generate pseudo-labels for the unlabeled target data and then incorporate these pseudo-labeled examples into the training process. During inference, PL classifies instances as unknown using the same method as our pseudo-label model $g$.
\end{itemize}

\subsection{Implementation Details}

Data, model implementation, and training script are included in the code \& data supplementary material. We train each model on each domain adaptation task with 10 different random seeds and report the average prediction performances. All experiments are conducted on a machine with 24-Core CPU, 4 RTX A4000 GPUs, and 128G RAM.

\subsection{Additional Results}\label{app:result}

In this section, we provide a comprehensive overview of the results for domain adaptation tasks across seven datasets. Detailed results are presented in Tables~\ref{tab:s5}-\ref{tab:s6}. Additionally, Table~\ref{tab:s7} includes the results of statistical tests used to assess the significance of our method in comparison to the baseline approaches across all domain adaptation tasks.

\begin{table}[H]
    \centering
    \caption{Prediction performances of RL-OSHeDA as well as baselines for all domain adaptation tasks in CIFAR10 \& ILSVRC2012 dataset.}
    \label{tab:s5}
    \resizebox{\textwidth}{!}{
\begin{tabular}{l|lll|lll|lll}
    \hline
    \multicolumn{10}{c}{CIFAR10 \& ILSVRC2012}\\\hline
                                                          & \multicolumn{3}{c|}{DS3L}                                                    & \multicolumn{3}{c|}{KPG}                                                     & \multicolumn{3}{c}{OPDA}                                                    \\\cline{2-10}
                                                          & \multicolumn{1}{c}{$HOS$} & \multicolumn{1}{c}{$OS^{\ast}$} & \multicolumn{1}{c|}{$UNK$} & \multicolumn{1}{c}{$HOS$} & \multicolumn{1}{c}{$OS^{\ast}$} & \multicolumn{1}{c|}{$UNK$} & \multicolumn{1}{c}{$HOS$} & \multicolumn{1}{c}{$OS^{\ast}$} & \multicolumn{1}{c}{$UNK$} \\\hline
ImageNet (ResNet-101) $\Rightarrow$ CIFAR (ResNet-50) (1) & 52.69$\pm$0.74          & 48.53$\pm$0.93          & 57.77$\pm$1.10          & 56.92$\pm$0.48          & 53.95$\pm$0.00          & 60.33$\pm$1.11          & 45.42$\pm$0.75          & 38.96$\pm$0.90          & 54.83$\pm$1.07          \\
ImageNet (ResNet-101) $\Rightarrow$ CIFAR (ResNet-50) (3) & 65.87$\pm$0.72          & 64.34$\pm$0.99          & 67.49$\pm$1.03          & 57.25$\pm$0.48          & 54.98$\pm$0.00          & 59.77$\pm$1.12          & 57.34$\pm$0.81          & 53.38$\pm$1.06          & 62.25$\pm$1.10          \\
ImageNet (ResNet-101) $\Rightarrow$ CIFAR (ResNet-50) (5) & 66.98$\pm$0.73          & 66.02$\pm$1.01          & 67.97$\pm$1.08          & 57.32$\pm$0.52          & 55.73$\pm$0.00          & 59.02$\pm$1.13          & 58.35$\pm$0.79          & 54.74$\pm$1.10          & 62.65$\pm$1.13          \\
ImageNet (ResNet-50) $\Rightarrow$ CIFAR (ResNet-101) (1) & 52.70$\pm$0.78          & 47.72$\pm$1.11          & 59.02$\pm$1.07          & 52.68$\pm$0.51          & 47.70$\pm$0.00          & 59.53$\pm$1.14          & 44.22$\pm$0.76          & 36.41$\pm$0.94          & 56.61$\pm$1.15          \\
\hline    
\end{tabular}
}
\end{table}

\begin{table}[H]
    \centering
    \caption{Prediction performances of RL-OSHeDA as well as baselines for all domain adaptation tasks in Multilingual Reuters Collection dataset.}
    \label{tab:todo_3}
    \resizebox{\textwidth}{!}{
    \begin{tabular}{l|lll|lll|lll}
    \hline
    \multicolumn{10}{c}{Multilingual Reuters Collection}\\\hline
                                                          & \multicolumn{3}{c|}{DS3L}                                                    & \multicolumn{3}{c|}{KPG}                                                     & \multicolumn{3}{c}{OPDA}                                                    \\\cline{2-10}
                                                          & \multicolumn{1}{c}{$HOS$} & \multicolumn{1}{c}{$OS^{\ast}$} & \multicolumn{1}{c|}{$UNK$} & \multicolumn{1}{c}{$HOS$} & \multicolumn{1}{c}{$OS^{\ast}$} & \multicolumn{1}{c|}{$UNK$} & \multicolumn{1}{c}{$HOS$} & \multicolumn{1}{c}{$OS^{\ast}$} & \multicolumn{1}{c}{$UNK$} \\\hline
English $\Rightarrow$ Spanish & 59.35$\pm$0.96          & 52.92$\pm$1.31          & 67.57$\pm$1.20          & 11.48$\pm$0.08          & 8.80$\pm$0.00           & 17.12$\pm$0.95          & 56.42$\pm$1.02          & 49.09$\pm$1.32          & 66.41$\pm$1.23          \\
French $\Rightarrow$ Spanish  & 59.35$\pm$0.95          & 52.92$\pm$1.25          & 67.57$\pm$1.26          & 11.32$\pm$0.08          & 8.65$\pm$0.00           & 17.02$\pm$0.96          & 56.37$\pm$0.94          & 49.14$\pm$1.24          & 66.14$\pm$1.22          \\
German $\Rightarrow$ Spanish  & 59.35$\pm$0.92          & 52.92$\pm$1.24          & 67.57$\pm$1.27          & 11.11$\pm$0.07          & 8.43$\pm$0.00           & 17.00$\pm$0.97          & 55.32$\pm$0.85          & 48.16$\pm$1.05          & 65.07$\pm$1.21          \\
Italian $\Rightarrow$ Spanish & 59.35$\pm$0.94          & 52.92$\pm$1.30          & 67.57$\pm$1.22          & 11.17$\pm$0.07          & 8.48$\pm$0.00           & 17.02$\pm$0.97          & 55.27$\pm$1.02          & 47.51$\pm$1.30          & 66.15$\pm$1.24   \\
\hline    
\end{tabular}
}
\end{table}

\begin{table}[H]
    \centering
    \caption{Prediction performances of RL-OSHeDA as well as baselines for all domain adaptation tasks in NUSWIDE \& ImageNet dataset.}
    \label{tab:todo_4}
    \resizebox{\textwidth}{!}{
    \begin{tabular}{l|lll|lll|lll}
    \hline
    \multicolumn{10}{c}{NUSWIDE \& ImageNet}\\\hline
                                                          & \multicolumn{3}{c|}{DS3L}                                                    & \multicolumn{3}{c|}{KPG}                                                     & \multicolumn{3}{c}{OPDA}                                                    \\\cline{2-10}
                                                          & \multicolumn{1}{c}{$HOS$} & \multicolumn{1}{c}{$OS^{\ast}$} & \multicolumn{1}{c|}{$UNK$} & \multicolumn{1}{c}{$HOS$} & \multicolumn{1}{c}{$OS^{\ast}$} & \multicolumn{1}{c|}{$UNK$} & \multicolumn{1}{c}{$HOS$} & \multicolumn{1}{c}{$OS^{\ast}$} & \multicolumn{1}{c}{$UNK$} \\\hline
Text $\Rightarrow$ Image & 67.61$\pm$1.65 & 66.17$\pm$2.22          & 69.20$\pm$2.30          & 55.18$\pm$1.17 & 52.60$\pm$0.00          & 58.10$\pm$2.45          & 71.06$\pm$1.44 & 66.60$\pm$1.98          & 76.38$\pm$2.09   \\\hline      
\end{tabular}
}
\resizebox{\textwidth}{!}{
\begin{tabular}{l|lll|lll|lll}
\hline
                                                          & \multicolumn{3}{c|}{PL}                                                      & \multicolumn{3}{c|}{SCT}                                                     & \multicolumn{3}{c}{SSAN}                                                    \\\cline{2-10}
                                                          & \multicolumn{1}{c}{$HOS$} & \multicolumn{1}{c}{$OS^{\ast}$} & \multicolumn{1}{c|}{$UNK$} & \multicolumn{1}{c}{$HOS$} & \multicolumn{1}{c}{$OS^{\ast}$} & \multicolumn{1}{c|}{$UNK$} & \multicolumn{1}{c}{$HOS$} & \multicolumn{1}{c}{$OS^{\ast}$} & \multicolumn{1}{c}{$UNK$} \\\hline
Text $\Rightarrow$ Image & 42.43$\pm$0.26 & 34.05$\pm$0.00          & 61.15$\pm$2.10          & 70.42$\pm$1.49 & 68.00$\pm$2.20          & 73.10$\pm$1.99          & 67.98$\pm$1.49 & 66.25$\pm$2.04          & 69.85$\pm$2.21       \\\hline  
\end{tabular}
}
\resizebox{\textwidth}{!}{
\begin{tabular}{l|lll|lll|lll}
\hline
                                                          & \multicolumn{3}{c|}{STN}                                                     & \multicolumn{3}{c|}{SL}                                                      & \multicolumn{3}{c}{OSHeDA}                                                  \\\cline{2-10}
                                                          & \multicolumn{1}{c}{$HOS$} & \multicolumn{1}{c}{$OS^{\ast}$} & \multicolumn{1}{c|}{$UNK$} & \multicolumn{1}{c}{$HOS$} & \multicolumn{1}{c}{$OS^{\ast}$} & \multicolumn{1}{c|}{$UNK$} & \multicolumn{1}{c}{$HOS$} & \multicolumn{1}{c}{$OS^{\ast}$} & \multicolumn{1}{c}{$UNK$} \\\hline
Text $\Rightarrow$ Image & 67.75$\pm$1.23 & 64.80$\pm$1.42          & 71.08$\pm$2.16          & 69.41$\pm$1.64 & 66.62$\pm$2.26          & 72.57$\pm$2.23          & \textbf{80.01$\pm$1.30} & \textbf{74.65$\pm$2.01} & \textbf{86.35$\pm$0.81}     \\\hline    
\end{tabular}
}
\end{table}

\begin{table}[H]
    \centering
    \caption{Prediction performances of RL-OSHeDA as well as baselines for all domain adaptation tasks in Wikipedia dataset.}
    \label{tab:todo_6}
    \resizebox{\textwidth}{!}{
    \begin{tabular}{l|lll|lll|lll}
    \hline
    \multicolumn{10}{c}{Wikipedia}\\\hline
                                                          & \multicolumn{3}{c|}{DS3L}                                                    & \multicolumn{3}{c|}{KPG}                                                     & \multicolumn{3}{c}{OPDA}                                                    \\\cline{2-10}
                                                          & \multicolumn{1}{c}{$HOS$} & \multicolumn{1}{c}{$OS^{\ast}$} & \multicolumn{1}{c|}{$UNK$} & \multicolumn{1}{c}{$HOS$} & \multicolumn{1}{c}{$OS^{\ast}$} & \multicolumn{1}{c|}{$UNK$} & \multicolumn{1}{c}{$HOS$} & \multicolumn{1}{c}{$OS^{\ast}$} & \multicolumn{1}{c}{$UNK$} \\\hline
Image $\Rightarrow$ Text & 35.90$\pm$2.42          & 26.96$\pm$2.70          & 54.68$\pm$3.09          & 25.37$\pm$0.40          & 16.80$\pm$0.00          & 51.92$\pm$3.19          & 39.60$\pm$2.46          & 29.48$\pm$2.60          & \textbf{61.16$\pm$3.26} \\
Text $\Rightarrow$ Image & 76.10$\pm$1.61          & 74.48$\pm$2.02          & 77.80$\pm$2.51          & 24.27$\pm$0.44          & 16.00$\pm$0.00          & 53.12$\pm$3.16          & 65.72$\pm$1.38          & 62.40$\pm$1.03          & 69.68$\pm$2.98   \\\hline      
\end{tabular}
}
\resizebox{\textwidth}{!}{
\begin{tabular}{l|lll|lll|lll}
\hline
                                                          & \multicolumn{3}{c|}{PL}                                                      & \multicolumn{3}{c|}{SCT}                                                     & \multicolumn{3}{c}{SSAN}                                                    \\\cline{2-10}
                                                          & \multicolumn{1}{c}{$HOS$} & \multicolumn{1}{c}{$OS^{\ast}$} & \multicolumn{1}{c|}{$UNK$} & \multicolumn{1}{c}{$HOS$} & \multicolumn{1}{c}{$OS^{\ast}$} & \multicolumn{1}{c|}{$UNK$} & \multicolumn{1}{c}{$HOS$} & \multicolumn{1}{c}{$OS^{\ast}$} & \multicolumn{1}{c}{$UNK$} \\\hline
Image $\Rightarrow$ Text & 27.11$\pm$2.26          & 18.28$\pm$2.10          & 54.00$\pm$3.18          & 38.51$\pm$2.45          & 29.28$\pm$2.77          & 56.56$\pm$3.00          & 39.52$\pm$2.03          & 30.16$\pm$2.18          & 57.52$\pm$3.10          \\
Text $\Rightarrow$ Image & 56.64$\pm$1.04          & 52.00$\pm$0.85          & 62.80$\pm$3.02          & 78.31$\pm$1.58          & 76.44$\pm$2.00          & 80.28$\pm$2.47          & 77.23$\pm$1.48          & 75.36$\pm$1.73          & 79.20$\pm$2.50       \\\hline  
\end{tabular}
}
\resizebox{\textwidth}{!}{
\begin{tabular}{l|lll|lll|lll}
\hline
                                                          & \multicolumn{3}{c|}{STN}                                                     & \multicolumn{3}{c|}{SL}                                                      & \multicolumn{3}{c}{OSHeDA}                                                  \\\cline{2-10}
                                                          & \multicolumn{1}{c}{$HOS$} & \multicolumn{1}{c}{$OS^{\ast}$} & \multicolumn{1}{c|}{$UNK$} & \multicolumn{1}{c}{$HOS$} & \multicolumn{1}{c}{$OS^{\ast}$} & \multicolumn{1}{c|}{$UNK$} & \multicolumn{1}{c}{$HOS$} & \multicolumn{1}{c}{$OS^{\ast}$} & \multicolumn{1}{c}{$UNK$} \\\hline
Image $\Rightarrow$ Text & 39.96$\pm$2.11          & 30.24$\pm$2.21          & 59.20$\pm$3.16          & 37.80$\pm$2.48          & 28.68$\pm$2.70          & 55.68$\pm$3.07          & \textbf{40.27$\pm$2.40} & \textbf{31.60$\pm$2.79} & 56.68$\pm$3.22          \\
Text $\Rightarrow$ Image & 75.54$\pm$1.70          & 72.56$\pm$2.05          & 78.80$\pm$2.79          & 76.40$\pm$1.47          & 74.52$\pm$1.68          & 78.40$\pm$2.46          & \textbf{85.93$\pm$1.39} & \textbf{82.92$\pm$2.11} & \textbf{89.40$\pm$1.51}     \\\hline    
\end{tabular}
}
\end{table}

\begin{table}[H]
    \centering
    \caption{Prediction performances of RL-OSHeDA as well as baselines for all domain adaptation tasks in PTB-XL dataset.}
    \label{tab:todo_7}
    \resizebox{\textwidth}{!}{
    \begin{tabular}{l|lll|lll|lll}
    \hline
    \multicolumn{10}{c}{PTB-XL}\\\hline
                                                          & \multicolumn{3}{c|}{DS3L}                                                    & \multicolumn{3}{c|}{KPG}                                                     & \multicolumn{3}{c}{OPDA}                                                    \\\cline{2-10}
                                                          & \multicolumn{1}{c}{$HOS$} & \multicolumn{1}{c}{$OS^{\ast}$} & \multicolumn{1}{c|}{$UNK$} & \multicolumn{1}{c}{$HOS$} & \multicolumn{1}{c}{$OS^{\ast}$} & \multicolumn{1}{c|}{$UNK$} & \multicolumn{1}{c}{$HOS$} & \multicolumn{1}{c}{$OS^{\ast}$} & \multicolumn{1}{c}{$UNK$} \\\hline
Digital ECG $\Rightarrow$ Paper ECG & 30.30$\pm$1.19 & 34.95$\pm$0.58          & 26.74$\pm$1.82          & N/A & N/A                     & N/A                     & 31.47$\pm$1.22 & 36.35$\pm$0.63          & 27.74$\pm$1.86   \\\hline      
\hline
                                                          & \multicolumn{3}{c|}{PL}                                                      & \multicolumn{3}{c|}{SCT}                                                     & \multicolumn{3}{c}{SSAN}                                                    \\\cline{2-10}
                                                          & \multicolumn{1}{c}{$HOS$} & \multicolumn{1}{c}{$OS^{\ast}$} & \multicolumn{1}{c|}{$UNK$} & \multicolumn{1}{c}{$HOS$} & \multicolumn{1}{c}{$OS^{\ast}$} & \multicolumn{1}{c|}{$UNK$} & \multicolumn{1}{c}{$HOS$} & \multicolumn{1}{c}{$OS^{\ast}$} & \multicolumn{1}{c}{$UNK$} \\\hline
Digital ECG $\Rightarrow$ Paper ECG & 26.18$\pm$1.37          & 36.43$\pm$0.55          & 20.43$\pm$1.66          & 26.23$\pm$1.65          & \textbf{46.48$\pm$0.71} & 18.27$\pm$1.60          & 25.16$\pm$1.47          & 40.40$\pm$0.65          & 18.27$\pm$1.54       \\
\hline    
\end{tabular}
}
\end{table}

\begin{table}[H]
    \centering
    \caption{Prediction performances of RL-OSHeDA as well as baselines for all domain adaptation tasks in ImageCLEF-DA dataset.}
    \label{tab:todo_2}
    \resizebox{\textwidth}{!}{
    \begin{tabular}{l|lll|lll|lll}
    \hline
    \multicolumn{10}{c}{ImageCLEF-DA}\\\hline
                                                          & \multicolumn{3}{c|}{DS3L}                                                    & \multicolumn{3}{c|}{KPG}                                                     & \multicolumn{3}{c}{OPDA}                                                    \\\cline{2-10}
                                                          & \multicolumn{1}{c}{$HOS$} & \multicolumn{1}{c}{$OS^{\ast}$} & \multicolumn{1}{c|}{$UNK$} & \multicolumn{1}{c}{$HOS$} & \multicolumn{1}{c}{$OS^{\ast}$} & \multicolumn{1}{c|}{$UNK$} & \multicolumn{1}{c}{$HOS$} & \multicolumn{1}{c}{$OS^{\ast}$} & \multicolumn{1}{c}{$UNK$} \\\hline
Bing (Reset-50) $\Rightarrow$ Caltech (VGG-19)       & 70.23$\pm$1.83          & 66.23$\pm$2.70          & 74.77$\pm$2.45          & 30.67$\pm$0.63          & 25.73$\pm$0.00          & 38.26$\pm$2.74          & 64.88$\pm$1.58          & 58.83$\pm$1.88          & 72.58$\pm$2.60          \\
Bing (Reset-50) $\Rightarrow$ ImageNet (VGG-19)      & 62.00$\pm$2.01          & 56.90$\pm$2.66          & 68.19$\pm$2.92          & 35.41$\pm$0.91          & 28.57$\pm$0.00          & 47.26$\pm$2.80          & 57.22$\pm$1.72          & 49.68$\pm$2.15          & 67.64$\pm$2.76          \\
Bing (Reset-50) $\Rightarrow$ PascalVOC (VGG-19)     & 57.21$\pm$1.86          & 52.42$\pm$2.34          & 63.01$\pm$2.84          & 31.10$\pm$0.49          & 24.74$\pm$0.00          & 42.55$\pm$2.93          & 54.47$\pm$1.75          & 48.13$\pm$2.15          & 62.84$\pm$2.92          \\
Bing (VGG-19) $\Rightarrow$ Caltech (Reset-50)       & 72.67$\pm$1.71          & 68.03$\pm$2.55          & 78.12$\pm$2.23          & 25.93$\pm$0.95          & 22.29$\pm$0.00          & 31.16$\pm$2.59          & 63.25$\pm$1.89          & 56.22$\pm$2.62          & 72.45$\pm$2.76          \\
\hline    
\end{tabular}
}
\end{table}

\begin{table}[H]
    \centering
    \caption{Prediction performances of RL-OSHeDA as well as baselines for all domain adaptation tasks in Office \& Caltech256 dataset.}
    \label{tab:s6}
    \resizebox{\textwidth}{!}{
    \begin{tabular}{l|lll|lll|lll}
    \hline
    \multicolumn{10}{c}{Office \& Caltech256}\\\hline
                                                          & \multicolumn{3}{c|}{DS3L}                                                    & \multicolumn{3}{c|}{KPG}                                                     & \multicolumn{3}{c}{OPDA}                                                    \\\cline{2-10}
                                                          & \multicolumn{1}{c}{$HOS$} & \multicolumn{1}{c}{$OS^{\ast}$} & \multicolumn{1}{c|}{$UNK$} & \multicolumn{1}{c}{$HOS$} & \multicolumn{1}{c}{$OS^{\ast}$} & \multicolumn{1}{c|}{$UNK$} & \multicolumn{1}{c}{$HOS$} & \multicolumn{1}{c}{$OS^{\ast}$} & \multicolumn{1}{c}{$UNK$} \\\hline
Amazon (DeCAF6) $\Rightarrow$ Caltech (SURF) & 46.08$\pm$1.41          & 37.41$\pm$1.60          & 60.15$\pm$2.05          & 36.51$\pm$0.69          & 30.20$\pm$0.00          & 46.22$\pm$2.22          & 43.26$\pm$1.58          & 33.61$\pm$1.87          & 60.84$\pm$2.01          \\
Amazon (DeCAF6) $\Rightarrow$ DSLR (SURF)    & 70.73$\pm$4.37          & 61.96$\pm$6.58          & 82.57$\pm$4.50          & 55.39$\pm$1.70          & 46.05$\pm$0.00          & 69.59$\pm$5.04          & 61.74$\pm$4.41          & 50.73$\pm$4.98          & 79.32$\pm$4.97          \\
Amazon (DeCAF6) $\Rightarrow$ Webcam (SURF)  & 71.64$\pm$2.73          & 67.43$\pm$4.05          & 76.62$\pm$3.64          & 61.25$\pm$1.62          & 56.91$\pm$0.00          & 66.34$\pm$3.81          & 61.79$\pm$2.88          & 53.28$\pm$3.61          & 74.00$\pm$4.06          \\
Amazon (SURF) $\Rightarrow$ Caltech (DeCAF6) & 68.29$\pm$1.31          & 65.51$\pm$1.83          & 71.37$\pm$1.98          & 36.27$\pm$0.91          & 34.10$\pm$0.00          & 38.80$\pm$2.06          & 58.80$\pm$1.40          & 52.44$\pm$2.00          & 67.48$\pm$1.85          \\
\hline    
\end{tabular}
}
\end{table}

\begin{table}[H]
    \centering
    \caption{Pairwise p-values from the Nemenyi test conducted on 56 domain adaptation tasks. P-values less than 0.05 indicate a statistically significant difference in prediction performance between the two corresponding methods.}
    \label{tab:s7}
    \begin{tabular}{r|ccccccccc}
    \hline
          & DS3L  & KPG   & OPDA  & PL    & SCT   & SSAN  & STN   & SL    & RL-OSHeDA \\\hline
DS3L      & 1.000 & 0.001 & 0.001 & 0.001 & 0.525 & 0.900 & 0.900 & 0.900 & \textbf{0.001}     \\
KPG       & 0.001 & 1.000 & 0.283 & 0.900 & 0.001 & 0.001 & 0.001 & 0.001 & \textbf{0.001}     \\
OPDA      & 0.001 & 0.283 & 1.000 & 0.050 & 0.001 & 0.001 & 0.001 & 0.001 & \textbf{0.001}     \\
PL        & 0.001 & 0.900 & 0.050 & 1.000 & 0.001 & 0.001 & 0.001 & 0.001 & \textbf{0.001}     \\
SCT       & 0.525 & 0.001 & 0.001 & 0.001 & 1.000 & 0.589 & 0.062 & 0.525 & \textbf{0.009}     \\
SSAN      & 0.900 & 0.001 & 0.001 & 0.001 & 0.589 & 1.000 & 0.900 & 0.900 & \textbf{0.001}     \\
STN       & 0.900 & 0.001 & 0.001 & 0.001 & 0.062 & 0.900 & 1.000 & 0.900 & \textbf{0.001}     \\
SL        & 0.900 & 0.001 & 0.001 & 0.001 & 0.525 & 0.900 & 0.900 & 1.000 & \textbf{0.001}     \\
RL-OSHeDA & \textbf{0.001} & \textbf{0.001} & \textbf{0.001} & \textbf{0.001} & \textbf{0.009} & \textbf{0.001} & \textbf{0.001} & \textbf{0.001} & \textbf{1.000}    \\\hline
\end{tabular}
\end{table}


\end{document}